\documentclass[final,5p,times,twocolumn]{elsarticle}

\usepackage{amssymb}
\usepackage{amsmath}

\usepackage{lineno} 

\usepackage[colorlinks=true]{hyperref}
\usepackage{booktabs}
\usepackage{threeparttable}
\usepackage{multirow}
\usepackage{float}
\usepackage{graphicx}
\usepackage{subcaption}
\usepackage{tabularx}
\usepackage{wrapfig}
\usepackage{array}
\usepackage{caption}
\usepackage{algorithm}
\usepackage{algpseudocode}

\begin{document}

\begin{frontmatter}



\title{CFGPNet: Cross-Attention-Based Fused Gradient Programmed Network Framework for Multispectral Object Detection}


\author[aut]{Nima Hatami}
\ead{nima.h@aut.ac.ir}

\author[aut]{Karim Faez}
\ead{kfaez@aut.ac.ir}

\author[aut]{Saeed Sharifian}
\ead{sharifian_s@aut.ac.ir}

\author[aut]{Hamidreza Amindavar\corref{cor1}}
\ead{hamidami@aut.ac.ir}

\cortext[cor1]{Corresponding author.}

\affiliation[aut]{
	organization={Department of Electrical Engineering, Amirkabir University of Technology},
	postcode={15914},
	state={Tehran},
	country={Iran}
}

\begin{abstract}
Multispectral object detection combines visible and thermal imagery to improve perception under challenging illumination and environmental conditions. However, differences in modality appearance and reliability can introduce redundant or conflicting responses, limiting the use of complementary information. Complex fusion mechanisms further increase computational cost, creating a persistent trade-off between detection accuracy and efficiency. To address these challenges, CFGPNet is proposed, a cross-attention-based fused gradient programmed network. The framework incorporates re-parameterized RepViT blocks into the YOLOv9 architecture to strengthen spatial and channel representations while maintaining efficient feature extraction. Cross Computation Efficient Attention (CrossCEA) exchanges spatial attention maps between modalities at multiple detection scales, allowing each stream to emphasize regions supported by the other while preserving modality-specific information. Attention Selection and Aggregation Fusion (ASAF) combines dense feature aggregation with selection of the strongest responses from multiple attention branches to form compact, discriminative fused representations. A programmable gradient information pathway provides auxiliary supervision during training to improve feature learning. This pathway is removed after training, adding no parameters or operations at inference. Experiments on FLIR, M3FD, LLVIP, VEDAI, and MFAD demonstrate favorable accuracy–efficiency trade-offs across three model scales, with the smallest variant requiring 15.3 million parameters and 56.9 GFLOPs. The code is available at \url{https://github.com/NimaHatami99/CFGPNet}.
 
\end{abstract}




\end{frontmatter}



\section{Introduction}
\label{sec_1}

\begin{figure*}[h]
	\centering
	\includegraphics[
	width=0.8\textwidth,
	trim={1mm 1mm 1mm 1mm},
	clip
	]{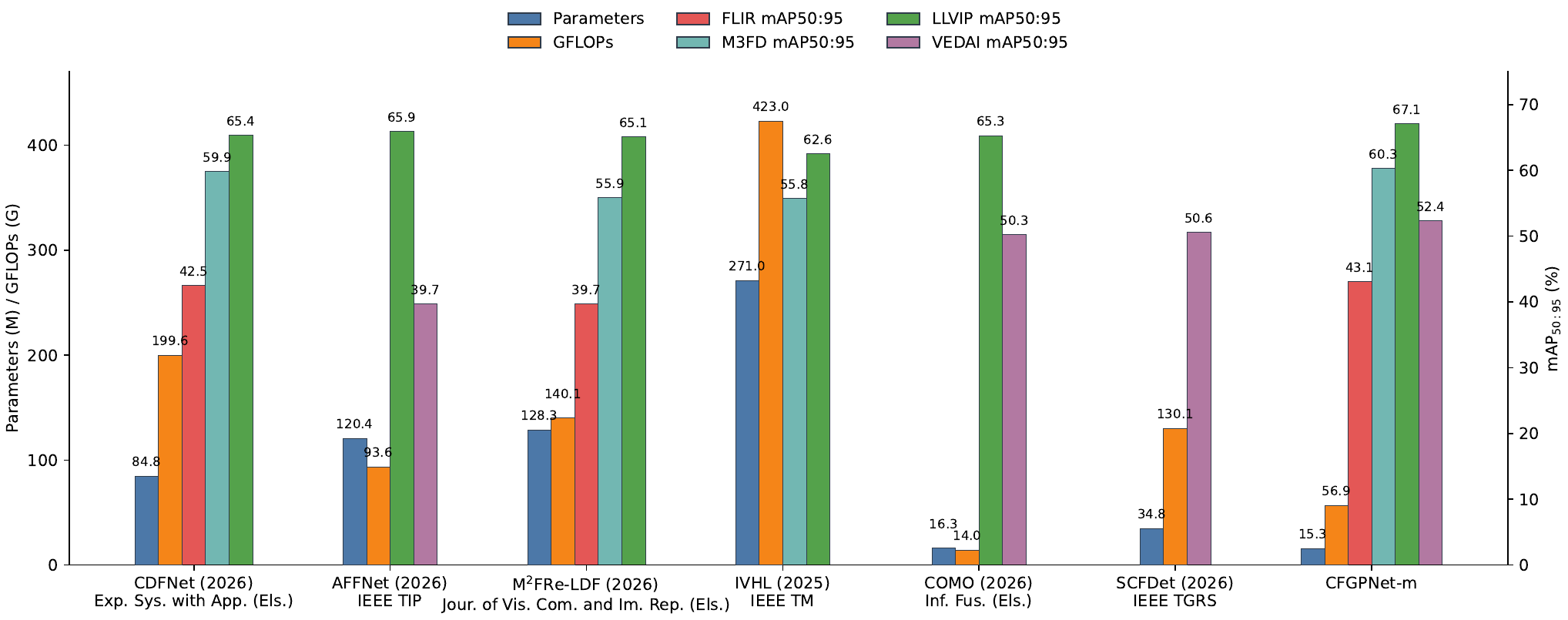}
	\caption{CFGPNet-m versus state-of-the-art multispectral detectors.}
	\label{fig_cfgpnet_comparison}
\end{figure*} 

Multispectral object detection combines visible and thermal imagery to improve perception under changing illumination and weather. Visible images provide color, texture, and fine spatial detail, whereas thermal images preserve object-related responses when visible contrast is degraded. Their complementarity is useful for driver assistance, pedestrian detection, and vehicle detection in aerial imagery, but the modalities differ in appearance, noise, and spatial reliability. Simple concatenation or fixed fusion can therefore propagate redundant or conflicting responses and may even reduce the benefit of the stronger modality \cite{cao2023multimodal,zeng2024mmi,yu2026m2fre}.

Recent methods improve fusion with global cross-modal reasoning, iterative attention, illumination-aware weighting, and hierarchical interaction \cite{qingyun2021cross,shen2024icafusion,hu2025ei,wu2025dhanet}. Other approaches address channel--spatial complementarity and feature alignment, including cross-dimension interaction, progressive channel exchange, adaptive dual cross-attention, and offset-aware or selective fusion \cite{wu2026cdfnet,li2025crossmodalnet,zhang2026adaptive,xue2026transfer,luo2026scfdet}. These designs improve robustness, yet their additional streams and attention blocks increase computational cost, while modality failure remains possible when one stream is weak or misaligned. Adaptive distillation reduce deployment cost while highlighting the trade-off between compactness and complementary feature learning \cite{chen2025amfd}.

To address these issues, CFGPNet is proposed, a cross-attention-based fused gradient programmed network for efficient RGB--T detection. The framework uses a GELAN\footnote{generalized efficient layer aggregation network}-based backbone and a training-only programmable gradient path to improve feature learning without retaining the auxiliary computation at inference \cite{wang2024yolov9}. CrossCEA exchanges modality-derived spatial attention maps at multiple detection scales, allowing each stream to emphasize regions supported by the other. The resulting features are combined by ASAF, which selects and aggregates complementary responses through dense and attention-guided paths. Three model scales provide different accuracy--efficiency operating points.

As shown in Fig.~\ref{fig_cfgpnet_comparison}, CFGPNet performs favorably across the benchmark datasets, combining high mAP$_{50:95}$ with low parameter count and GFLOPs. The main contributions are: 
\begin{itemize}
	\item CFGPNet, a cross-attention-based fused gradient programmed framework for RGB--T detection, with three model scales spanning practical accuracy--efficiency operating points.
	
	\item A RepViT-based re-parameterized GELAN backbone that strengthens spatial and channel representations.
	
	\item CrossCEA, which exchanges modality-derived spatial attention maps across detection scales before fusion.
	
	\item ASAF, which combines dense aggregation with branch-selective attention to produce compact, discriminative multiscale fused features.
	
	\item Extensive evaluation on FLIR \cite{teledyneflir_adas_dataset}, M3FD \cite{liu2022target}, LLVIP \cite{jia2021llvip}, VEDAI \cite{razakarivony2016vehicle}, and MFAD \cite{hu2025ei}, demonstrating favorable accuracy--efficiency performance against recent multispectral detectors.
\end{itemize}

\section{Related Work}
\label{sec_2}

\subsection{Multispectral Object Detection Methods}
\label{sec_2_1}

Multispectral object detection combines the complementary properties of visible and infrared (IR) imagery: visible images provide rich texture and color information, whereas IR images can remain informative under some adverse conditions. Most early systems therefore adopted two-stream backbones with intermediate fusion. CFT uses Transformer self-attention to model global relationships between RGB and thermal features, while ICAFusion performs iterative query-guided cross-attention to refine the two streams \cite{qingyun2021cross,shen2024icafusion}. MMI-Det jointly optimizes fusion and detection with contour enhancement, contrastive modality bridging, and information-guided supervision, thereby avoiding the objective mismatch of a separate fusion-then-detection pipeline \cite{zeng2024mmi}. For lightweight mid-level fusion, CSSA replaces weak channels across modalities and applies parameter-free spatial attention, whereas C$^2$DFF-Net augments differential feature learning with spatial--frequency fusion for small objects in remote-sensing imagery \cite{cao2023multimodal,11180153}.

Recent detectors increasingly make fusion adaptive to illumination, scale, and deployment constraints. EI$^2$Det combines cross-modal interaction with illumination-aware weighting and edge-guided fusion, and DHANet uses modality-specific enhancement, hierarchical interaction, and decision fusion for object detection in images captured by drones \cite{hu2025ei,wu2025dhanet}. CDFNet models complementary information at both channel and spatial levels before cross-dimension fusion; CrossModalNet progressively aligns low-level features and refines high-level semantics with convolutional and bidirectional attention \cite{wu2026cdfnet,li2025crossmodalnet}. Lightweight alternatives include PFI-Net, which separates parallel interaction paths to reduce cross-modal interference, and SAFF, which emphasizes spatially reliable IR cues while retaining RGB semantics \cite{wang2025pfi,xiao2026saff}. Other efforts reduce the cost of two-stream inference through adaptive modal-fusion distillation or foundation-model features \cite{chen2025amfd,liu2025fusion}. Nevertheless, redundant branches, modality imbalance, and fusion-induced degradation remain recurring concerns, especially when the two sensors are misaligned or one modality is unreliable \cite{yu2026m2fre}. 

\subsection{Cross-Modality Information Interaction}
\label{sec_2_2}

Cross-modal interaction has evolved from direct concatenation or summation toward reciprocal, selective, and alignment-aware exchange. CFT and ICAFusion use global Transformer attention and iterative cross-attention, respectively, to capture long-range dependencies that are difficult to obtain with local convolutions \cite{qingyun2021cross,shen2024icafusion}. CrossModalNet combines local convolutional attention with bidirectional channel interaction, while ADCA-Net dynamically switches the reference modality through dual cross-attention and adaptive max--average downsampling \cite{li2025crossmodalnet,zhang2026adaptive}. CDFIT further separates dual-stream feature interaction to alleviate modality imbalance, and CDFNet explicitly models cross-modal dependencies in both channel and spatial dimensions \cite{huang2026cdfit,wu2026cdfnet}.

Several methods refine how information is transferred when one modality is noisy or spatially displaced. CARNet decomposes features into strong and weak components and reconstructs each modality through cross-attention-guided interaction; PFI-Net maintains parallel interaction paths to limit destructive transfer; and M$^2$FRe-LDF first strengthens intra-modality channels before using deformable convolution for local cross-modal fusion \cite{wang2025carnet,wang2025pfi,yu2026m2fre}. MMI-Det uses contrastive learning to reduce the modality gap, whereas EI$^2$Det and DHANet adapt the interaction to illumination and modality characteristics \cite{zeng2024mmi,hu2025ei,wu2025dhanet}. For severe registration errors, COMO introduces cross-Mamba interaction with offset-guided fusion, TGRDet performs filter-assisted alignment followed by transfer graph reasoning, and SCFDet combines mutual-guided alignment with selective salient-region fusion \cite{liu2026cross,xue2026transfer,luo2026scfdet}. AFFNet extends this idea to local consistency-based weighting, assigning different fusion weights to spatial regions according to the reliability of RGB and IR evidence \cite{tang2026adaptive}. These approaches improve correspondence and complementarity, but their interaction modules can increase memory, latency, or architectural complexity.
 
\subsection{Multispectral Attention Approach}
\label{sec_2_3}

Attention is used in multispectral detection to select informative channels, spatial regions, and scales while suppressing modality-specific noise. CrossFormer propagates bidirectional cross-guided attention hierarchically across scales \cite{lee2024crossformer}. DHANet uses asymmetric global--local attention for modality-specific enhancement, while CSSA combines channel switching with parameter-free spatial attention to obtain a compact fusion operation \cite{wu2025dhanet,cao2023multimodal}. C$^2$DFF-Net also couples channel and spatial attention with frequency-domain features, illustrating the benefit of jointly modeling spatial and spectral evidence \cite{11180153}.

Attention has also been coupled with task-aware, fine-grained, or interpretable fusion. EI$^2$Det predicts illumination-dependent modality weights and preserves boundaries with edge-guided fusion; CDFIT uses dual-stream Transformer interaction to control cross-modal interference; and SCFDet adds top-$K$ region selection, sparse expert routing, and channel-aware enhancement \cite{hu2025ei,huang2026cdfit,luo2026scfdet}. SAFF and AFFNet use spatially aware or connected-region consistency to focus computation on reliable local cues, whereas CAMF derives interpretable channel weights from class activation mapping for image fusion \cite{xiao2026saff,tang2026adaptive,tang2023camf}. A unified fusion-enhanced network further transfers spatial and channel attention across modalities and across high-level tasks \cite{liu2025fusion}. These designs demonstrate the value of adaptive attention, but they often rely on multiple attention branches or heavy Transformer blocks.

The efficiency problem is also addressed at the backbone and optimization levels. YOLOv9 introduces GELAN and training-only Programmable Gradient Information (PGI), providing a compact inference path with improved gradient supervision \cite{wang2024yolov9}. AMFD distills modality-specific teacher features into an efficient student rather than forcing the student to imitate a fused representation \cite{chen2025amfd}. The modality-failure analysis of M$^2$FRe-LDF further shows that strong cross-modal fusion is insufficient when individual streams are under-represented \cite{yu2026m2fre}. These observations motivate CFGPNet, which combines a YOLOv9-based efficient backbone and training path with cross-modal attention exchange and compact attention-guided aggregation.

\section{Method}
\label{sec_3}

\subsection{Overview of the Proposed Framework}
\label{sec_3_1} 

\begin{figure*}[h]
	\centering
	\includegraphics[width=0.8\textwidth, trim={6mm 6mm 6mm 6mm}, clip]{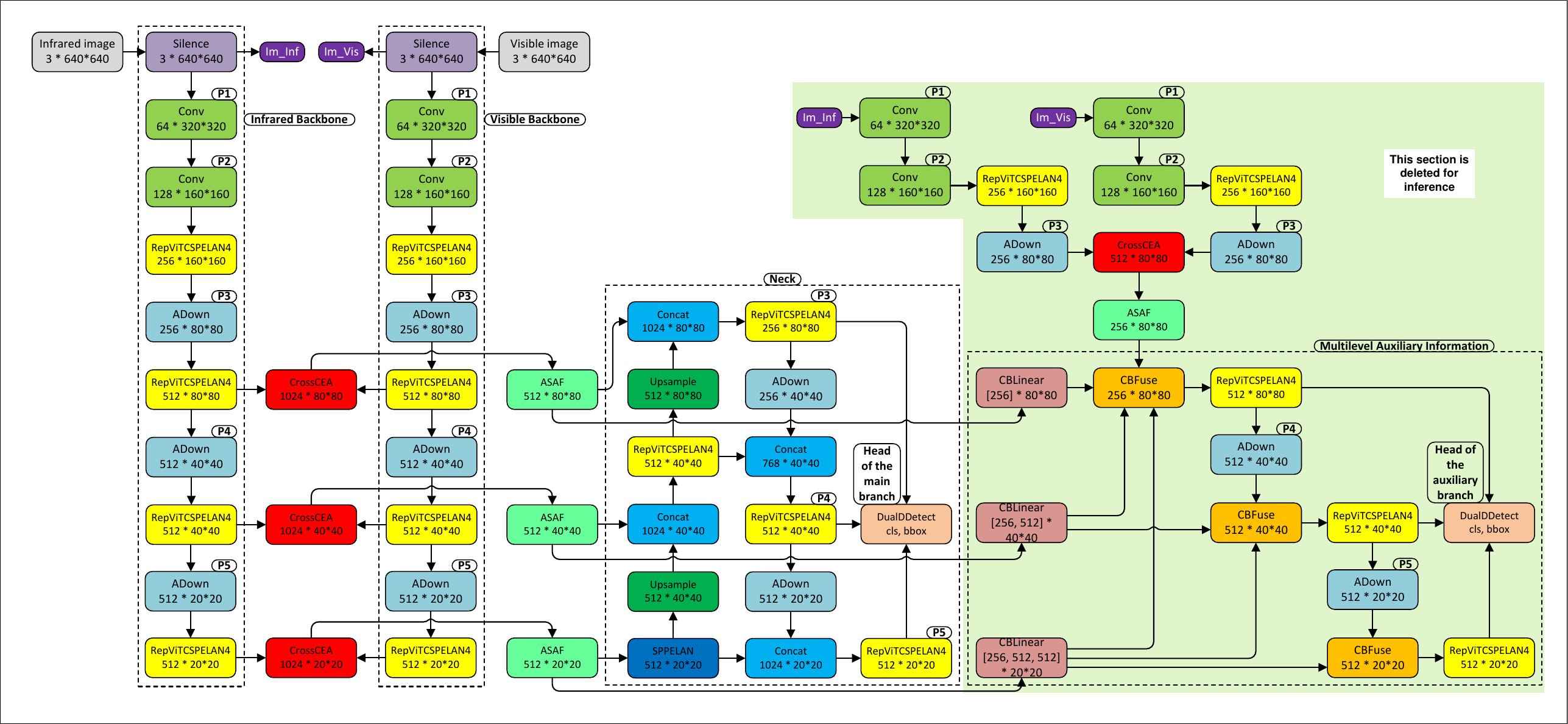}
	\captionsetup{width=0.8\textwidth}
	\caption{\centering Overview of the proposed CFGPNet framework. The auxiliary branch (green area) is used only during training for gradient enhancement and is removed during inference. The Silence module splits the six-channel input into two three-channel tensors: infrared (channels 1--3) and visible (channels 4--6). The diagram reports each module name, the number of output channels in brackets, and the output spatial dimension, all based on a 640$\times$640 input image size.}\label{fig_1}
\end{figure*}

In CFGPNet, modality-specific feature extraction, cross-modal refinement, and multispectral fusion are integrated within a YOLOv9-based architecture for RGB--T object detection \cite{wang2024yolov9}. As shown in Fig.~\ref{fig_1}, registered visible and thermal images are processed by independent backbones, from which modality-specific multiscale features are obtained.
At the P3--P5 detection scales, spatial attention maps are derived from each modality by CrossCEA and applied to the opposite stream before fusion. The enhanced features are then fused by ASAF and passed through the multiscale neck to the detection head. The improved backbone, CrossCEA, ASAF, and training-time auxiliary gradient path are described in the subsequent subsections.

\subsection{Improved Backbone Structure}
\label{sec_3_2} 

\begin{figure}[h]
	\centering
	\begin{minipage}{0.50\textwidth}
		\centering
		
		\resizebox{\linewidth}{!}{%
			\begin{tabular}{@{}c@{\hspace{1mm}}c@{}}
				\begin{tabular}[c]{@{}c@{}}
					\includegraphics[
					height=8cm,
					keepaspectratio,
					trim={6mm 6mm 6mm 6mm},
					clip
					]{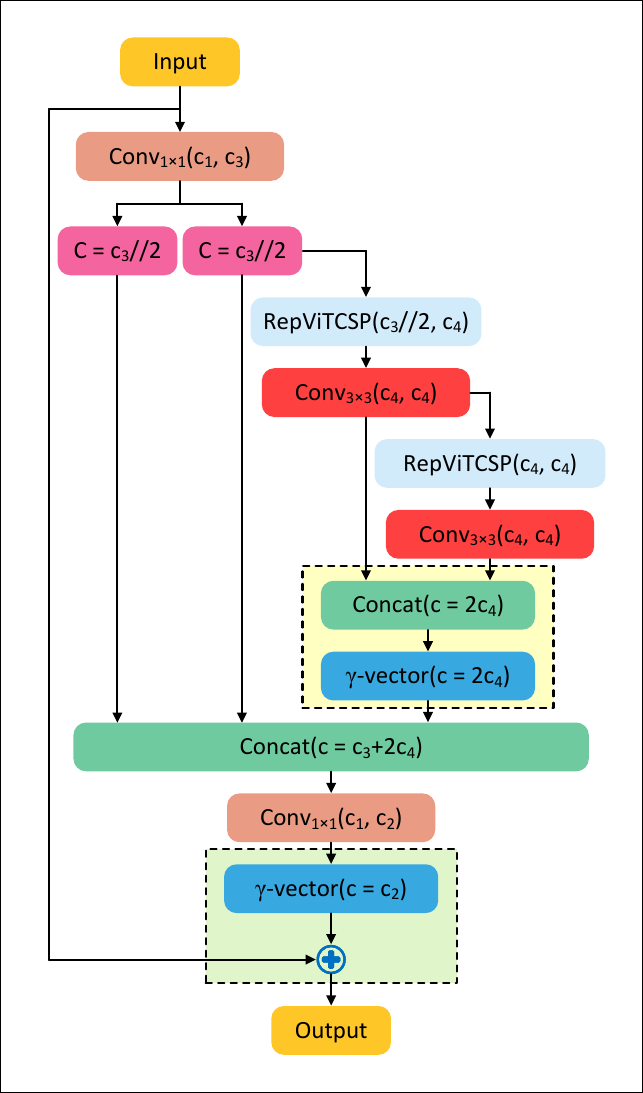}
					\\[-0.3ex]
					(a)
				\end{tabular}
				&
				\begin{tabular}[c]{@{}c@{}}
					\includegraphics[
					height=3.7cm,
					keepaspectratio,
					trim={6mm 6mm 6mm 6mm},
					clip
					]{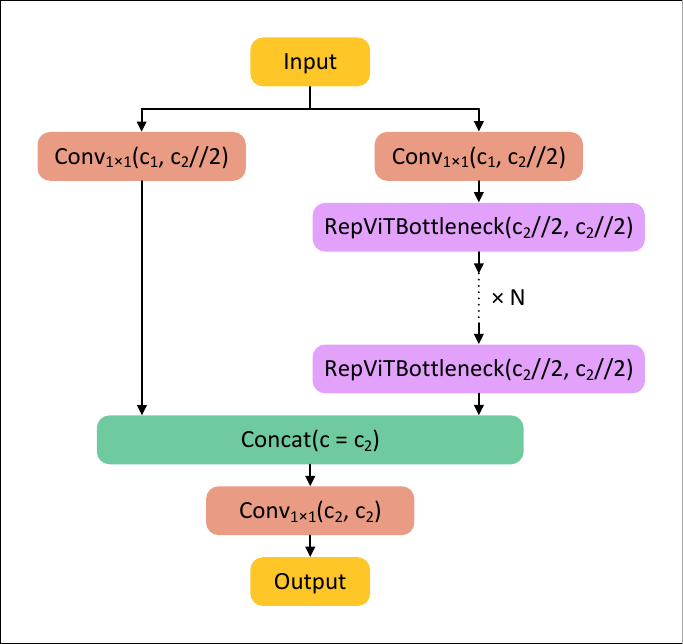}
					\\[-0.3ex]
					(b)
					\\[0.5ex]
					\includegraphics[
					height=3.7cm,
					keepaspectratio,
					trim={6mm 6mm 6mm 6mm},
					clip
					]{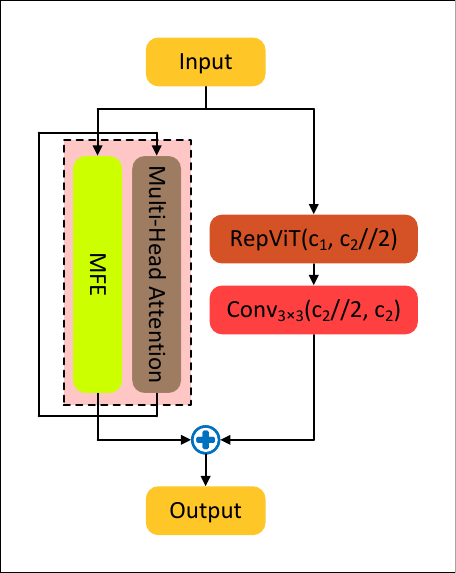}
					\\[-0.3ex]
					(c)
				\end{tabular}
			\end{tabular}%
		}
		
		\caption{Architectures of (a) RepViTCSPELAN4, (b) RepViTCSP, and (c) RepViTBottleneck. For each operation, the two channel arguments denote the input and output channels, respectively; $c_1$ and $c_2$ denote the module-level input and output channels, and $\oplus$ denotes element-wise addition.}
		\label{fig_2}
	\end{minipage}
\end{figure}

The GELAN architecture of YOLOv9 is adopted as the backbone template because different computational units can be incorporated within its flexible aggregation structure \cite{wang2024yolov9}. As shown in Fig.~\ref{fig_2}, the original RepVGG-based units are replaced with RepViT-based blocks \cite{wang2024repvit}, and RepViTBottleneck, RepViTCSP, and RepViTCSPELAN4 are thereby constructed for use in both the backbone and neck. 

\subsubsection{RepViT-Based GELAN Backbone}
\label{sec_3_2_1}

Within GELAN, the original RepVGG-derived computational units \cite{ding2021repvgg} are replaced by RepViT blocks \cite{wang2024repvit}, while the layer-aggregation topology is retained. Spatial and channel mixing are separated within each block, and structural re-parameterization is used to maintain a compact inference structure.

SE attention \cite{hu2018squeeze} is enabled in all RepViT blocks for channel-wise recalibration. As shown in Fig.~\ref{fig_2}, RepViTBottleneck is formed by replacing the original RepNBottleneck transformation, and RepViTCSP and RepViTCSPELAN4 are subsequently constructed by cascading this block.

\subsubsection{Multi-Head Attention}
\label{sec_3_2_2}

Multi-head self-attention \cite{vaswani2017attention} is incorporated into selected RepViTBottleneck blocks for global contextual refinement. Query, key, and value projections are implemented using $1\times1$ convolutions, while a depthwise $5\times5$ convolution is applied to the value branch for positional encoding. As shown in Fig.~\ref{fig_2}(c), the attention path is placed parallel to the skip connection and is enabled only in the GELAN modules of the neck to limit computational cost. 

\subsubsection{Marginal Feature Extractor}
\label{sec_3_2_3}

\begin{figure*}[h]
	\centering
	\begin{minipage}{0.70\textwidth}
		\centering
		\includegraphics[
		width=\linewidth,
		keepaspectratio,
		trim={6mm 6mm 6mm 6mm},
		clip
		]{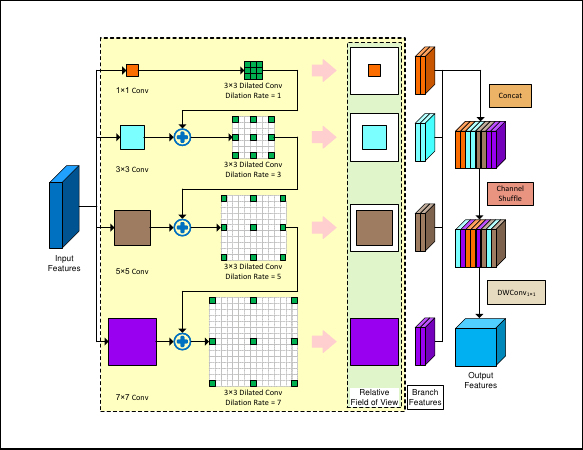}
		\caption{Architecture of the MFE module.}
		\label{fig_3}
	\end{minipage}
\end{figure*}

The Marginal Feature Extractor (MFE) is introduced to preserve peripheral and boundary responses that can be weakened by the center-concentrated effective receptive fields of deep CNNs \cite{luo2016understanding}. As shown in Fig.~\ref{fig_3}, four progressively connected branches are constructed using $1\times1$, $3\times3$, $5\times5$, and $7\times7$ convolutional kernels paired with $3\times3$ dilated convolutions at rates 1, 3, 5, and 7. 
Multiscale context is thereby captured without spatial downsampling \cite{yu2015multi}, while a new input projection is combined with the preceding output at each subsequent stage.

The branch outputs are concatenated, channel-shuffled with four groups \cite{zhang2018shufflenet}, and refined by a $1\times1$ depthwise convolution. Within RepViTBottleneck (Fig.~\ref{fig_2}(c)), MFE replaces the identity path. MFE is confined to the training-only auxiliary branch in CFGPNet-m/c and is retained in the CFGPNet-e backbone at inference.

\subsubsection{Optional Learnable Aggregation Factors}
\label{sec_3_2_4}

Two optional channel-wise scaling factors are considered in RepViTCSPELAN4, as illustrated in Fig.~\ref{fig_2}(a). The concatenated outputs of the two $3\times3$ branches are modulated by $\gamma_{x_{12}}$ before the final $1\times1$ projection (yellow area), whereas the projected output is modulated by $\gamma_{\mathrm{out}}$ before residual addition when $c_1=c_2$ (green area). Both factors are initialized to $10^{-2}$ so that their initial contributions remain small. These factors are examined as potential mechanisms for regulating feature aggregation and stabilizing optimization, with their actual effect evaluated in the ablation study. 

\subsection{Programmable Gradient Information}
\label{sec_3_3}

Programmable Gradient Information (PGI), introduced in YOLOv9 \cite{wang2024yolov9}, is adopted as a training-only auxiliary pathway. Multilevel features from the main RGB--T pathway are routed through the auxiliary branch, where predictions are generated solely for computation of an auxiliary loss. Gradients from this loss are propagated to the main-path parameters through the routed features, and supplementary supervision is thereby provided during optimization.

After training, the entire auxiliary branch is removed. Only the main detection head is retained, and its predictions alone are passed to non-maximum suppression. No predictions, parameters, or operations from the auxiliary branch are included during inference; consequently, the parameter count and GFLOPs are reduced relative to the full training architecture.

\subsection{Cross Computation Efficient Attention}
\label{sec_3_4}

\begin{figure}[t]
	\centering
	\begin{minipage}{0.5\textwidth}
		\centering
		
		\resizebox{\linewidth}{!}{%
			\begin{tabular}{@{}c@{\hspace{1mm}}c@{}}
				\begin{tabular}[c]{@{}c@{}}
					\includegraphics[
					trim={6mm 6mm 6mm 6mm},
					clip
					]{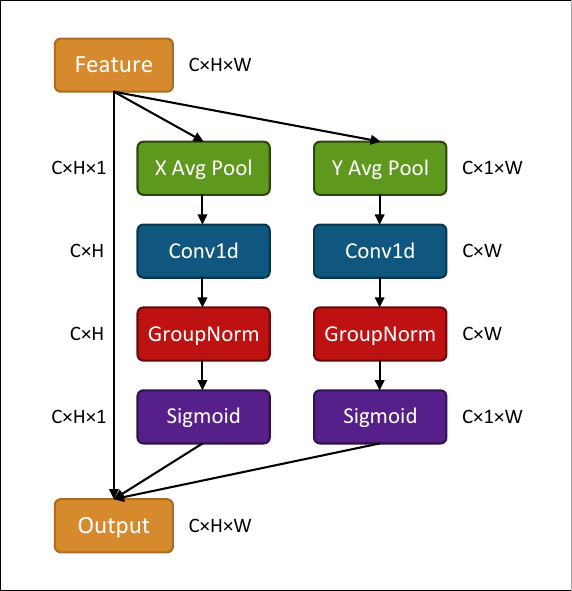}
					\\[-0.3ex]
					(a)
					\\[0.5ex]
					\includegraphics[
					trim={6mm 6mm 6mm 6mm},
					clip
					]{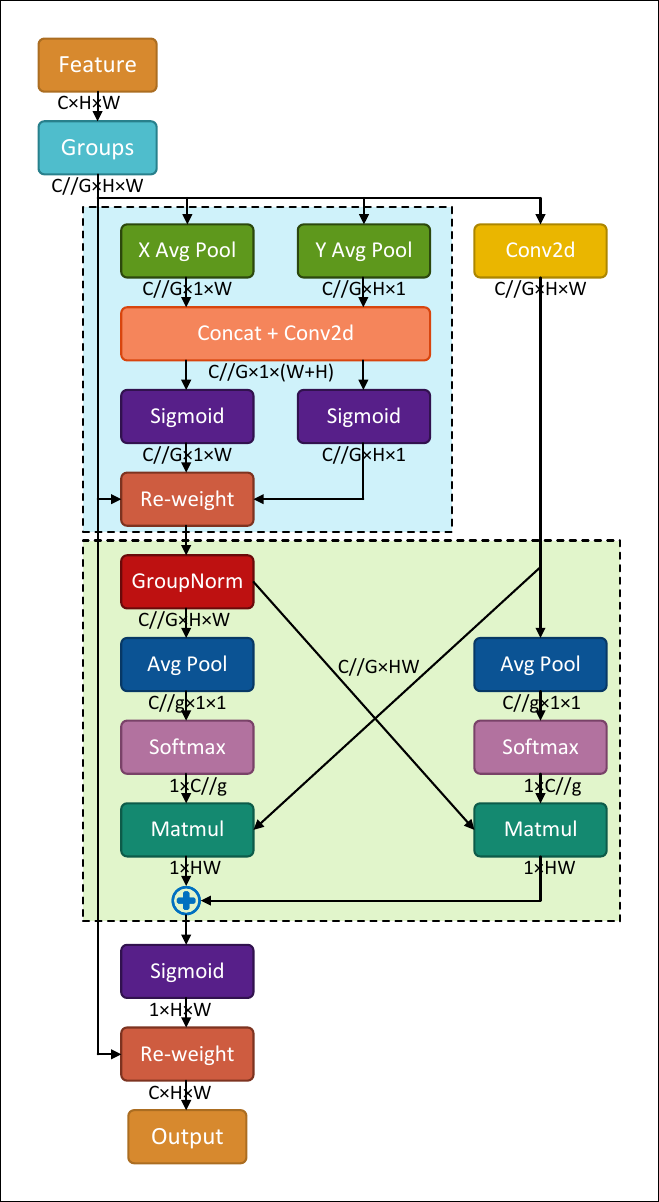}
					\\[-0.3ex]
					(b)
				\end{tabular}
				&
				\begin{tabular}[c]{@{}c@{}}
					\includegraphics[
					trim={6mm 6mm 6mm 6mm},
					clip
					]{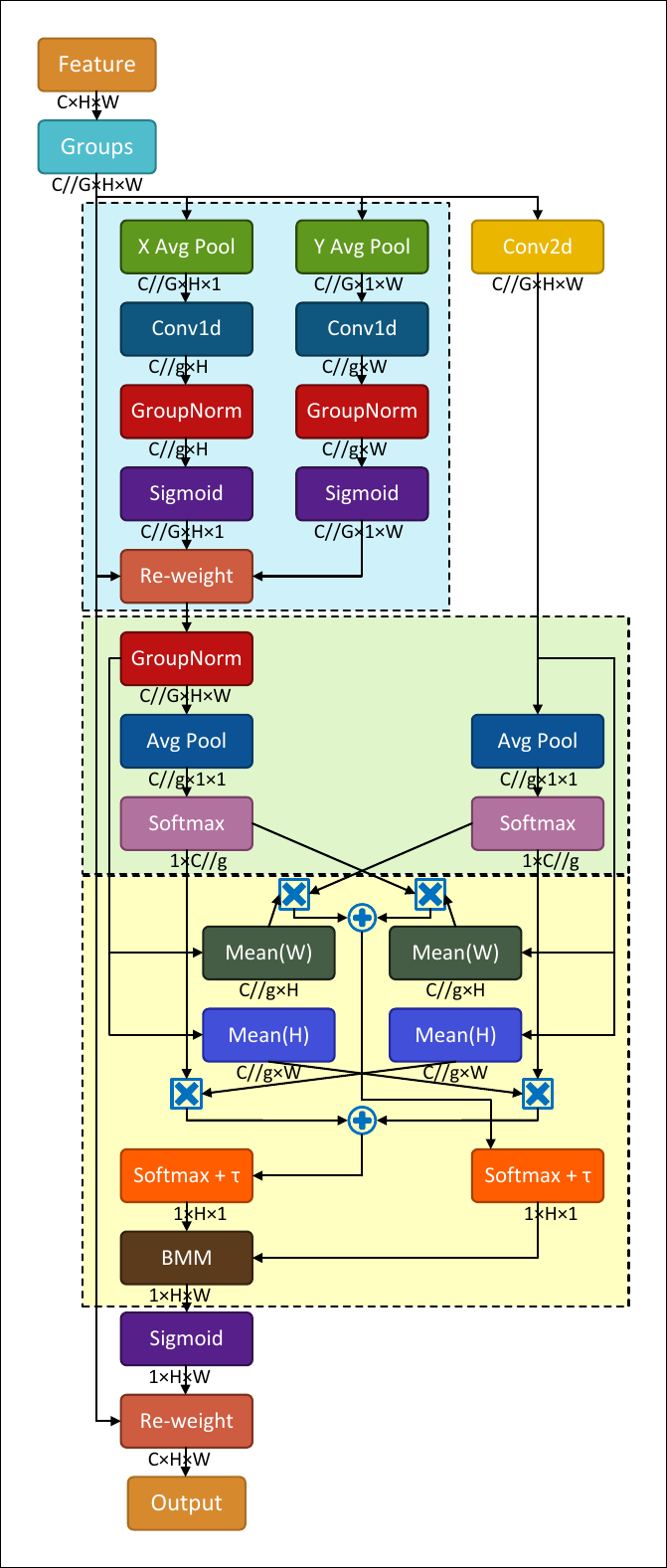}
					\\[-0.3ex]
					(c)
				\end{tabular}
			\end{tabular}%
		}
		
		\caption{Architectures of (a) ELA, (b) EMA, and (c) CEA.}
		\label{fig_4}
	\end{minipage}
\end{figure}

CrossCEA is applied to the P3--P5 features before ASAF fusion. Its underlying Computation Efficient Attention (CEA) operator, as illustrated in Fig.~\ref{fig_4}(c), is built on the grouped local--global structure of EMA\footnote{efficient multi-scale attention} \cite{ouyang2023efficient} (Fig.~\ref{fig_4}(b)). The Coordinate Attention (CA) module \cite{hou2021coordinate} used in the local branch of EMA is replaced by the axis-wise encoding of ELA\footnote{efficient location attention} \cite{xu2025ela} (Fig.~\ref{fig_4}(a)). Height- and width-wise descriptors from the resulting local branch are then cross-coupled with those from the global convolutional branch before a two-dimensional attention map is formed. CEA therefore remains closely related to EMA and ELA, with strengthened local--global interaction being its principal extension.

In CrossCEA, a CEA attention map is computed independently for each modality and applied to the opposite stream. Only the spatial attention maps are exchanged, while the modality-specific feature values are preserved until ASAF fusion. The operation at each feature scale is summarized in Algorithm~\ref{alg_crosscea}.

\begin{algorithm}[h]
	\caption{CrossCEA at one feature scale}
	\label{alg_crosscea}
	\footnotesize
	\begin{algorithmic}[1]
		\Require $F^{I},F^{V}\in\mathbb{R}^{B\times C\times H\times W}$,
		groups $G$, temperature $\tau$
		\Ensure Cross-enhanced features $\widetilde{F}^{I},\widetilde{F}^{V}$
		
		\Function{CEAMap}{$F,G,\tau$}
		\State $X \gets \Call{GroupChannels}{F,G}$
		\State $(p_h,p_w) \gets \Call{StripPool}{X}$
		\State $g_h \gets \sigma(\operatorname{GN}(\operatorname{DWConv1D}(p_h)))$
		\State $g_w \gets \sigma(\operatorname{GN}(\operatorname{DWConv1D}(p_w)))$
		\State $X_{\ell} \gets \operatorname{GN}(X\odot g_h\odot g_w)$
		\State $X_{g} \gets \operatorname{Conv}_{3\times3}(X)$
		\State $(H_{\ell},W_{\ell}) \gets \Call{StripPool}{X_{\ell}}$
		\State $(H_g,W_g) \gets \Call{StripPool}{X_g}$
		\State $q_{\ell} \gets \operatorname{Softmax}_{c}(\operatorname{GAP}(X_{\ell}))$
		\State $q_g \gets \operatorname{Softmax}_{c}(\operatorname{GAP}(X_g))$
		\State $s_h \gets \langle q_{\ell},H_g\rangle_c
		+\langle q_g,H_{\ell}\rangle_c$
		\State $s_w \gets \langle q_{\ell},W_g\rangle_c
		+\langle q_g,W_{\ell}\rangle_c$
		\State $\alpha_h \gets \operatorname{Softmax}_{h}(s_h/\tau)$
		\State $\alpha_w \gets \operatorname{Softmax}_{w}(s_w/\tau)$
		\State $\Omega \gets
		\Call{RestoreAndBroadcast}{
			\sigma(\operatorname{Outer}(\alpha_h,\alpha_w)),G}$
		\State \Return $\Omega$
		\EndFunction
		
		\State $\Omega^{I}\gets\Call{CEAMap}{F^{I},G,\tau}$
		\State $\Omega^{V}\gets\Call{CEAMap}{F^{V},G,\tau}$
		\State $\widetilde{F}^{I}\gets F^{I}\odot\Omega^{V}$
		\State $\widetilde{F}^{V}\gets F^{V}\odot\Omega^{I}$
		\State \Return $\widetilde{F}^{I},\widetilde{F}^{V}$
	\end{algorithmic}
\end{algorithm}

Here $B,C,H,W$ denote batch size, channels, height, and width. \textsc{GroupChannels} reshapes the input to $(BG,C/G,H,W)$, with $G=32$ and $G\mid C$. \textsc{StripPool} returns $(\operatorname{Avg}_w(X), \operatorname{Avg}_h(X))$, averaging over width and height, respectively; GAP averages both spatial axes. $\sigma$ denotes sigmoid, $\odot$ denotes element-wise multiplication with broadcasting, and the softmax subscript specifies its normalization axis. $\langle\cdot,\cdot\rangle_c$ sums channel-wise products, and $\operatorname{Outer}(a,b)=ab^\top$. \textsc{RestoreAndBroadcast} broadcasts each group map over its $C/G$ channels and restores shape $(B,C,H,W)$. DWConv1D uses kernel size 7; GN denotes group normalization with one channel per normalization group. CEA parameters are shared across modalities, and the axis encoders share convolution and normalization parameters. The temperature $\tau$ is learned and initialized to 1.

\subsection{Attention Selection and Aggregation Fusion Network}
\label{sec_3_5}

\begin{figure*}[h]
	\centering
	\includegraphics[
	width=0.8\textwidth,
	trim={6mm 6mm 6mm 6mm},
	clip
	]{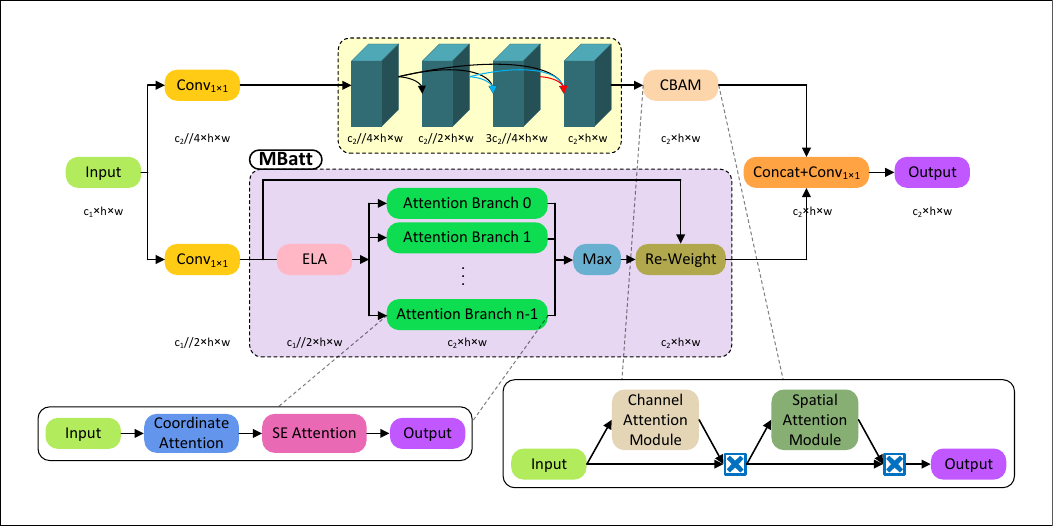}
	\caption{Architecture of the ASAF network.}
	\label{fig_5}
\end{figure*}

At each detection scale, the CrossCEA-enhanced visible and infrared features are fused into a single representation by the Attention Selection and Aggregation Fusion (ASAF) network. As shown in Fig.~\ref{fig_5}, the upper path combines dense connectivity \cite{huang2017densely} with CBAM\footnote{convolutional block attention module} refinement \cite{woo2018cbam}, whereas the MBatt\footnote{multi-branch attention} path incorporates ELA-based axis encoding \cite{xu2025ela}, coordinate attention \cite{hou2021coordinate}, and SE\footnote{squeeze and excitation} recalibration \cite{hu2018squeeze}.

In the dense path, modality features are progressively aggregated and refined. In MBatt, three attention candidates are generated after axis-wise encoding, and their element-wise maximum is selected at each channel and spatial position. The outputs of both paths are subsequently concatenated and projected to the required channel dimension. The complete procedure is summarized in Algorithm~\ref{alg_asaf}.

\begin{algorithm}[h]
	\caption{Attention Selection and Aggregation Fusion}
	\label{alg_asaf}
	\begin{algorithmic}[1]
		\Require Enhanced features $\tilde{F}^{I}$ and $\tilde{F}^{V}$;
		output channels $C_2=C_1/2$; $N_b=3$
		\Ensure Fused feature $\mathcal{Y}_{\mathrm{ASAF}}$
		\Statex \textit{$\Phi_{1\times1}^{a\rightarrow b}$ denotes a $1\times1$ convolution from $a$ to $b$ channels, followed by batch normalization and SiLU.}
		
		\State $\mathcal{X}\gets
		\operatorname{Cat}_{c}(\tilde{F}^{I},\tilde{F}^{V})$
		
		\State $\mathcal{D}_{0}\gets
		\Phi_{1\times1}^{C_1\rightarrow C_2/4}(\mathcal{X})$
		\State $\mathcal{D}_{1}\gets
		\Phi_{1\times1}^{C_2/4\rightarrow C_2/4}(\mathcal{D}_{0})$
		\For{$\ell=2,3,4$}
		\State $\mathcal{Q}_{\ell}\gets
		\mathcal{R}_{\ell}(\mathcal{D}_{\ell-1})$
		\State $\mathcal{D}_{\ell}\gets
		\operatorname{Cat}_{c}(\mathcal{D}_{\ell-1},\mathcal{Q}_{\ell})$
		\EndFor
		\State $\mathcal{U}_{\mathrm{d}}\gets
		\operatorname{CBAM}(\mathcal{D}_{4})$
		
		\State $\mathcal{P}\gets
		\Phi_{1\times1}^{C_1\rightarrow C_2}(\mathcal{X})$
		\State $\mathcal{E}_{h}\gets
		\sigma\!\left(\operatorname{GN}
		(\operatorname{Conv1D}(\operatorname{Avg}_{w}(\mathcal{P})))\right)$
		\State $\mathcal{E}_{w}\gets
		\sigma\!\left(\operatorname{GN}
		(\operatorname{Conv1D}(\operatorname{Avg}_{h}(\mathcal{P})))\right)$
		\State $\mathcal{P}_{\mathrm{e}}\gets
		\mathcal{P}\odot\mathcal{E}_{h}\odot\mathcal{E}_{w}$
		
		\For{$i=1,\ldots,N_b$}
		\State $\mathcal{A}_{i}\gets
		\operatorname{SE}_{i}
		(\operatorname{CA}_{i}(\mathcal{P}_{\mathrm{e}}))$
		\EndFor
		\State $\mathcal{A}_{\max}\gets
		\operatorname*{Max}_{i=1,\ldots,N_b}(\mathcal{A}_{i})$
		\Comment{element-wise}
		\State $\mathcal{U}_{\mathrm{a}}\gets
		\mathcal{P}_{\mathrm{e}}\odot\mathcal{A}_{\max}$
		
		\State $\mathcal{Y}_{\mathrm{ASAF}}\gets
		\Phi_{1\times1}^{2C_2\rightarrow C_2}
		\left(
		\operatorname{Cat}_{c}
		(\mathcal{U}_{\mathrm{d}},\mathcal{U}_{\mathrm{a}})
		\right)$
		\State \Return $\mathcal{Y}_{\mathrm{ASAF}}$
	\end{algorithmic}
\end{algorithm} 

Here $C_1$ is the concatenated input channel count, $C_2=C_1/2$, and $\operatorname{Cat}_c$ concatenates along channels. With $g=C_2/4$, $\mathcal{R}_{\ell}$ produces $g$ new channels using $1\times1$ and $3\times3$ convolutions with $4g$ and $g$ outputs, respectively, each preceded by batch normalization and ReLU. In the ELA path, Conv1D is depthwise with kernel size 5; GN uses 32 groups, and convolution and normalization parameters are shared across axes. $\operatorname{CA}_i$ and $\operatorname{SE}_i$ denote coordinate and squeeze-and-excitation attention, with separate parameters for each candidate $i$. Their outputs $\mathcal{A}_i$ are recalibrated feature tensors; Max selects the largest value across candidates at each tensor entry. CBAM applies channel attention followed by spatial attention.

\subsection{Loss Function}
\label{sec_3_6}

The standard YOLOv9 detection objective is retained for the main and training-only auxiliary heads \cite{wang2024yolov9}. In addition, a feature-distribution regularizer is applied to the CrossCEA outputs at each detection scale. Let the paired features at scale $s$ be represented as
$T^{(s)}=[Z_{\mathrm{I}}^{(s)}\mid Z_{\mathrm{V}}^{(s)}]$, where the two channel partitions correspond to the infrared and visible streams. After vectorization, a distribution is formed for each modality as
\begin{equation}
	z_{\mathrm{M}}^{(s)}
	=
	\operatorname{Softmax}
	\left(
	\operatorname{vec}(Z_{\mathrm{M}}^{(s)})
	\right),
	\qquad
	\mathrm{M}\in\{\mathrm{I},\mathrm{V}\},
	\label{eq_loss_1}
\end{equation}
where $\operatorname{vec}$ flattens the channel and spatial dimensions of each sample, and softmax normalizes its $D_s=C_sH_sW_s$ feature elements.

Using cross-entropy \cite{shannon1948mathematical} and KL divergence \cite{kullback1997information}, the quantities implemented at scale $s$ are
\begin{equation}
	\begin{aligned}
		\mathcal{H}_{\varepsilon}^{(s)}(p,q)
		&=
		-\frac{1}{B}
		\sum_{b=1}^{B}\sum_{d=1}^{D_s}
		p_{bd}\log(q_{bd}+\varepsilon),\\
		\mathcal{D}_{\mathrm{KL},\varepsilon}^{(s)}(p\|q)
		&=
		\frac{1}{B}
		\sum_{b=1}^{B}\sum_{d=1}^{D_s}
		p_{bd}
		\left[
		\log p_{bd}-\log(q_{bd}+\varepsilon)
		\right],
	\end{aligned}
	\label{eq_loss_2}
\end{equation}
where $B$ is the batch size, $\varepsilon=10^{-10}$, and logarithms are natural. The entries $p_{bd}$ and $q_{bd}$ are probabilities for sample $b$ and feature element $d$. Below, $S$ denotes the number of regularized scales in the branch, and $\mathcal{H}(p)=-B^{-1}\sum_{b,d}p_{bd}\log p_{bd}$ denotes batch-averaged entropy. The implemented CE--KL regularizer is then
\begin{equation}
	\begin{aligned}
		\mathcal{L}_{\mathrm{CE\text{-}KL}}
		=
		\sum_{s=1}^{S}
		\Big[
		&\mathcal{H}_{\varepsilon}^{(s)}
		(z_{\mathrm{I}}^{(s)},z_{\mathrm{V}}^{(s)})
		+
		\mathcal{H}_{\varepsilon}^{(s)}
		(z_{\mathrm{V}}^{(s)},z_{\mathrm{I}}^{(s)})\\
		&-
		\mathcal{D}_{\mathrm{KL},\varepsilon}^{(s)}
		(z_{\mathrm{I}}^{(s)}\|z_{\mathrm{V}}^{(s)})
		-
		\mathcal{D}_{\mathrm{KL},\varepsilon}^{(s)}
		(z_{\mathrm{V}}^{(s)}\|z_{\mathrm{I}}^{(s)})
		\Big].
	\end{aligned}
	\label{eq_loss_3}
\end{equation}
Because
$\mathcal{H}_{\varepsilon}(p,q)
-\mathcal{D}_{\mathrm{KL},\varepsilon}(p\|q)
=\mathcal{H}(p)$,
Eq.~\ref{eq_loss_3} is equivalently written as
\begin{equation}
	\mathcal{L}_{\mathrm{CE\text{-}KL}}
	=
	\sum_{s=1}^{S}
	\left[
	\mathcal{H}(z_{\mathrm{I}}^{(s)})
	+
	\mathcal{H}(z_{\mathrm{V}}^{(s)})
	\right].
	\label{eq_loss_4}
\end{equation}
Accordingly, concentrated CrossCEA responses are encouraged independently in both modalities; direct minimization of their distributional discrepancy is not imposed.

During training, this regularizer is added to the YOLOv9 detection objective with a weight of $1.0$. The same losses are applied to the auxiliary branch with a global weight of $0.25$. Only the main branch is retained during inference.

\subsection{CFGPNet Framework Scales}
\label{sec_3_7}

\begin{figure*}[h]
	\centering
	\includegraphics[
	width=0.8\textwidth,
	trim={5mm 5mm 5mm 5mm},
	clip
	]{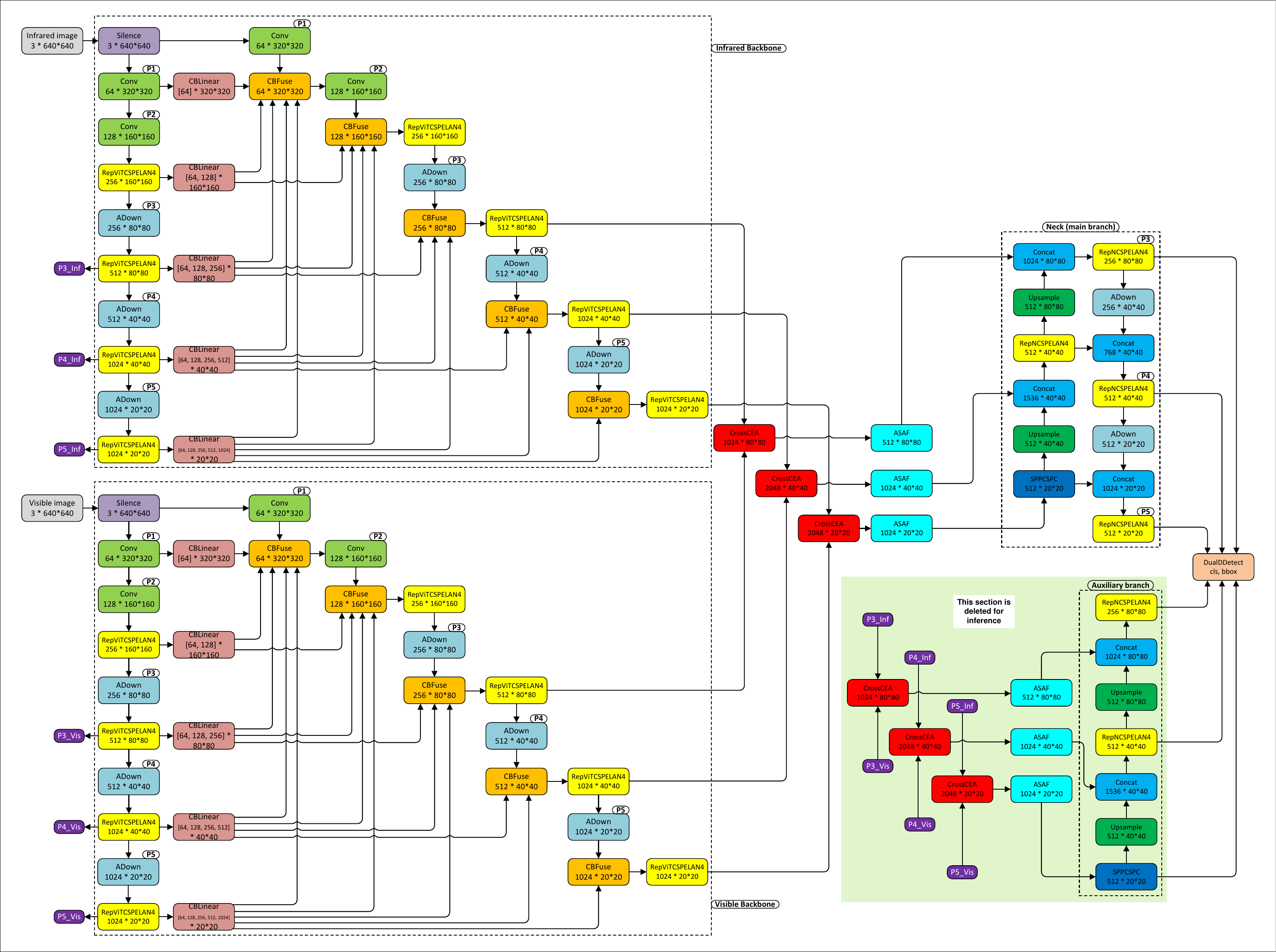}
	\captionsetup{width=0.8\textwidth}
	\caption{\centering Training architecture of CFGPNet-e with additional multilevel routing in the auxiliary pathway. Module names, output channels in brackets, and spatial resolutions are reported for a $640\times640$ input.}
	\label{fig_6}
\end{figure*}

Three variants, CFGPNet-m\footnote{medium}, CFGPNet-c\footnote{compact}, and CFGPNet-e\footnote{extended}, are instantiated for different computational budgets. The topology shown in Fig.~\ref{fig_1} is shared by CFGPNet-m and CFGPNet-c, with reduced channel widths used in CFGPNet-m. In CFGPNet-e, the channel widths and block depths are increased, and additional multilevel routing is incorporated into the training-time auxiliary pathway, as shown in Fig.~\ref{fig_6}. The CrossCEA, ASAF, and main detection stages are retained across all variants. In each case, the auxiliary pathway is removed after training, and only the main pathway is used for inference. Detailed configurations are provided in Table~\ref{tab_16}.

\section{Experiments}
\label{sec_4}

CFGPNet is evaluated on FLIR, M3FD, LLVIP, VEDAI, and MFAD. Accuracy, efficiency, ablation on MFAD, and qualitative evaluations are reported.

\subsection{Datasets}
\label{sec_4_1}

\textbf{FLIR:} The aligned subset provided by \cite{zhang2020multispectral} is used, with misregistered RGB--T pairs excluded. Person, car, and bicycle instances from daytime and nighttime scenes are evaluated. (Download link: \url{https://drive.google.com/file/d/1xHDMGl6HJZwtarNWkEV3T4O9X4ZQYz2Y/view})

\textbf{M3FD:} Registered RGB--T road scenes covering daytime, overcast, nighttime, and challenging conditions are provided. An 8:2 train--test split is used. (Download link: \url{https://github.com/JinyuanLiu-CV/TarDAL})

\textbf{LLVIP:} Registered visible--infrared pedestrian pairs captured under low-light conditions are provided. The official split is used. (Download link: \url{https://github.com/bupt-ai-cz/LLVIP})

\textbf{VEDAI:} Co-registered color and infrared aerial images containing small vehicles are provided. The $1024\times1024$ images and predefined Split~1 are used, with the split fixed \emph{a priori} for all variants. For external comparisons, reported single-split results are retained; when all splits are reported separately, Split~1 is used, whereas ten-split averages are excluded. (Download link: \url{https://downloads.greyc.fr/vedai/})

\textbf{MFAD:} Registered RGB--T driving scenes covering varied roads, illumination levels, and weather conditions are provided. The official split is used. (Download link: \url{https://github.com/hukefy/EI2Det})

Only publicly released images and annotations were used. The evaluated dataset statistics are summarized in Table~\ref{tab_1}.

\begin{table}[h]
	\centering
	\begin{threeparttable}[t]
		\footnotesize
		\caption{Statistics of the five multispectral datasets used in our experiments.}
		\label{tab_1}
		\setlength\tabcolsep{1.5pt}
		\begin{tabular}{lccccc}
			\toprule
			\textbf{Dataset} & \textbf{Resolution (VIS/IR)} & \textbf{\#Classes} & \textbf{Train} & \textbf{Test} & \textbf{Total} \\
			\midrule
			\textbf{FLIR}  &
			$640\times512$ & 3 & 4,129 & 1,013 & 5,142 \\
			\midrule
			\textbf{M3FD}  &
			\begin{tabular}[c]{@{}c@{}}$1024\times768$\tnote{\textdaggerdbl}\\$640\times512$\end{tabular} &
			6 & 3,360 & 840 & 4,200 \\
			\midrule
			\textbf{LLVIP} &
			$1080\times720$ & 1 & 12,024 & 3,464 & 15,488 \\
			\midrule
			\textbf{VEDAI} &
			$1024\times1024$ & 9 & 1,125 & 123 & 1,248 \\
			\midrule
			\textbf{MFAD}  &
			\begin{tabular}[c]{@{}c@{}}$1280\times960$\tnote{\textdaggerdbl}\\$640\times512$\end{tabular} &
			6 & 9,897 & 2,473 & 12,370 \\
			\bottomrule
		\end{tabular}
		\begin{tablenotes}[flushleft]
			\footnotesize
			\item[\textdaggerdbl] This image size is used for both modalities.
		\end{tablenotes}
	\end{threeparttable}
	\vspace{0pt}
\end{table}

Collectively, the benchmarks cover low-light pedestrian, road, adverse-weather, and small-object detection in aerial imagery.

\subsection{Evaluation Metrics}
\label{sec_4_2}

Detection accuracy is evaluated using Precision, Recall, mAP50, and mAP50:95. For a predicted box $B_p$ and ground-truth box $B_{gt}$, intersection over union is defined as
\begin{equation}
	\operatorname{IoU}\left(B_p,B_{gt}\right)
	=
	\frac{\left|B_p\cap B_{gt}\right|}
	{\left|B_p\cup B_{gt}\right|}.
	\label{eq_45}
\end{equation}
At threshold $\tauup$, a correctly classified prediction with
$\operatorname{IoU}\geq\tauup$ is counted as a true positive (TP); unmatched predictions and ground-truth instances are counted as false positives (FP) and false negatives (FN), respectively. Precision and Recall are computed as
\begin{equation}
	\operatorname{Precision}
	=
	\frac{TP}{TP+FP},
	\qquad
	\operatorname{Recall}
	=
	\frac{TP}{TP+FN}.
	\label{eq_46}
\end{equation}
For $C$ classes, mean average precision at threshold $\tauup$ is
\begin{equation}
	\operatorname{mAP}[\tauup]
	=
	\frac{1}{C}
	\sum_{c=1}^{C}
	\operatorname{AP}_{c}[\tauup],
	\label{eq_47}
\end{equation}
where
\begin{equation}
	\operatorname{AP}_{c}[\tauup]
	=
	\int_{0}^{1}
	P_c(R;\tauup)\,dR
	\label{eq_48}
\end{equation}
is obtained from the precision--recall curve by varying the confidence threshold. mAP50 is evaluated at $\tauup=0.50$, whereas mAP50:95 is averaged over ten thresholds from 0.50 to 0.95 in increments of 0.05.

Computational cost is characterized by parameter count, GFLOPs, and inference throughput. Parameters and GFLOPs are computed from the inference graph after removal of the training-only auxiliary pathway. For each dataset, all $N_{\mathrm{test}}$ paired test samples listed in Table~\ref{tab_1} are processed with a batch size of 1. If $t_i$ is the forward-pass latency in seconds for the $i$-th RGB--T pair, excluding data loading and preprocessing, throughput is computed as
\begin{equation}
	\overline{t}
	=
	\frac{1}{N_{\mathrm{test}}}
	\sum_{i=1}^{N_{\mathrm{test}}}t_i,
	\qquad
	\operatorname{FPS}
	=
	\frac{1}{\overline{t}}
	=
	\frac{N_{\mathrm{test}}}
	{\sum_{i=1}^{N_{\mathrm{test}}}t_i}.
	\label{eq_49}
\end{equation}
One paired RGB--T sample is treated as one inference instance. 

\subsection{Training Setup}
\label{sec_4_3}

The models were implemented in PyTorch 1.12.1 with Python 3.10, CUDA 11.8, and Ubuntu 20.04. Training was performed from scratch on a workstation with an Intel Core i7-12700 CPU, 96\,GB RAM, and two NVIDIA RTX 3090 GPUs. Inference was measured on one RTX 3090 using native PyTorch, without deployment acceleration, quantization, or pruning.

All variants were trained for 600 epochs at $640\times640$ resolution. Global batch sizes of 18, 10, and 6 were used for CFGPNet-m, CFGPNet-c, and CFGPNet-e, respectively. The optimization and augmentation settings are reported in Table~\ref{tab_15}. Geometric augmentations were synchronized between paired visible and thermal images to preserve their spatial correspondence.

Dataset-specific NMS thresholds were selected using a held-out subset containing 10\% of each training set, without access to the evaluation images, and were fixed across all variants. The selected values are reported in Table~\ref{tab_2}.

\begin{table}[h] 
	\centering 
	\begin{threeparttable}[t] 
		\footnotesize 
		\caption{Dataset-level NMS IoU thresholds used for CFGPNet inference.} 
		\label{tab_2} 
		\setlength\tabcolsep{1.5pt} 
		\begin{tabular}{lccccc} 
			\toprule 
			\textbf{Dataset} & \textbf{FLIR} & \textbf{M3FD} & \textbf{LLVIP} & \textbf{VEDAI} & \textbf{MFAD} \\ 
			\midrule 
			\textbf{NMS IoU threshold} & 0.6 & 0.5 & 0.65 & 0.45 & 0.55 \\ 
			\bottomrule 
		\end{tabular} 
		
	\end{threeparttable} 
	\vspace{0pt} 
\end{table}

\subsection{Comparison with State-of-the-Art Methods}
\label{sec_4_4}

CFGPNet is compared with published RGB--T detectors on FLIR, M3FD, LLVIP, VEDAI, and MFAD using the metrics defined in Section~\ref{sec_4_2}. Results for the compared methods are taken from their respective publications, whereas all CFGPNet variants are evaluated under the protocol described in Section~\ref{sec_4_3}. Dataset-specific accuracy is reported in the following subsections, and the computational profiles of the three variants are summarized in Table~\ref{tab_3}.

\begin{table}[!h] 
	\centering 
	\begin{threeparttable}[t] 
		\footnotesize
		
		\caption{Framework specifications of CFGPNet model scales 
			(input size: 640$\times$640). These specifications are obtained after deleting the auxiliary reversible branch from all models. Specifications of original trained weights are given in parentheses.} 
		\label{tab_3} 
		
		\setlength{\tabcolsep}{2.5pt} 
		
		\begin{tabularx}{\columnwidth}{ 
				@{} 
				>{\raggedright\arraybackslash}p{0.25\columnwidth} 
				*{3}{>{\centering\arraybackslash}X} 
				@{} 
			} 
			\toprule 
			& \textbf{CFGPNet-m} 
			& \textbf{CFGPNet-c} 
			& \textbf{CFGPNet-e} \\ 
			\midrule 
			
			\textbf{\#Param. (M)} 
			& 15.3 (21.0)
			& 44.9 (71.7)
			& 128.0 (180.9) \\ 
			
			\textbf{GFLOPs (G)} 
			& 56.9 (94.6)
			& 199.7 (362.2)
			& 406.1 (560.7) \\ 
			
			\textbf{FPS} 
			& 91.7 (52.2)
			& 52.6 (32.0)
			& 25.7 (18.6) \\ 
			
			\textbf{Model Size (MB)} 
			& 30.3 (41.6)
			& 87.1 (138.8)
			& 248.2 (349.6) \\ 
			
			\textbf{\#Layers} 
			& 816 (1165)
			& 816 (1165)
			& 1388 (1806) \\ 
			
			\bottomrule 
		\end{tabularx} 
		
		\begin{tablenotes}[flushleft] 
			\footnotesize 
			\item \textit{Notes}: \#Param. and GFLOPs are reported for a 
			single forward pass at 
			1$\times$6$\times$640$\times$640. FPS is calculated as the reciprocal of the mean forward-pass latency over all paired test samples listed in Table~\ref{tab_1}, excluding data loading and preprocessing, as defined in Section~\ref{sec_4_2}. 
		\end{tablenotes} 
		
	\end{threeparttable} 
\end{table}

\begin{figure*}[!h]
	\centering
	
	\captionsetup[subfigure]{
		font=scriptsize,
		justification=centering,
		singlelinecheck=false,
		skip=1pt
	}
	
	\begin{subfigure}[t]{0.2\textwidth}%
		\vspace{0pt}
		\centering
		\includegraphics[
		width=\linewidth,
		keepaspectratio,
		trim={2mm 2mm 2mm 2mm},
		clip
		]{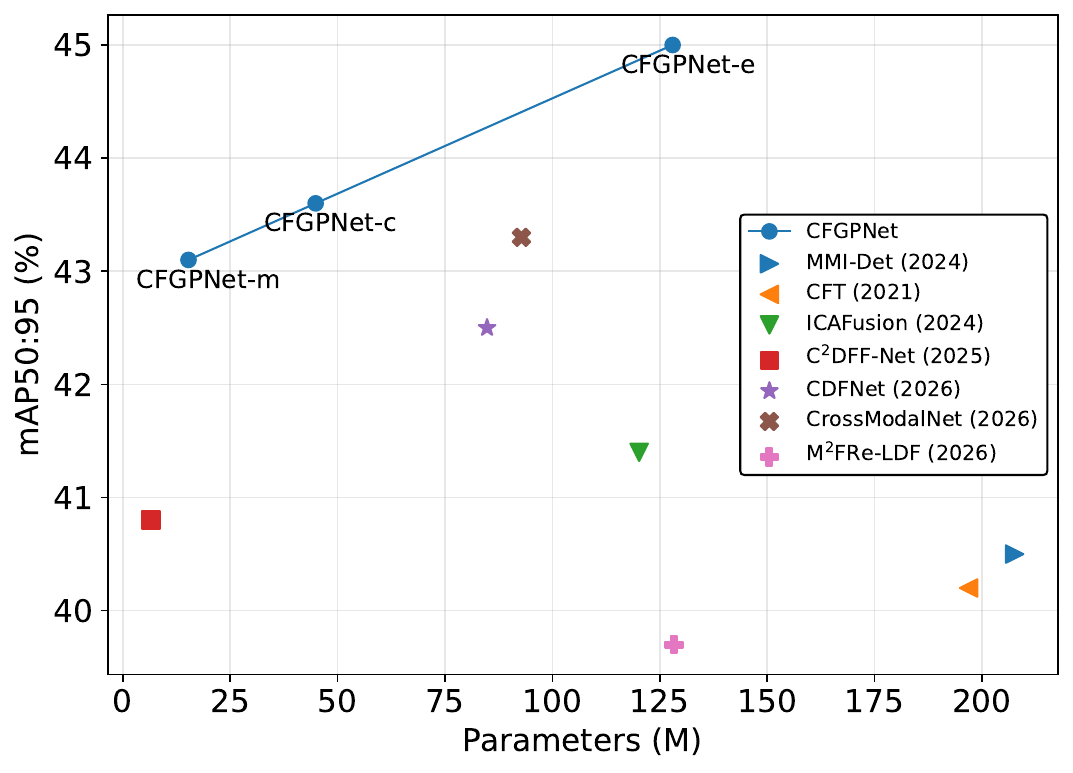}
		\caption{}
		\label{fig_7a}
	\end{subfigure}%
	\begin{subfigure}[t]{0.2\textwidth}%
		\vspace{0pt}
		\centering
		\includegraphics[
		width=\linewidth,
		keepaspectratio,
		trim={2mm 2mm 2mm 2mm},
		clip
		]{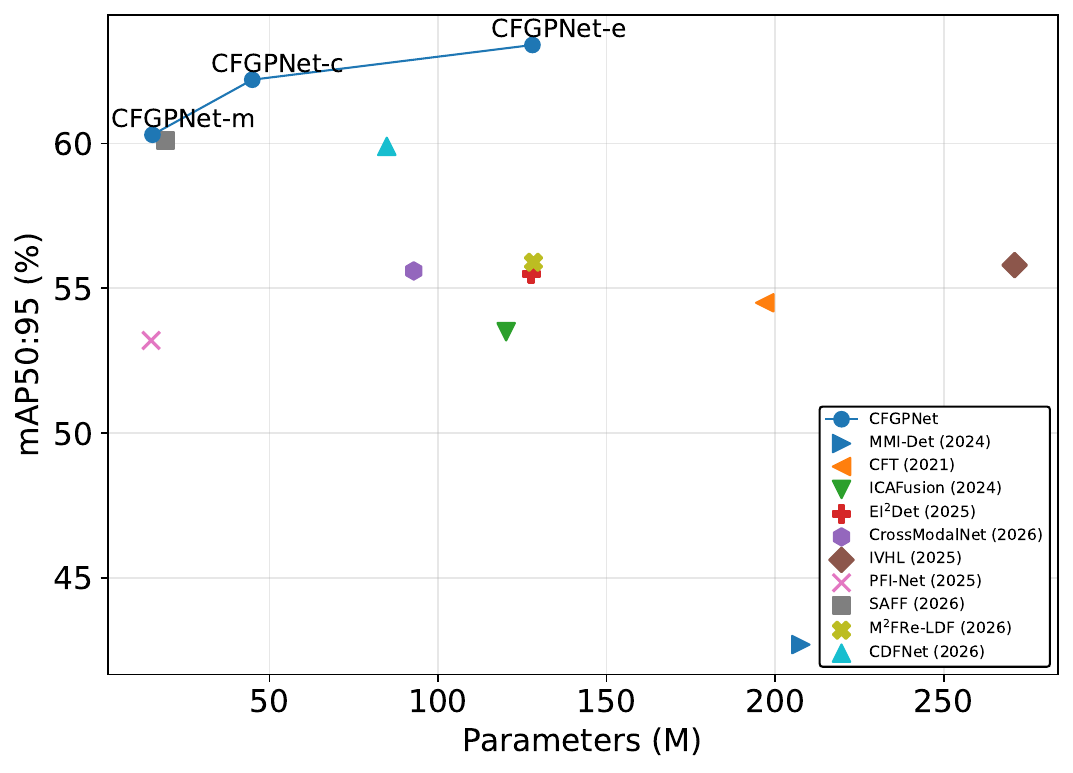}
		\caption{}
		\label{fig_7b}
	\end{subfigure}%
	\begin{subfigure}[t]{0.2\textwidth}%
		\vspace{0pt}
		\centering
		\includegraphics[
		width=\linewidth,
		keepaspectratio,
		trim={2mm 2mm 2mm 2mm},
		clip
		]{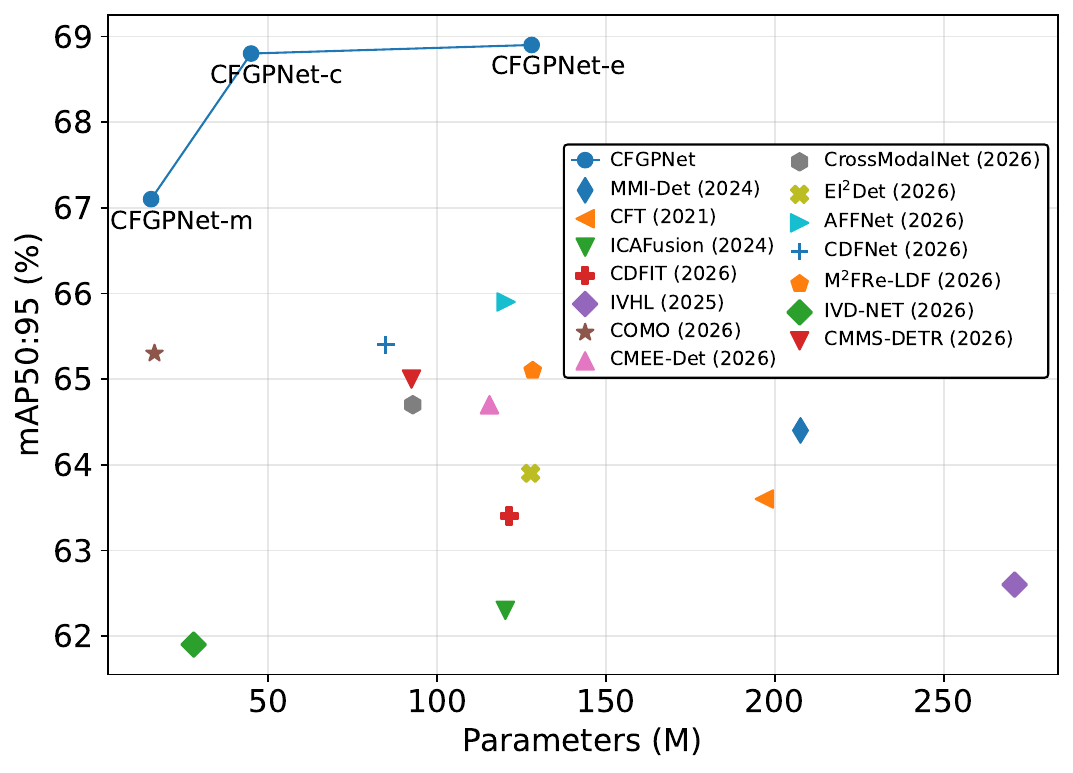}
		\caption{}
		\label{fig_7c}
	\end{subfigure}%
	\begin{subfigure}[t]{0.2\textwidth}%
		\vspace{0pt}
		\centering
		\includegraphics[
		width=\linewidth,
		keepaspectratio,
		trim={2mm 2mm 2mm 2mm},
		clip
		]{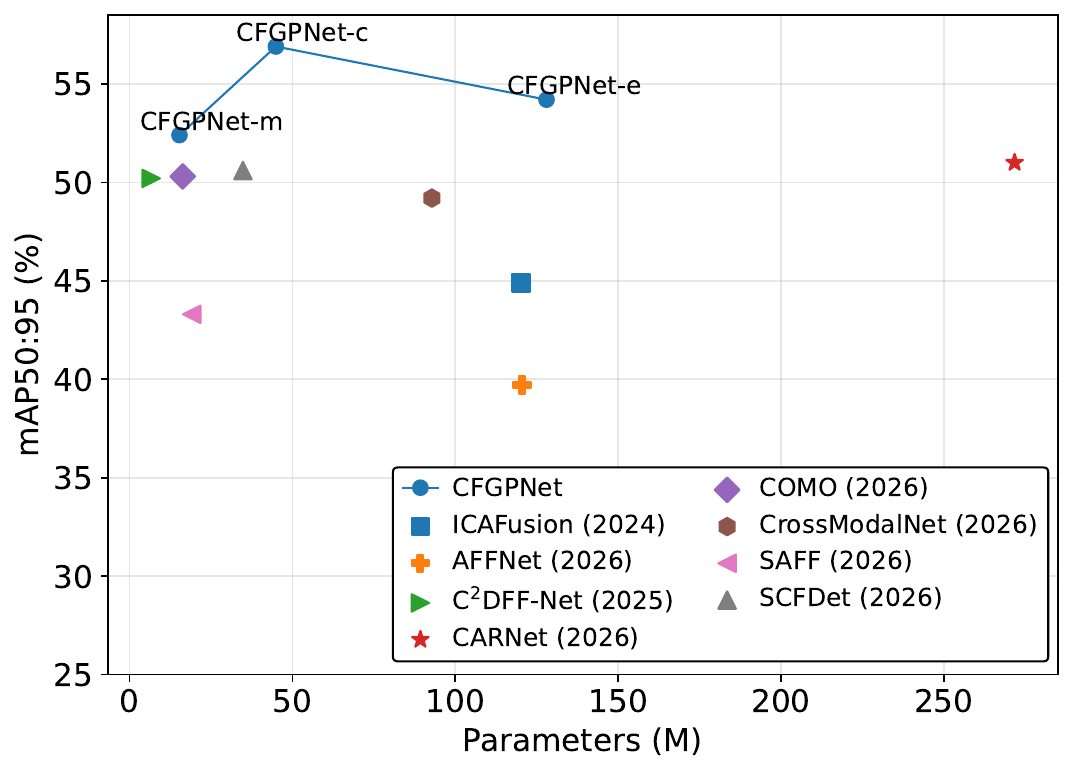}
		\caption{}
		\label{fig_7d}
	\end{subfigure}%
	\begin{subfigure}[t]{0.2\textwidth}%
		\vspace{0pt}
		\centering
		\includegraphics[
		width=\linewidth,
		keepaspectratio,
		trim={2mm 2mm 2mm 2mm},
		clip
		]{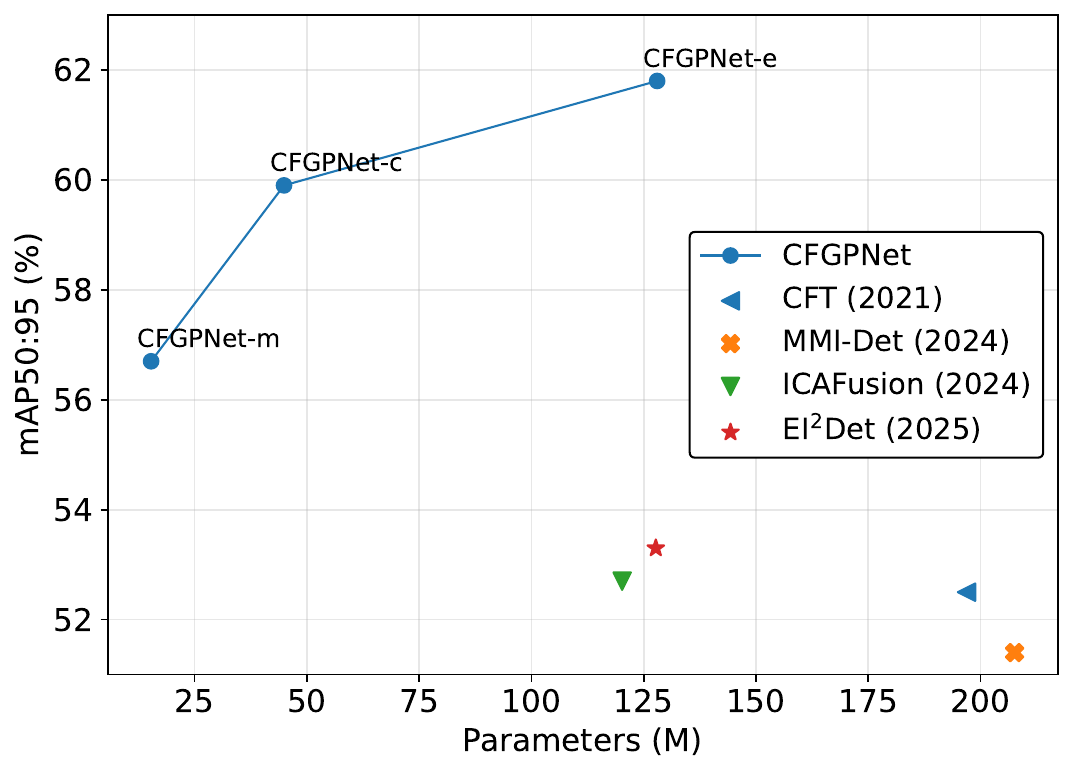}
		\caption{}
		\label{fig_7e}
	\end{subfigure}%
	
	\par\vspace{1mm}
	
	\begin{subfigure}[t]{0.2\textwidth}%
		\vspace{0pt}
		\centering
		\includegraphics[
		width=\linewidth,
		keepaspectratio,
		trim={2mm 2mm 2mm 2mm},
		clip
		]{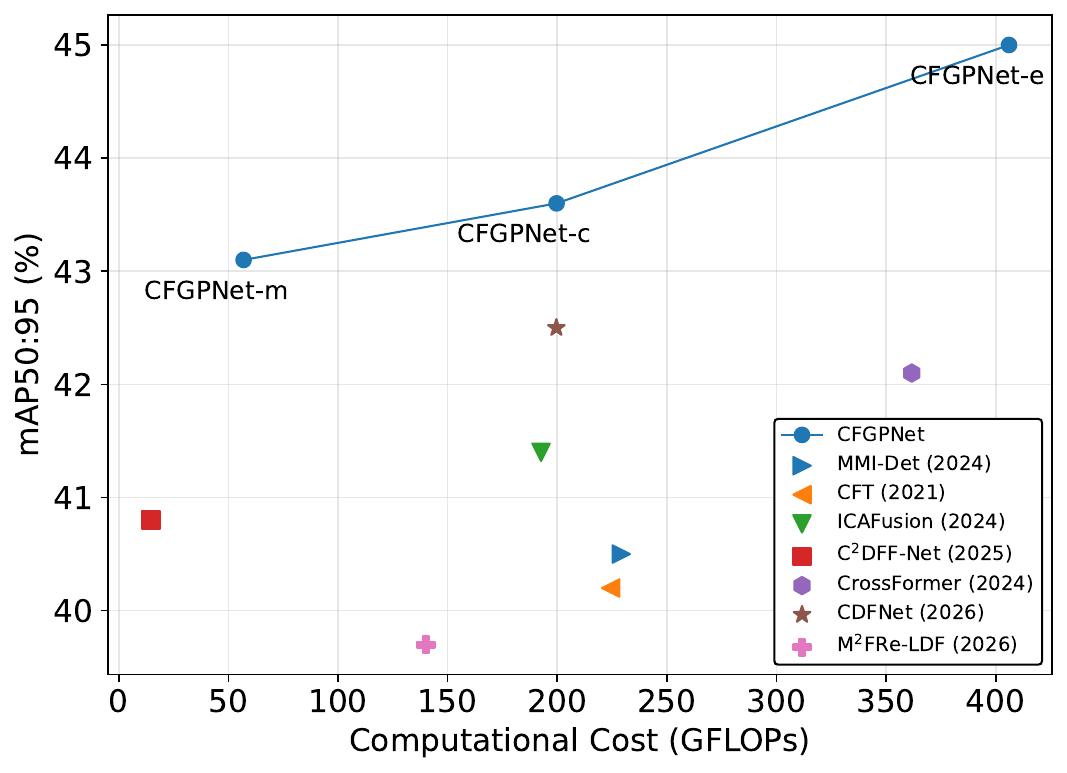}
		\caption{}
		\label{fig_7f}
	\end{subfigure}%
	\begin{subfigure}[t]{0.2\textwidth}%
		\vspace{0pt}
		\centering
		\includegraphics[
		width=\linewidth,
		keepaspectratio,
		trim={2mm 2mm 2mm 2mm},
		clip
		]{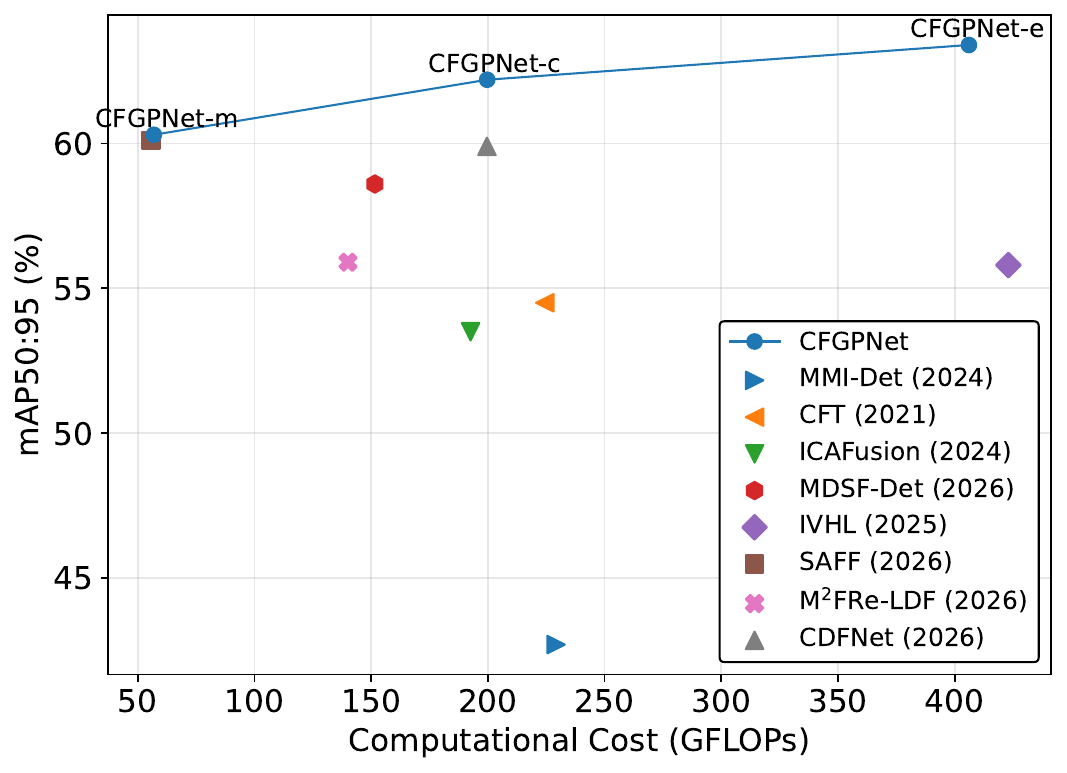}
		\caption{}
		\label{fig_7g}
	\end{subfigure}%
	\begin{subfigure}[t]{0.2\textwidth}%
		\vspace{0pt}
		\centering
		\includegraphics[
		width=\linewidth,
		keepaspectratio,
		trim={2mm 2mm 2mm 2mm},
		clip
		]{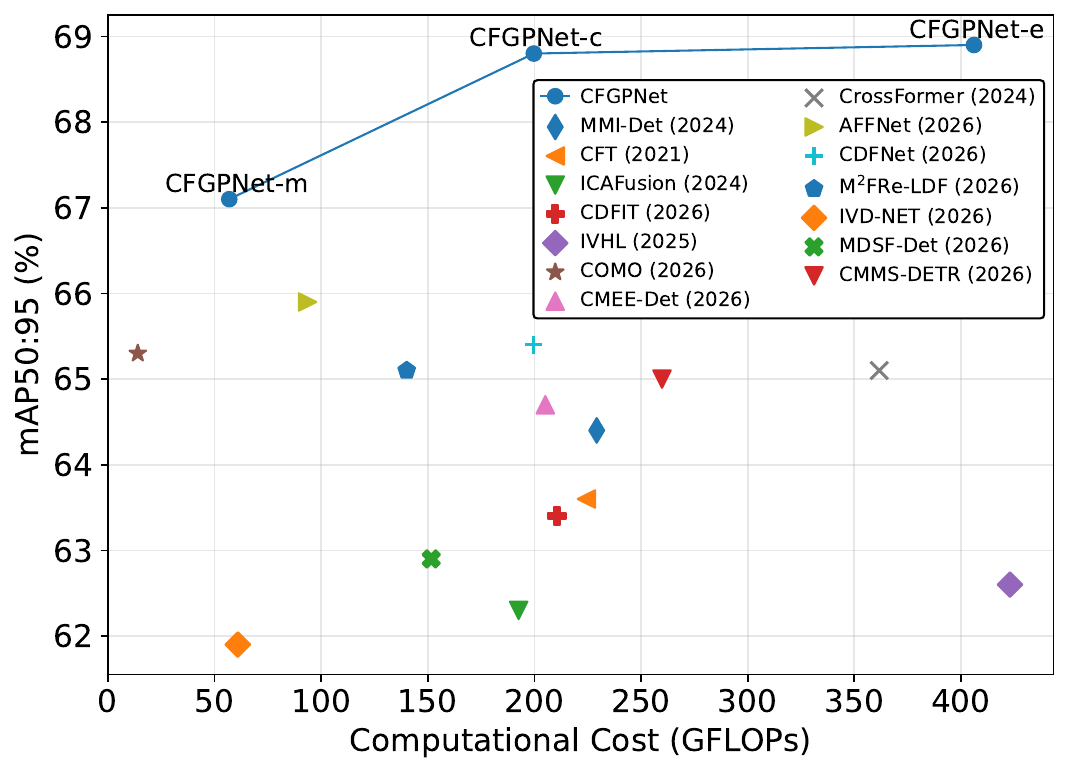}
		\caption{}
		\label{fig_7h}
	\end{subfigure}%
	\begin{subfigure}[t]{0.2\textwidth}%
		\vspace{0pt}
		\centering
		\includegraphics[
		width=\linewidth,
		keepaspectratio,
		trim={2mm 2mm 2mm 2mm},
		clip
		]{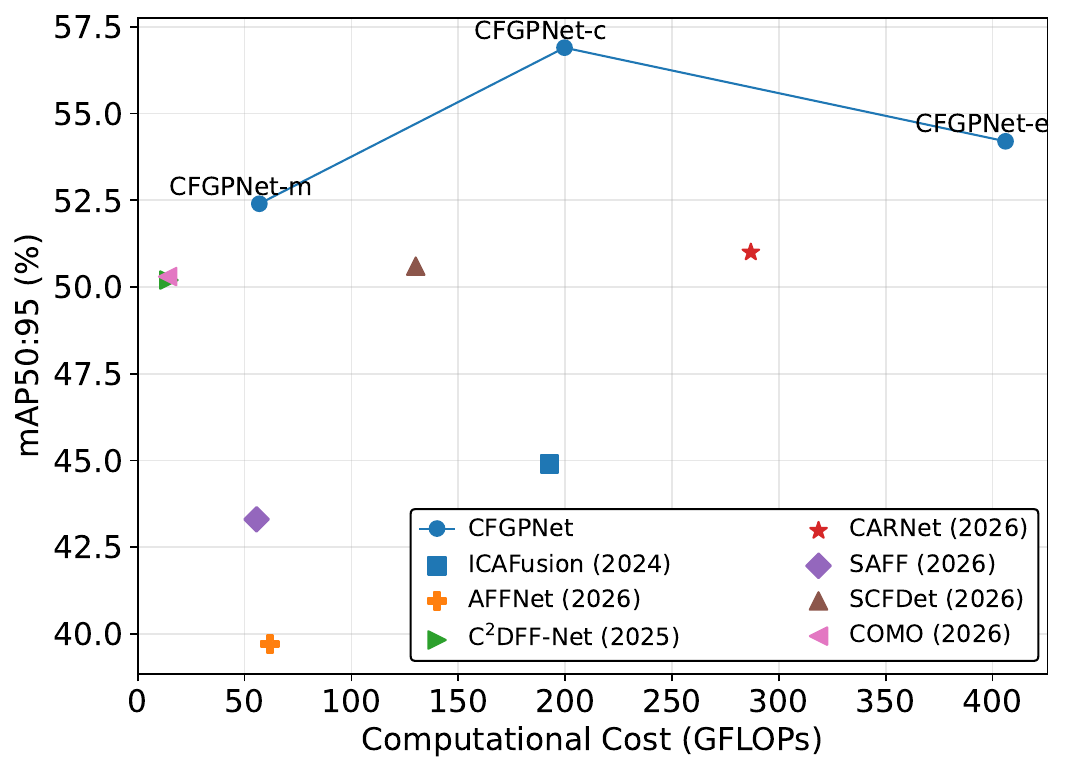}
		\caption{}
		\label{fig_7i}
	\end{subfigure}%
	\begin{subfigure}[t]{0.2\textwidth}%
		\vspace{0pt}
		\centering
		\includegraphics[
		width=\linewidth,
		keepaspectratio,
		trim={2mm 2mm 2mm 2mm},
		clip
		]{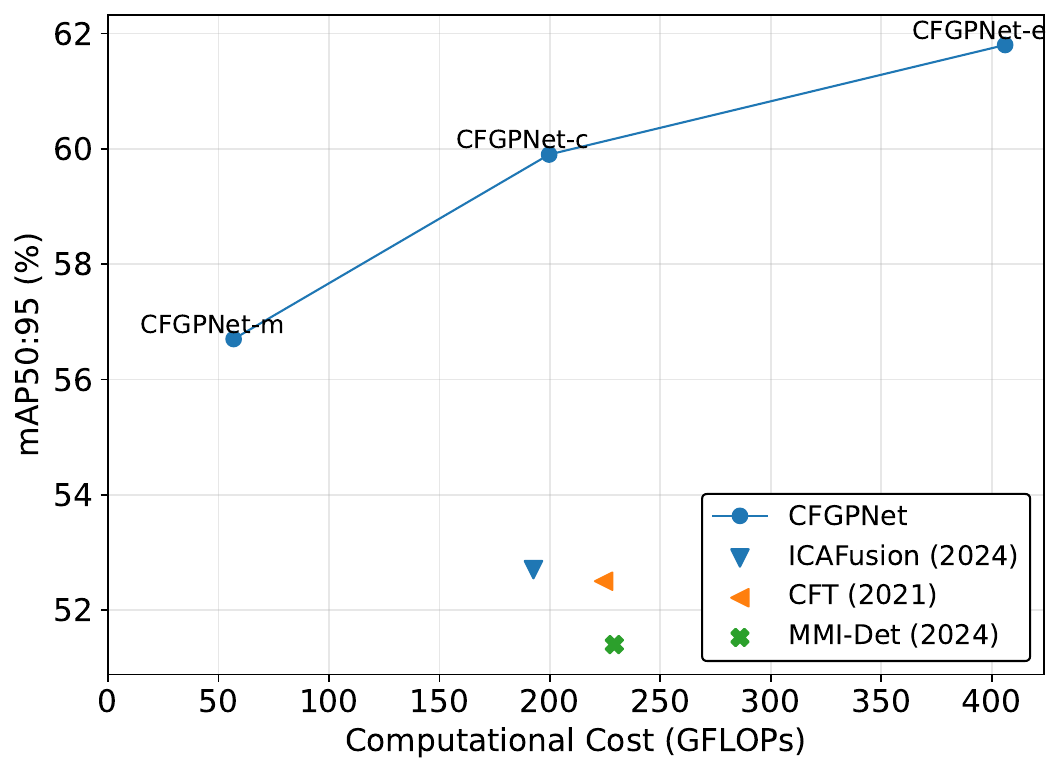}
		\caption{}
		\label{fig_7j}
	\end{subfigure}%
	
	\caption{Effectiveness--efficiency comparisons of CFGPNet and 
		state-of-the-art methods on (a,f) FLIR, (b,g) M3FD, (c,h) LLVIP,
		(d,i) VEDAI, and (e,j) MFAD. The first row shows accuracy versus
		the number of parameters, while the second row shows accuracy versus
		computational cost in GFLOPs.}
	\label{fig_7}
\end{figure*}

\textbf{FLIR.} In Table~\ref{tab_4}, the FLIR results are separated most clearly by the stricter localization metric. 
\begin{table}[!h]
	\centering
	\begin{threeparttable}[t]
		\scriptsize
		
		\caption{Quantitative comparison with state-of-the-art methods on the
			\textbf{FLIR} dataset. The best, second-best, and third-best accuracy
			results are highlighted in bold.}
		\label{tab_4}
		
		\setlength{\tabcolsep}{1.5pt}
		
		\begin{tabularx}{\columnwidth}{
				@{}
				c|
				l
				*{4}{>{\centering\arraybackslash}X}
				@{}
			}
			\toprule
			& \textbf{Model}
			& \shortstack{\textbf{\#Param.}\textbf{(M)}}
			& \shortstack{\textbf{GFLOPs}\textbf{(G)}}
			& \shortstack{\textbf{mAP}\textbf{50}}
			& \shortstack{\textbf{mAP}\textbf{50:95}} \\
			\midrule
			
			\multirow{4}{*}{\rotatebox{90}{\textbf{T}}}
			& Fast R-CNN\cite{girshick2015fast} (2015) & -- & -- & 74.2 & 38.0 \\
			& YOLOv11\cite{khanam2024yolov11} (2024) & -- & -- & 75.3 & 38.9 \\
			& MobileFormer\cite{chen2022mobile} (2022) & -- & -- & 72.9 & 36.0 \\
			& EfficientViT\cite{liu2023efficientvit} (2023) & -- & -- & 73.1 & 36.6 \\
			\midrule
			
			\multirow{4}{*}{\rotatebox{90}{\textbf{RGB}}}
			& Fast R-CNN\cite{girshick2015fast} (2015) & -- & -- & 67.6 & 30.1 \\
			& YOLOv11\cite{khanam2024yolov11} (2024) & -- & -- & 68.9 & 31.9 \\
			& MobileFormer\cite{chen2022mobile} (2022) & -- & -- & 66.9 & 30.2 \\
			& EfficientViT\cite{liu2023efficientvit} (2023) & -- & -- & 67.3 & 30.7 \\
			\midrule
			
			\multirow{17}{*}{\rotatebox{90}{\textbf{RGB--T}}}
			& MMI-Det\cite{zeng2024mmi} (2024) & 207.6 & 229.2 & 79.8 & 40.5 \\
			& CFT\cite{qingyun2021cross} (2021) & 196.9 & 224.4 & 78.7 & 40.2 \\
			& ICAFusion\cite{shen2024icafusion} (2024) & 120.2 & 192.6 & 79.2 & 41.4 \\
			& CSSA\cite{cao2023multimodal} (2023) & -- & -- & 79.2 & 41.3 \\
			& C$^2$DFF-Net\cite{11180153} (2025) & 6.6 & 14.6 & 76.9 & 40.8 \\
			& CrossFormer\cite{lee2024crossformer} (2024) & -- & 361.7 & 79.3 & 42.1 \\
			& EI$^2$Det\cite{hu2025ei} (2025) & 127.7 & -- & 80.2 & -- \\
			& CAMDet\cite{jang2025camdet} (2025) & -- & -- & 76.3 & 37.1 \\
			& DHANet\cite{wu2025dhanet} (2025) & -- & -- & 74.3 & -- \\
			& MCOR\cite{jang2025multispectral} (2025) & -- & -- & 78.2 & 39.9 \\
			& CDFNet\cite{wu2026cdfnet} (2026) & 84.8 & 199.6 & \textbf{81.9} & 42.5 \\
			& CrossModalNet\cite{li2025crossmodalnet} (2026) & 92.8 & -- & \textbf{81.7} & \textbf{43.3} \\
			& ADCA-Net\cite{zhang2026adaptive} (2026) & 34.0 & -- & 78.9 & -- \\ 
			& M$^2$FRe-LDF\cite{yu2026m2fre} (2026) & 128.3 & 140.1 & 78.3 & 39.7 \\ 
			& CFGPNet-m & 15.3 & 56.9 & 80.0 & 43.1 \\
			& CFGPNet-c & 44.9 & 199.7 & 79.8 & \textbf{43.6} \\
			& CFGPNet-e & 128.0 & 406.1 & \textbf{80.7} & \textbf{45.0} \\
			
			\bottomrule
		\end{tabularx}
		
		\begin{tablenotes}[flushleft]
			\footnotesize
			\item \textit{Notes}: \#Param. denotes the number of parameters
			in millions. ``--'' indicates that the value is not reported in the corresponding reference. Single-modality results in Tables~\ref{tab_4}-\ref{tab_7} are obtained from \cite{li2025crossmodalnet}.
		\end{tablenotes}
		
	\end{threeparttable}
	\vspace{0pt}
\end{table}
With CFGPNet-e, 80.7\% mAP50 and 45.0\% mAP50:95 are obtained; its mAP50 is within 1.2 points of the highest listed values from CDFNet (81.9\%) and CrossModalNet (81.7\%) \cite{wu2026cdfnet,li2025crossmodalnet}, while margins of 1.7 and 2.5 points over CrossModalNet and CDFNet, respectively, are recorded for mAP50:95. Values of 43.1\% and 43.6\% mAP50:95 are obtained with CFGPNet-m and CFGPNet-c. The corresponding positions in Fig.~\ref{fig_7} are distributed along a scale-dependent accuracy--cost curve, with lower-cost operating points obtained from the compact variants and the highest localization score obtained from CFGPNet-e.

\textbf{M3FD.} In Table~\ref{tab_5}, strong performance is recorded on M3FD under both evaluation criteria. 
\begin{table}[!h]
	\centering
	\begin{threeparttable}[t]
		\scriptsize
		
		\caption{Quantitative comparison with state-of-the-art methods on the
			\textbf{M3FD} dataset. The best, second-best, and third-best accuracy
			results are highlighted in bold.}
		\label{tab_5}
		
		\setlength{\tabcolsep}{1.5pt}
		
		\begin{tabularx}{\columnwidth}{
				@{}
				c|
				l
				*{4}{>{\centering\arraybackslash}X}
				@{}
			}
			\toprule
			& \textbf{Model}
			& \shortstack{\textbf{\#Param.}\textbf{(M)}}
			& \shortstack{\textbf{GFLOPs}\textbf{(G)}}
			& \shortstack{\textbf{mAP}\textbf{50}}
			& \shortstack{\textbf{mAP}\textbf{50:95}} \\
			\midrule
			
			\multirow{4}{*}{\rotatebox{90}{\textbf{T}}}
			& Fast R-CNN\cite{girshick2015fast} (2015) & -- & -- & 78.5 & 49.0 \\
			& YOLOv11\cite{khanam2024yolov11} (2024) & -- & -- & 81.2 & 50.8 \\
			& MobileFormer\cite{chen2022mobile} (2022) & -- & -- & 79.3 & 49.5 \\
			& EfficientViT\cite{liu2023efficientvit} (2023) & -- & -- & 80.7 & 50.3 \\
			
			\midrule
			
			\multirow{4}{*}{\rotatebox{90}{\textbf{RGB}}}
			& Fast R-CNN\cite{girshick2015fast} (2015) & -- & -- & 80.8 & 51.1 \\
			& YOLOv11\cite{khanam2024yolov11} (2024) & -- & -- & 83.7 & 54.3 \\
			& MobileFormer\cite{chen2022mobile} (2022) & -- & -- & 82.1 & 51.3 \\
			& EfficientViT\cite{liu2023efficientvit} (2023) & -- & -- & 84.0 & 52.7 \\
			
			\midrule
			
			\multirow{17}{*}{\rotatebox{90}{\textbf{RGB--T}}}
			& MMI-Det\cite{zeng2024mmi} (2024) & 207.6 & 229.2 & 76.6 & 42.7 \\
			& CFT\cite{qingyun2021cross} (2021) & 196.9 & 224.4 & 85.0 & 54.5 \\
			& ICAFusion\cite{shen2024icafusion} (2024) & 120.2 & 192.6 & 85.1 & 53.5 \\
			& EI$^2$Det\cite{hu2025ei} (2025) & 127.7 & -- & 86.2 & 55.5 \\
			& MCOR\cite{jang2025multispectral} (2025) & -- & -- & 87.2 & 57.3 \\
			& CrossModalNet\cite{li2025crossmodalnet} (2026) & 92.8 & -- & 87.3 & 55.6 \\
			& IVHL\cite{liu2025fusion} (2025) & 271.0 & 423.0 & 83.4 & 55.8 \\
			& CARNet\cite{wang2025carnet} (2026) & 271.7 & 286.9 & 84.3 & -- \\
			& PFI-Net\cite{wang2025pfi} (2025) & 14.9 & -- & 82.9 & 53.2 \\
			& SAFF\cite{xiao2026saff} (2026) & 19.1 & 55.6 & 87.4 & 60.1 \\
			& AlignFree-Net\cite{lu2026alignfree} (2026) & 38.5 & 46.5 & 86.3 & -- \\
			& M$^2$FRe-LDF\cite{yu2026m2fre} (2026) & 128.3 & 140.1 & 86.8 & 55.9 \\ 
			& CDFNet\cite{wu2026cdfnet} (2026) & 84.8 & 199.6 & 88.6 & 59.9 \\
			& MDSF-Det\cite{zhang2026mdsf} (2026) & -- & 151.6 & \textbf{89.3} & 58.6 \\
			& CFGPNet-m & 15.3 & 56.9 & 87.8 & \textbf{60.3} \\
			& CFGPNet-c & 44.9 & 199.7 & \textbf{89.0} & \textbf{62.2} \\
			& CFGPNet-e & 128.0 & 406.1 & \textbf{89.9} & \textbf{63.4} \\
			
			\bottomrule
		\end{tabularx}
		
		\begin{tablenotes}[flushleft]
			\footnotesize
			\item \textit{Notes}: ``--'' indicates that the value is not reported in the corresponding reference. 
		\end{tablenotes}
		
	\end{threeparttable}
	\vspace{0pt}
\end{table}
For CFGPNet-e, 89.9\% mAP50 and 63.4\% mAP50:95 are obtained, compared with 89.3\%/58.6\% for MDSF-Det and 88.6\%/59.9\% for CDFNet \cite{zhang2026mdsf,wu2026cdfnet}; margins of 0.6 and 4.8 points over MDSF-Det are thereby recorded. For the compact and medium variants, 89.0\%/62.2\% and 87.8\%/60.3\% are obtained, respectively, so competitive localization is retained below the largest configuration. The larger separation under mAP50:95 is consistent with improved localization across IoU thresholds.

\textbf{LLVIP.} In Table~\ref{tab_6}, near-saturation behavior is observed on LLVIP at mAP50: 98.9\% is reported for MMI-Det, while 97.8\%, 97.7\%, and 97.5\% are obtained with CFGPNet-m, CFGPNet-c, and CFGPNet-e, respectively \cite{zeng2024mmi}. 
\begin{table}[!h]
	\centering
	\begin{threeparttable}[t]
		\scriptsize
		
		\caption{Quantitative comparison with state-of-the-art methods on the
			\textbf{LLVIP} dataset. The best, second-best, and third-best accuracy
			results are highlighted in bold.}
		\label{tab_6}
		
		\setlength{\tabcolsep}{1.5pt}
		
		\begin{tabularx}{\columnwidth}{
				@{}
				c|
				l
				*{4}{>{\centering\arraybackslash}X}
				@{}
			}
			\toprule
			& \textbf{Model}
			& \shortstack{\textbf{\#Param.}\textbf{(M)}}
			& \shortstack{\textbf{GFLOPs}\textbf{(G)}}
			& \shortstack{\textbf{mAP}\textbf{50}}
			& \shortstack{\textbf{mAP}\textbf{50:95}} \\
			\midrule
			
			\multirow{4}{*}{\rotatebox{90}{\textbf{T}}}
			& Fast R-CNN\cite{girshick2015fast} (2015) & -- & -- & 95.4 & 61.5 \\
			& YOLOv11\cite{khanam2024yolov11} (2024) & -- & -- & 95.1 & 60.9 \\
			& MobileFormer\cite{chen2022mobile} (2022) & -- & -- & 93.2 & 59.7 \\
			& EfficientViT\cite{liu2023efficientvit} (2023) & -- & -- & 93.8 & 60.2 \\
			\midrule
			
			\multirow{4}{*}{\rotatebox{90}{\textbf{RGB}}}
			& Fast R-CNN\cite{girshick2015fast} (2015) & -- & -- & 90.1 & 52.2 \\
			& YOLOv11\cite{khanam2024yolov11} (2024) & -- & -- & 90.2 & 52.6 \\
			& MobileFormer\cite{chen2022mobile} (2022) & -- & -- & 88.6 & 50.3 \\
			& EfficientViT\cite{liu2023efficientvit} (2023) & -- & -- & 89.5 & 51.4 \\
			\midrule
			
			\multirow{30}{*}{\rotatebox{90}{\textbf{RGB--T}}}
			& MMI-Det\cite{zeng2024mmi} (2024) & 207.6 & 229.2 & \textbf{98.9} & 64.4 \\
			& CFT\cite{qingyun2021cross} (2021) & 196.9 & 224.4 & 97.5 & 63.6 \\
			& CAMDet\cite{jang2025camdet} (2025) & -- & -- & 96.5 & 62.7 \\
			& MCOR\cite{jang2025multispectral} (2025) & -- & -- & 97.6 & 64.9 \\
			& CSSA\cite{cao2023multimodal} (2023) & -- & -- & 94.3 & 59.2 \\
			& CAMF\cite{tang2023camf} (2023) & -- & -- & 89.0 & 55.6 \\
			& ICAFusion\cite{shen2024icafusion} (2024) & 120.2 & 192.6 & 96.3 & 62.3 \\
			& DHANet\cite{wu2025dhanet} (2025) & -- & -- & 97.7 & -- \\
			& CDFIT\cite{huang2026cdfit} (2026) & 121.4 & 210.6 & 97.6 & 63.4 \\
			& IVHL\cite{liu2025fusion} (2025) & 271.0 & 423.0 & 95.8 & 62.6 \\
			& CrossFormer\cite{lee2024crossformer} (2024) & -- & 361.7 & 97.4 & 65.1 \\
			& AMFD\cite{chen2025amfd} (2025) & -- & -- & 95.2 & 58.3 \\
			& COMO\cite{liu2026cross} (2026) & 16.3 & 14.0 & 97.2 & 65.3 \\
			& CMEE-Det\cite{li2025cross} (2026) & 115.5 & 205.1 & 97.0 & 64.7 \\
			& CrossModalNet\cite{li2025crossmodalnet} (2026) & 92.8 & -- & 97.7 & 64.7 \\
			& KDET-HPFL\cite{lan2025kdet} (2025) & -- & -- & -- & 65.1 \\
			& EI$^2$Det\cite{hu2025ei} (2025) & 127.7 & -- & \textbf{98.0} & 63.9 \\
			& MCFF-Det\cite{zhai2025mcff} (2025) & -- & -- & 96.0 & 64.0 \\
			& VIF-YOLO\cite{chen2025vif} (2025) & -- & -- & 96.3 & 64.5 \\
			& AFFNet\cite{tang2026adaptive} (2026) & 120.4 & 93.6 & \textbf{98.2} & 65.9 \\
			& ADCA-Net\cite{zhang2026adaptive} (2026) & 34.0 & -- & 97.6 & -- \\
			& CDFNet\cite{wu2026cdfnet} (2026) & 84.8 & 199.6 & 97.5 & 65.4 \\
			& M$^2$FRe-LDF\cite{yu2026m2fre} (2026) & 128.3 & 140.1 & 97.5 & 65.1 \\ 
			& IVD-NET\cite{wu2026ivd} (2026) & 27.9 & 60.9 & 96.7 & 61.9 \\
			& MDSF-Det\cite{zhang2026mdsf} (2026) & -- & 151.6 & 94.9 & 62.9 \\
			& TGRDet\cite{xue2026transfer} (2026) & 27.8 & 20.5 & 96.5 & -- \\
			& CMMS-DETR\cite{wang2026cmms} (2026) & 92.4 & 259.8 & 97.3 & 65.0 \\
			& CFGPNet-m & 15.3 & 56.9 & 97.8 & \textbf{67.1} \\
			& CFGPNet-c & 44.9 & 199.7 & 97.7 & \textbf{68.8} \\
			& CFGPNet-e & 128.0 & 406.1 & 97.5 & \textbf{68.9} \\
			
			\bottomrule
		\end{tabularx}
		
		\begin{tablenotes}[flushleft]
			\footnotesize
			\item \textit{Notes}: ``--'' indicates that the value is not reported in the corresponding reference. 
		\end{tablenotes}
		
	\end{threeparttable}
	\vspace{0pt}
\end{table}
The principal separation is observed at mAP50:95, where 67.1\%, 68.8\%, and 68.9\% are obtained; relative to the strongest listed prior value, 65.9\% from AFFNet, a gain of 3.0 points is recorded by CFGPNet-e \cite{tang2026adaptive}. Nearly the same localization score is obtained with CFGPNet-c using 44.9M parameters as with CFGPNet-e using 128.0M, while 67.1\% is obtained with CFGPNet-m at 15.3M parameters. The principal differences are consequently found in localization and parameter efficiency, since the headline mAP50 values are tightly grouped.

\textbf{VEDAI.} Under the fixed Split~1 protocol used for VEDAI, the highest result in Table~\ref{tab_7} is obtained with CFGPNet-c: 83.3\% mAP50 and 56.9\% mAP50:95. 
\begin{table}[!h]
	\centering
	\begin{threeparttable}[t]
		\scriptsize
		
		\caption{Quantitative comparison with state-of-the-art methods on the
			\textbf{VEDAI} dataset. Individually reported single-split results are
			included; when results for all ten splits are available separately, the
			Split~1 result is used. Studies reporting only the ten-split average are
			excluded. The best, second-best, and third-best results are highlighted
			in bold.}
		\label{tab_7}
		
		\setlength{\tabcolsep}{1.5pt}
		
		\begin{tabularx}{\columnwidth}{
				@{}
				c|
				l
				*{4}{>{\centering\arraybackslash}X}
				@{}
			}
			\toprule
			& \textbf{Model}
			& \shortstack{\textbf{\#Param.}\textbf{(M)}}
			& \shortstack{\textbf{GFLOPs}\textbf{(G)}}
			& \shortstack{\textbf{mAP}\textbf{50}}
			& \shortstack{\textbf{mAP}\textbf{50:95}} \\
			\midrule
			
			\multirow{4}{*}{\rotatebox{90}{\textbf{T}}}
			& Fast R-CNN\cite{girshick2015fast} (2015) & -- & -- & 54.2 & 38.2 \\
			& YOLOv11\cite{khanam2024yolov11} (2024) & -- & -- & 62.2 & 43.6 \\
			& MobileFormer\cite{chen2022mobile} (2022) & -- & -- & 56.6 & 37.4 \\
			& EfficientViT\cite{liu2023efficientvit} (2023) & -- & -- & 57.5 & 37.8 \\
			\midrule
			
			\multirow{4}{*}{\rotatebox{90}{\textbf{RGB}}}
			& Fast R-CNN\cite{girshick2015fast} (2015) & -- & -- & 60.8 & 41.2 \\
			& YOLOv11\cite{khanam2024yolov11} (2024) & -- & -- & 67.9 & 44.1 \\
			& MobileFormer\cite{chen2022mobile} (2022) & -- & -- & 62.1 & 42.2 \\
			& EfficientViT\cite{liu2023efficientvit} (2023) & -- & -- & 64.5 & 42.9 \\
			\midrule
			
			\multirow{15}{*}{\rotatebox{90}{\textbf{RGB--T}}}
			& CFT\cite{qingyun2021cross} (2021) & 196.9 & 224.4 & 77.2 & -- \\
			& ICAFusion\cite{shen2024icafusion} (2024) & 120.2 & 192.6 & 76.6 & 44.9 \\
			& VIF-YOLO\cite{chen2025vif} (2025) & -- & -- & 75.1 & 44.9 \\
			& DHANet\cite{wu2025dhanet} (2025) & -- & -- & 78.2 & -- \\
			& MCOR\cite{jang2025multispectral} (2025) & -- & -- & 76.2 & 46.3 \\
			& AFFNet\cite{tang2026adaptive} (2026) & 120.4 & 61.8 & 75.2 & 39.7 \\
			& C$^2$DFF-Net\cite{11180153} (2025) & 6.6 & 14.6 & 79.8 & 50.2 \\
			& CARNet\cite{wang2025carnet} (2026) & 271.7 & 286.9 & 81.4 & 51.0 \\
			& COMO\cite{liu2026cross} (2026) & 16.3 & 14.0 & \textbf{81.7} & 50.3 \\
			& CrossModalNet\cite{li2025crossmodalnet} (2026) & 92.8 & -- & 79.3 & 49.2 \\
			& SAFF\cite{xiao2026saff} (2026) & 19.1 & 55.6 & 63.3 & 43.3 \\
			& SCFDet\cite{luo2026scfdet} (2026) & 34.8 & 130.1 & 80.5 & 50.6 \\
			& CFGPNet-m & 15.3 & 56.9 & 79.8 & \textbf{52.4} \\
			& CFGPNet-c & 44.9 & 199.7 & \textbf{83.3} & \textbf{56.9} \\
			& CFGPNet-e & 128.0 & 406.1 & \textbf{82.9} & \textbf{54.2} \\
			
			\bottomrule
		\end{tabularx}
		
		\begin{tablenotes}[flushleft]
			\footnotesize
			\item \textit{Notes}: ``--'' indicates that the value is not reported in the
			corresponding reference.
		\end{tablenotes}
		
	\end{threeparttable}
	\vspace{0pt}
\end{table}
Against the closest high-scoring entries, gains of 1.6 mAP50 points over COMO (81.7\%) and 5.9 mAP50:95 points over CARNet (51.0\%) are recorded \cite{liu2026cross,wang2025carnet}; 82.9\%/54.2\% and 79.8\%/52.4\% are obtained with CFGPNet-e and CFGPNet-m, respectively. The absence of a monotonic gain with model size is explicit: higher values on both metrics are obtained with the 44.9M-parameter CFGPNet-c than with the 128.0M-parameter CFGPNet-e. This pattern is consistent with a dataset-specific capacity requirement for small objects in aerial imagery.

\textbf{MFAD.} In Table~\ref{tab_8}, a clear scale response is observed on MFAD. 
\begin{table}[!h] 
	\centering 
	\begin{threeparttable}[t] 
		\scriptsize 
		
		\caption{Quantitative comparison with state-of-the-art methods on the 
			\textbf{MFAD} dataset. The best, second-best, and third-best results are highlighted in bold.} 
		\label{tab_8} 
		
		\setlength{\tabcolsep}{1.5pt} 
		
		\begin{tabularx}{\columnwidth}{ 
				@{} 
				c| 
				l 
				*{4}{>{\centering\arraybackslash}X} 
				@{} 
			} 
			\toprule 
			& \textbf{Model} 
			& \shortstack{\textbf{\#Param.}\textbf{(M)}} 
			& \shortstack{\textbf{GFLOPs}\textbf{(G)}} 
			& \shortstack{\textbf{mAP}\textbf{50}} 
			& \shortstack{\textbf{mAP}\textbf{50:95}} \\ 
			\midrule 
			
			\multirow{3}{*}{\rotatebox{90}{\textbf{T}}} 
			& YOLOv5-l\cite{jocher2022ultralyticsv7} (2022) & -- & -- & 70.0 & 42.8 \\ 
			& YOLOv7-x\cite{wang2023yolov7} (2022) & -- & -- & 66.5 & 40.6 \\ 
			& YOLOv10-l\cite{wang2024yolov10} (2024) & -- & -- & 65.7 & 41.8 \\ 
			\midrule 
			
			\multirow{3}{*}{\rotatebox{90}{\textbf{RGB}}} 
			& YOLOv5-l\cite{jocher2022ultralyticsv7} (2022) & -- & -- & 74.9 & 49.1 \\ 
			& YOLOv7-x\cite{wang2023yolov7} (2022) & -- & -- & 72.9 & 48.8 \\ 
			& YOLOv10-l\cite{wang2024yolov10} (2024) & -- & -- & 71.1 & 48.9 \\ 
			\midrule 
			
			\multirow{7}{*}{\rotatebox{90}{\textbf{RGB--T}}} 
			& CFT\cite{qingyun2021cross} (2021) & 196.9 & 224.4 & 77.8 & 52.5 \\ 
			& MMI-Det\cite{zeng2024mmi} (2024) & 207.6 & 229.2 & 76.9 & 51.4 \\ 
			& ICAFusion\cite{shen2024icafusion} (2024) & 120.2 & 192.6 & 77.6 & 52.7 \\ 
			& EI$^2$Det\cite{hu2025ei} (2025) & 127.7 & -- & 79.0 & 53.3 \\ 
			& CFGPNet-m & 15.3 & 56.9 & \textbf{79.6} & \textbf{56.7} \\ 
			& CFGPNet-c & 44.9 & 199.7 & \textbf{82.0} & \textbf{59.9} \\ 
			& CFGPNet-e & 128.0 & 406.1 & \textbf{83.4} & \textbf{61.8} \\ 
			
			\bottomrule 
		\end{tabularx} 
		
		\begin{tablenotes}[flushleft] 
			\footnotesize 
			\item \textit{Notes}: ``--'' indicates that the value is not reported in the 
			corresponding reference. Single-modality results are obtained from \cite{hu2025ei}.
		\end{tablenotes} 
		
	\end{threeparttable} 
	\vspace{0pt} 
\end{table}
The pairs 79.6\%/56.7\%, 82.0\%/59.9\%, and 83.4\%/61.8\% are obtained with CFGPNet-m, CFGPNet-c, and CFGPNet-e, respectively, and the highest values on both metrics are therefore obtained with CFGPNet-e. Relative to the strongest prior entry listed, EI$^2$Det at 79.0\%/53.3\% \cite{hu2025ei}, gains of 4.4 mAP50 points and 8.5 mAP50:95 points are recorded by CFGPNet-e; gains of 3.0 and 6.6 points are still obtained with CFGPNet-c using 44.9M parameters. Across the five benchmarks, the deployment choices are made explicit by the parameter-- and GFLOP--accuracy plots in Fig.~\ref{fig_7}: 91.7 FPS at 56.9 GFLOPs is obtained with CFGPNet-m, 52.6 FPS at 199.7 GFLOPs with CFGPNet-c, and 25.7 FPS at 406.1 GFLOPs with CFGPNet-e.

\begin{figure}[t]
	\centering
	\includegraphics[width=0.5\textwidth, trim={6mm 6mm 6mm 6mm}, clip]{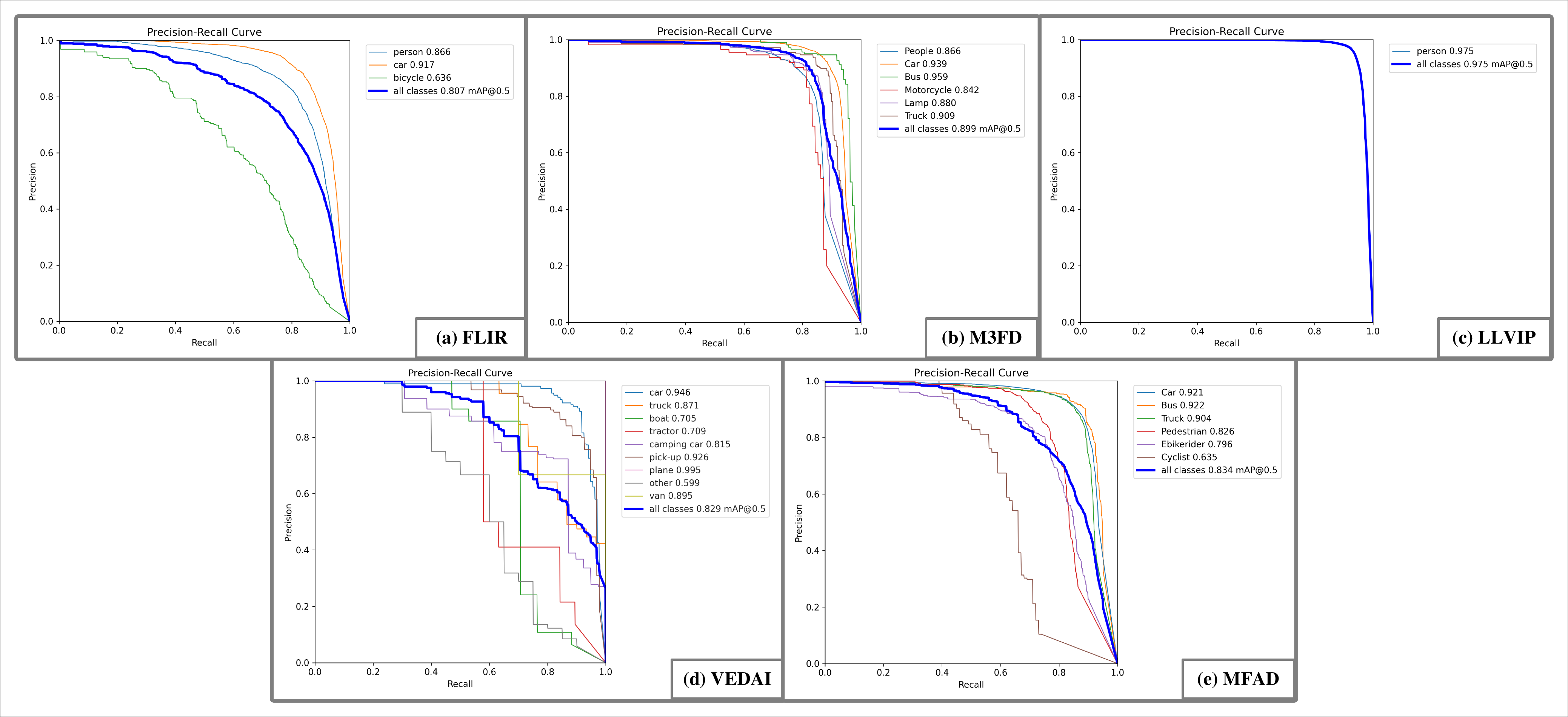}
	\caption{\centering Illustration of precision--recall curves for CFGPNet-e on (a) FLIR, (b) M3FD, (c) LLVIP, (d) VEDAI, and (e) MFAD datasets.}\label{fig_12}
\end{figure}

Cross-dataset precision--recall behavior for CFGPNet-e is summarized in Fig.~\ref{fig_12}; high precision is retained over the recall range on all five benchmarks, with aggregate mAP50 values spanning 80.7\% on FLIR to 97.5\% on LLVIP. 
The class-wise values in Table~\ref{tab_pr} provide the corresponding detail: high per-class mAP50 is obtained for M3FD Bus (95.9\%), VEDAI Plane (99.5\%), and MFAD Car and Bus (92.1\% and 92.2\%), whereas lower values are observed for FLIR Bicycle (63.6\%), VEDAI ``Other'' (59.9\%), and MFAD Cyclist (63.5\%). 
\begin{table}[!t]
	\centering
	\footnotesize
	\caption{Detailed per-class and overall detection results of CFGPNet variants on all evaluated datasets. Values are reported in the format Precision/Recall/mAP50/mAP50:95 (\%).}
	\label{tab_pr}
	
	\begin{threeparttable}[t]
		\setlength{\tabcolsep}{2.0pt}
		\renewcommand{\arraystretch}{0.90}
		
		\resizebox{\columnwidth}{!}{%
			\begin{tabular}{l|lccc}
				\toprule
				& \textbf{Class}
				& \textbf{CFGPNet-m}
				& \textbf{CFGPNet-c}
				& \textbf{CFGPNet-e} \\
				\midrule
				
				\multirow{4}{*}{\textbf{FLIR}}
				& \textbf{person}
				& 83.2/78.4/85.9/44.5
				& 82.3/79.8/86.0/45.1
				& 83.5/78.9/86.6/46.1 \\
				
				& \textbf{car}
				& 81.4/87.1/91.2/61.8
				& 80.8/87.9/91.5/62.6
				& 83.0/86.8/91.7/63.5 \\
				
				& \textbf{bicycle}
				& 73.1/54.3/62.8/23.1
				& 67.3/54.3/61.9/23.2
				& 68.4/54.9/63.6/25.3 \\
				
				\cline{2-5}
				& \textbf{all}
				& 79.2/73.3/80.0/43.1
				& 76.8/74.0/79.8/43.6
				& 78.3/73.5/80.7/45.0 \\
				
				\midrule
				
				\multirow{7}{*}{\textbf{M3FD}}
				& \textbf{People}
				& 86.0/78.1/84.7/53.5
				& 87.6/78.6/86.1/55.6
				& 86.9/80.9/86.6/56.6 \\
				
				& \textbf{Car}
				& 90.4/88.1/92.9/70.4
				& 91.1/87.9/93.1/71.2
				& 90.8/89.0/93.9/72.1 \\
				
				& \textbf{Bus}
				& 91.7/92.5/95.5/80.3
				& 91.9/93.0/95.5/82.4
				& 91.7/92.5/95.9/82.2 \\
				
				& \textbf{Motorcycle}
				& 87.5/71.6/81.0/46.0
				& 88.0/79.4/84.9/48.4
				& 89.5/80.4/84.2/49.1 \\
				
				& \textbf{Lamp}
				& 87.1/77.2/83.7/46.5
				& 89.2/78.0/85.0/48.9
				& 90.3/81.1/88.0/51.7 \\
				
				& \textbf{Truck}
				& 88.9/84.4/88.9/65.2
				& 89.7/87.2/89.5/66.6
				& 89.8/88.4/90.9/68.8 \\
				
				\cline{2-5}
				& \textbf{all}
				& 88.6/82.0/87.8/60.3
				& 89.6/84.0/89.0/62.2
				& 89.9/85.4/89.9/63.4 \\
				
				\midrule
				
				\multirow{2}{*}{\textbf{LLVIP}}
				& \textbf{person}
				& 96.2/93.8/97.8/67.1
				& 96.2/93.1/97.7/68.8
				& 96.8/92.5/97.5/68.9 \\
				
				\cline{2-5}
				& \textbf{all}
				& 96.2/93.8/97.8/67.1
				& 96.2/93.1/97.7/68.8
				& 96.8/92.5/97.5/68.9 \\
				
				\midrule
				
				\multirow{10}{*}{\textbf{VEDAI}}
				& \textbf{car}
				& 88.4/73.9/88.8/58.4
				& 90.1/81.2/91.5/59.6
				& 92.1/86.7/94.6/60.7 \\
				
				& \textbf{truck}
				& 77.5/63.3/78.2/52.0
				& 78.2/59.9/78.0/49.3
				& 68.3/76.7/87.1/62.4 \\
				
				& \textbf{boat}
				& 72.9/41.2/66.4/38.9
				& 74.2/58.8/69.6/41.0
				& 89.6/50.9/70.5/45.7 \\
				
				& \textbf{tractor}
				& 92.6/65.7/76.7/46.6
				& 83.4/63.2/84.0/53.3
				& 100.0/56.6/70.9/45.3 \\
				
				& \textbf{camping car}
				& 60.9/76.9/73.7/53.2
				& 67.1/78.6/78.2/54.4
				& 71.6/84.2/81.5/55.8 \\
				
				& \textbf{pick-up}
				& 81.3/71.6/83.8/54.2
				& 85.2/83.2/87.5/54.6
				& 86.0/84.3/92.6/60.9 \\
				
				& \textbf{plane}
				& 65.6/100.0/99.5/79.6
				& 71.2/100.0/99.5/99.5
				& 43.7/100.0/99.5/59.7 \\
				
				& \textbf{other}
				& 87.4/34.7/56.7/29.1
				& 84.3/53.8/70.8/36.2
				& 69.2/45.0/59.9/32.2 \\
				
				& \textbf{van}
				& 100.0/85.9/94.3/59.2
				& 87.7/70.0/90.9/64.0
				& 90.7/70.0/89.5/65.0 \\
				
				\cline{2-5}
				& \textbf{all}
				& 80.7/68.1/79.8/52.4
				& 80.2/72.1/83.3/56.9
				& 79.0/72.7/82.9/54.2 \\
				
				\midrule
				
				\multirow{7}{*}{\textbf{MFAD}}
				& \textbf{Car}
				& 88.4/85.3/90.8/69.9
				& 90.0/85.1/91.4/71.6
				& 90.0/86.1/92.1/72.8 \\
				
				& \textbf{Bus}
				& 87.8/87.8/90.9/71.3
				& 91.1/88.6/92.2/73.7
				& 89.9/89.1/92.2/74.4 \\
				
				& \textbf{Truck}
				& 87.9/82.5/87.0/65.6
				& 90.1/81.9/88.6/68.9
				& 91.3/84.3/90.4/71.4 \\
				
				& \textbf{Pedestrian}
				& 83.1/70.7/79.1/49.5
				& 87.5/71.1/81.3/52.5
				& 86.9/73.9/82.6/54.8 \\
				
				& \textbf{Ebikerider}
				& 78.1/72.7/76.8/47.6
				& 83.6/73.6/79.3/51.4
				& 81.6/71.8/79.6/52.7 \\
				
				& \textbf{Cyclist}
				& 69.8/46.0/52.7/36.3
				& 76.5/48.8/59.1/41.6
				& 79.9/56.0/63.5/45.0 \\
				
				\cline{2-5}
				& \textbf{all}
				& 82.5/74.2/79.6/56.7
				& 86.5/74.9/82.0/59.9
				& 86.6/76.8/83.4/61.8 \\
				
				\bottomrule
			\end{tabular}%
		}
	\end{threeparttable}
	
	\vspace{-4pt}
\end{table}
For mAP50:95, the highest aggregate values are obtained with CFGPNet-e on FLIR, M3FD, LLVIP, and MFAD, while the strongest VEDAI result is obtained with CFGPNet-c.

\subsection{Ablation Studies}
\label{sec_4_5}

All ablations were conducted on MFAD using CFGPNet-m. Except for the factor under study, the architecture, training schedule, data split, and evaluation protocol were held fixed.

\textbf{RepViT as the backbone baseline.} Table~\ref{tab_9} shows that replacing RepVGG with RepViT increased mAP50/mAP50:95 from $76.4\%/53.9\%$ to $79.6\%/56.7\%$. 
\begin{table}[h]
	\centering
	\begin{threeparttable}[t]
		\footnotesize
		\caption{Controlled ablation of RepVGG and RepViT as the GELAN backbone baseline. All other CFGPNet components and training/inference settings are kept fixed. The best result is highlighted in bold.}
		\label{tab_9}
		\setlength\tabcolsep{1.5pt}
		\begin{tabular}{lccccc}
			\toprule
			\textbf{Baseline} & \textbf{mAP50} & \textbf{mAP50:95} & \textbf{GFLOPs (G)} & \textbf{\#Param. (M)} & \textbf{FPS} \\
			\midrule
			\textbf{RepVGG} & 76.4 & 53.9 & 60.8 & 15.9 & 82.3\\
			\textbf{RepViT} & \textbf{79.6} & \textbf{56.7} & \textbf{56.9} & \textbf{15.3} & \textbf{91.7}\\
			\bottomrule
		\end{tabular}
	\end{threeparttable}
	\vspace{0pt}
\end{table}
At the same time, GFLOPs decreased from 60.8 to 56.9, parameters from 15.9M to 15.3M, and FPS increased from 82.3 to 91.7. This improvement is consistent with the separation of spatial and channel mixing and the use of structural re-parameterization in RepViT. RepViT was therefore retained as the GELAN backbone.

\textbf{Effect of PGI.} With training-time PGI supervision, mAP50 and mAP50:95 were increased from 78.9\% and 56.4\% to 79.6\% and 56.7\%, respectively (Table~\ref{tab_9_1}). 
\begin{table}[!h]
	\centering
	\begin{threeparttable}[t]
		\footnotesize
		\caption{Controlled ablation of model performance with and without PGI. All other CFGPNet components and training/inference settings are kept fixed. The best value in each column is highlighted in bold.}
		\label{tab_9_1}
		\setlength\tabcolsep{1.5pt}
		\begin{tabular}{lccccc}
			\toprule
			\textbf{Method} & \textbf{mAP50} & \textbf{mAP50:95} & \textbf{GFLOPs (G)} & \textbf{\#Param. (M)} & \textbf{FPS} \\
			\midrule
			\textbf{Without PGI} & 78.9 & 56.4 & \textbf{56.9} & \textbf{15.3} & \textbf{93.5}\\
			\textbf{With PGI} & \textbf{79.6} & \textbf{56.7} & \textbf{56.9} & \textbf{15.3} & 91.7\\
			\bottomrule
		\end{tabular}
	\end{threeparttable}
	\vspace{0pt}
\end{table}
Because the auxiliary branch is removed after training, the inference cost remained unchanged at 15.3M parameters and 56.9 GFLOPs. A modest accuracy gain was therefore obtained without increasing deployment complexity.

\textbf{Ablation on fusion loss terms.} The fusion-loss ablation is reported in Table~\ref{tab_10}. 
\begin{table}[!h]
	\centering
	\begin{threeparttable}[t]
		\footnotesize
		\caption{Ablation study on fusion loss terms. The best value in each column is highlighted in bold.}
		\label{tab_10}
		\setlength\tabcolsep{1.5pt}
		\begin{tabular}{lcc}
			\toprule
			\textbf{Loss term} & \textbf{mAP50} & \textbf{mAP50:95} \\
			\midrule
			\textbf{None} & \textbf{79.7} & 56.6\\
			\textbf{MSE} & 79.1 & 55.4\\
			\textbf{SSIM} & 73.8 & 51.1\\
			\textbf{CE-KL} & 79.6 & \textbf{56.7}\\
			\textbf{MSE + SSIM + CE-KL} & 74.3 & 51.5\\
			\bottomrule
		\end{tabular}
	\end{threeparttable}
	\vspace{0pt}
\end{table}
The CE--KL regularizer defined in Section~\ref{sec_3_6} was compared with no auxiliary fusion loss, MSE, SSIM, and their combination. The alternative terms were computed from the split CrossCEA outputs. Structural similarity (SSIM) \cite{wang2004image} was used to constrain structural consistency:
\begin{gather}
	\mathcal{L}_{\text{SSIM}} = \sum_{s=1}^{S} \left(1-\operatorname{SSIM}\left(p^{(s)}, q^{(s)}\right)\right),\notag\\ 
	p^{(s)} = \sigma\left(Z_{\text{I}}^{(s)}\right),\notag\\ 
	q^{(s)} = \sigma\left(Z_{\text{V}}^{(s)}\right),
	\label{eq_39}
\end{gather}

\begin{equation}
	\begin{aligned}
		\operatorname{SSIM}(p,q)
		&=\operatorname{Mean}\!\left[
		\frac{2\mu_p\mu_q+C_1}{\mu_p^2+\mu_q^2+C_1}
		\frac{2\sigma_{pq}+C_2}{\sigma_p^2+\sigma_q^2+C_2}
		\right],\\
		\mu_p&=\mathcal{G}*p,\\
		\sigma_p^2&=\mathcal{G}*(p^2)-\mu_p^2,\\
		\sigma_{pq}&=\mathcal{G}*(pq)-\mu_p\mu_q,\\
		C_1&=(0.01L)^2,\qquad C_2=(0.03L)^2 .
	\end{aligned}
	\label{eq_40}
\end{equation}
Here $p=p^{(s)}$ and $q=q^{(s)}$ at each scale, $\sigma(\cdot)$ denotes sigmoid, and $L=1$ is the feature range after sigmoid. $\mathcal{G}$ is a normalized $11\times11$ Gaussian kernel with standard deviation 1.5; $*$ denotes channel-wise spatial convolution without padding. Mean averages over samples, channels, and valid window positions. The statistics for $q$ are defined analogously. An MSE\footnote{mean squared error} term was used to penalize pointwise deviations after sigmoid normalization:
\begin{equation}
	\mathcal{L}_{\mathrm{MSE}}
	=\sum_{s=1}^{S}\frac{1}{BD_s}
	\sum_{b=1}^{B}\sum_{d=1}^{D_s}
	\left(p_{bd}^{(s)}-q_{bd}^{(s)}\right)^2,
	\label{eq_41}
\end{equation}
where $p^{(s)}$ and $q^{(s)}$ are the sigmoid-normalized features in Eq.~\ref{eq_39}, and $d$ indexes their flattened channel and spatial dimensions.

The main-branch coefficients were $\lambda_{CE-KL}=1.0$, $\lambda_{SSIM}=3.0$, and $\lambda_{MSE}=300.0$; each auxiliary-branch term was multiplied by $0.25$. Without fusion loss, $79.7\%/56.6\%$ was obtained for mAP50/mAP50:95, whereas CE--KL yielded $79.6\%/56.7\%$. CE--KL was therefore retained for its higher mAP50:95. MSE, SSIM, and MSE+SSIM+CE--KL yielded $79.1\%/55.4\%$, $73.8\%/51.1\%$, and $74.3\%/51.5\%$, respectively. Pointwise and structural alignment thus reduced performance, consistent with over-regularization of modality-specific cues.

\begin{figure}[!h]
	\centering
	
	\captionsetup[subfigure]{
		font=scriptsize,
		justification=centering,
		singlelinecheck=false,
		skip=2pt
	}
	
	\begin{subfigure}[t]{0.49\columnwidth}
		\vspace{0pt}
		\centering
		\includegraphics[
		width=\linewidth,
		trim={2mm 2mm 2mm 2mm},
		clip
		]{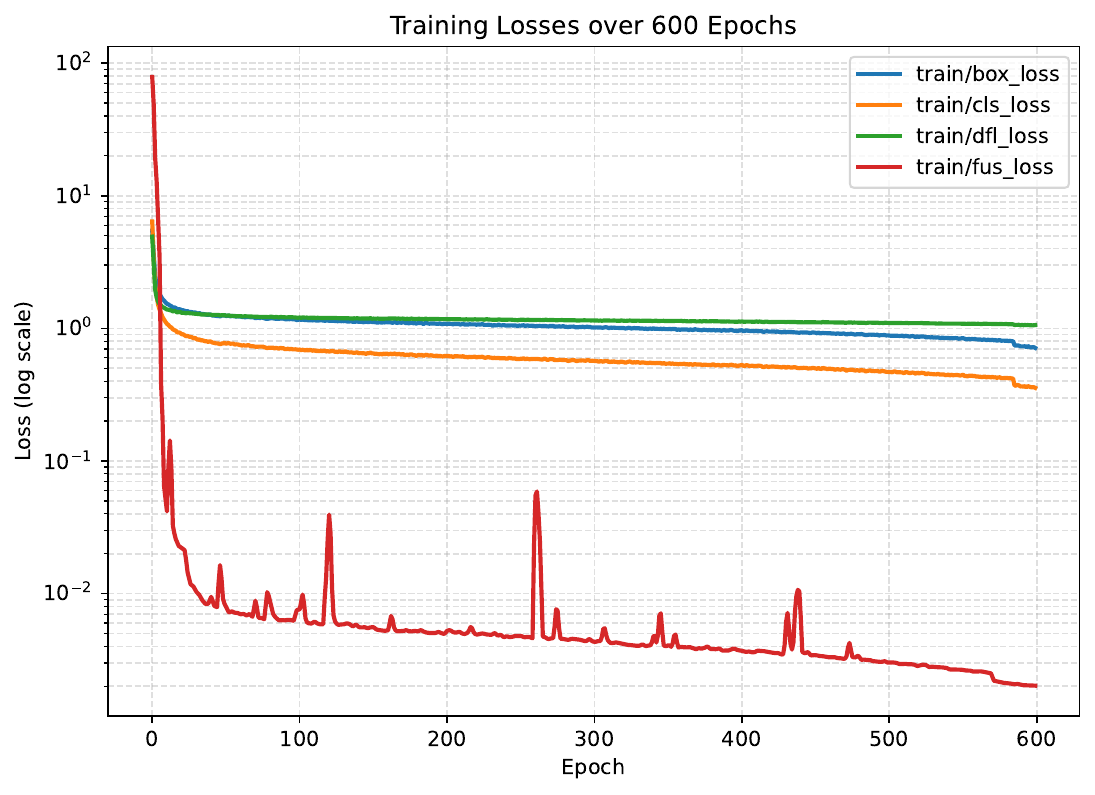}
		\caption{}
		\label{fig_loss}
	\end{subfigure}
	\hfill
	\begin{subfigure}[t]{0.49\columnwidth}
		\vspace{0pt}
		\centering
		\includegraphics[
		width=\linewidth,
		trim={2mm 2mm 2mm 2mm},
		clip
		]{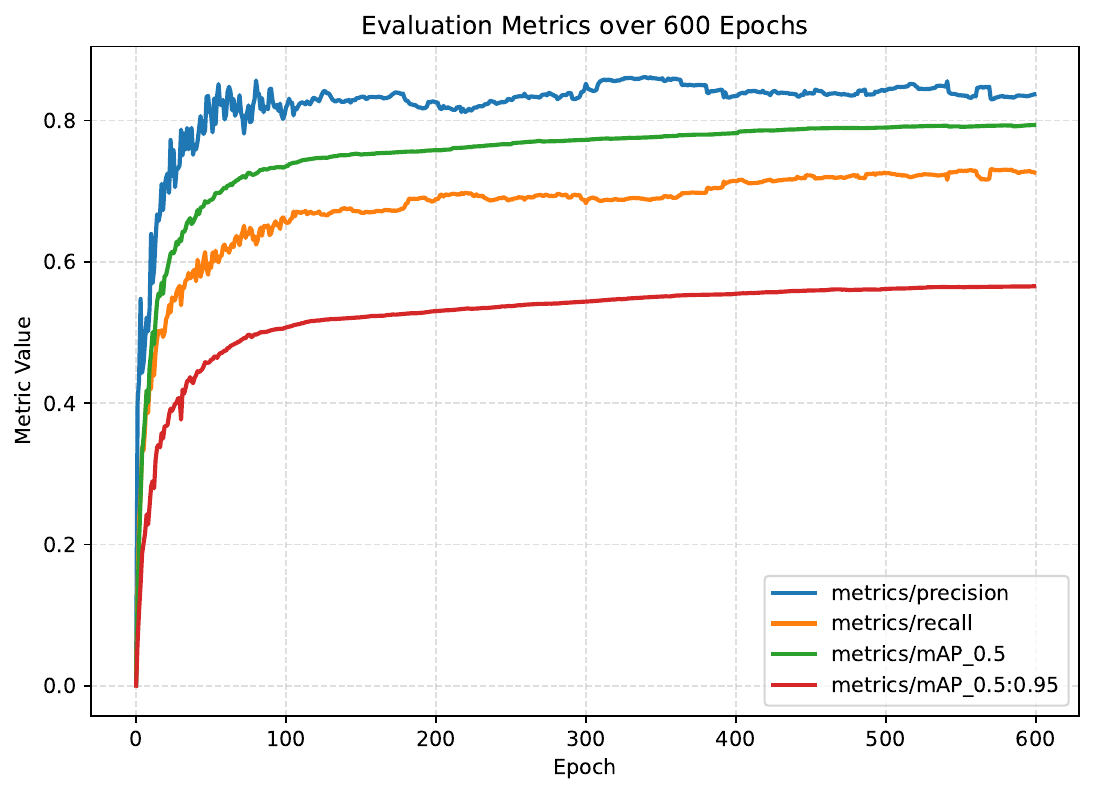}
		\caption{}
		\label{fig_metric}
	\end{subfigure}
	
	\caption{Evolution of training losses and evaluation metrics for CFGPNet-m on the MFAD dataset during 600 training epochs: 
		(a) box, class, DFL, and fusion loss values; 
		(b) precision, recall, mAP@50, and mAP@50:95 scores.}
	\label{fig_loss_metric}
\end{figure}

As shown in Fig.~\ref{fig_loss_metric}, declines were observed in the box, class, DFL\footnote{distribution focal loss}, and fusion losses, while precision, recall, mAP50, and mAP50:95 converged over 600 epochs. 
The fusion-loss value decreased sharply over the first ten epochs, from approximately $8\times10^{1}$ at epoch 1 to $10^{-1}$ at epoch 10.

\textbf{Ablation on learnable aggregation factors.} Table~\ref{tab_11} shows that the highest performance was obtained without an aggregation factor, reaching $79.6\%$ mAP50 and $56.7\%$ mAP50:95. 
\begin{table}[!h]
	\centering
	\begin{threeparttable}[t]
		\footnotesize
		\caption{Ablation study on learnable aggregation factors in the improved GELAN backbone. The best value in each column is highlighted in bold.}
		\label{tab_11}
		\setlength\tabcolsep{1.5pt}
		\begin{tabular}{lcc}
			\toprule
			\textbf{Aggregation factor} & \textbf{mAP50} & \textbf{mAP50:95} \\
			\midrule
			\textbf{None} & \textbf{79.6} & \textbf{56.7}\\
			\textbf{$\gamma_{x_{12}}$} & 70.8 & 48.3\\
			\textbf{$\gamma_{out}$} & 78.2 & 56.0\\
			\textbf{$\gamma_{x_{12}} + \gamma_{out}$} & 74.9 & 51.8\\
			\bottomrule
		\end{tabular}
	\end{threeparttable}
	\vspace{0pt}
\end{table}
Introducing $\gamma_{out}$ caused a limited decrease to $78.2\%/56.0\%$, whereas $\gamma_{x_{12}}$ reduced the scores to $70.8\%/48.3\%$; using both factors resulted in $74.9\%/51.8\%$. The stronger degradation caused by $\gamma_{x_{12}}$, which rescales the intermediate concatenated features, indicates that channel-wise modulation disrupts feature aggregation in this configuration. The unscaled aggregation path was therefore retained.

\textbf{Number of attention branches in MBatt.} Table~\ref{tab_12} shows that the highest accuracy was obtained with three branches in ASAF, reaching $79.6\%$ mAP50 and $56.7\%$ mAP50:95 at 56.9 GFLOPs, 15.3M parameters, and 91.7 FPS. 
\begin{table}[!h]
	\centering
	\begin{threeparttable}[t]
		\footnotesize
		\caption{Ablation study on the number of attention branches inside the MBatt structure implemented in the ASAF module. The best value in each column is highlighted in bold.}
		\label{tab_12}
		\setlength\tabcolsep{1.5pt}
		\begin{tabular}{lccccc}
			\toprule
			\textbf{Number of branches} & \textbf{mAP50} & \textbf{mAP50:95} & \textbf{GFLOPs (G)} & \textbf{\#Param. (M)} & \textbf{FPS} \\
			\midrule
			\textbf{2} & 77.0 & 53.8 & \textbf{56.9} & \textbf{15.3} & \textbf{92.9}\\
			\textbf{3} & \textbf{79.6} & \textbf{56.7} & \textbf{56.9} & \textbf{15.3} & 91.7\\
			\textbf{4} & 74.8 & 52.8 & 57.1 & 15.4 & 89.8\\
			\textbf{5} & 79.4 & 56.6 & 57.3 & 15.5 & 87.3\\
			\textbf{6} & 76.7 & 54.4 & 57.5 & 15.5 & 85.5\\
			\bottomrule
		\end{tabular}
	\end{threeparttable}
	\vspace{0pt}
\end{table}
Two branches reduced accuracy without lowering computational cost, while four branches caused a further decline. Five branches approached the three-branch result ($79.4\%/56.6\%$) but increased computational cost and reduced throughput; six branches provided no gain. Three branches were therefore retained as a suitable balance between attention diversity and efficiency.

\textbf{Effect of SE, MHA\footnote{multi-head attention}, and MFE in the improved backbone.} Table~\ref{tab_13} shows that each component improved detection accuracy individually. 
\begin{table}[!h]
	\centering
	\begin{threeparttable}[t]
		\footnotesize
		\caption{Ablation study on SE attention (in RepViT baseline), multi-head attention, and MFE module (in RepViTBottleneck). The best value in each column is highlighted in bold.}
		\label{tab_13}
		\setlength\tabcolsep{1.5pt}
		\begin{tabular}{lccccc}
			\toprule
			\textbf{Component} & \textbf{mAP50} & \textbf{mAP50:95} & \textbf{GFLOPs (G)} & \textbf{\#Param. (M)} & \textbf{FPS} \\
			\midrule
			\textbf{None} & 74.6 & 53.5 & \textbf{54.4} & \textbf{14.2} & \textbf{143.0}\\
			\textbf{SE} & 77.7 & 56.6 & \textbf{54.4} & 14.3 & 139.3\\
			\textbf{MFE} & 77.1 & 55.9 & \textbf{54.4} & \textbf{14.2} & \textbf{143.0}\\
			\textbf{MHA} & 77.3 & 55.2 & 55.3 & 14.5 & 96.6\\
			\textbf{SE + MFE + MHA} & \textbf{79.6} & \textbf{56.7} & 56.9 & 15.3 & 91.7\\
			\bottomrule
		\end{tabular}
	\end{threeparttable}
	\vspace{0pt}
\end{table}
SE produced the strongest single-component result at $77.7\%/56.6\%$ mAP50/mAP50:95, while MFE and MHA yielded $77.1\%/55.9\%$ and $77.3\%/55.2\%$, respectively. With all three components, the highest scores of $79.6\%/56.7\%$ were obtained at 56.9 GFLOPs, 15.3M parameters, and 91.7 FPS, suggesting complementary contributions; the joint configuration was therefore retained.

\textbf{Effect of attention design in the cross-attention stage.} Table~\ref{tab_14} shows that attention-based interaction improved mAP50:95 over direct concatenation, from $53.8\%$ to at least $54.7\%$. 
\begin{table}[!h]
	\centering
	\begin{threeparttable}[t]
		\footnotesize
		\caption{Ablation study on the effect of different attention modules in the cross-attention stage. The best value in each column is highlighted in bold.}
		\label{tab_14}
		\setlength\tabcolsep{1.5pt}
		\begin{tabular}{lccccc}
			\toprule
			\textbf{module} & \textbf{mAP50} & \textbf{mAP50:95} & \textbf{GFLOPs (G)} & \textbf{\#Param. (M)} & \textbf{FPS} \\
			\midrule
			\textbf{Concat} & 76.8 & 53.8 & \textbf{56.4} & \textbf{15.3} & \textbf{95.0}\\
			\midrule
			\textbf{ELA} & 76.5 & 54.7 & \textbf{56.4} & \textbf{15.3} & 93.7\\
			\textbf{EMA} & 79.4 & 55.8 & 56.9 & \textbf{15.3} & 89.4\\
			\textbf{CEA} & 79.2 & 56.1 & 56.9 & \textbf{15.3} & 91.7\\
			\midrule
			\textbf{Cross ELA} & 77.0 & 54.9 & \textbf{56.4} & \textbf{15.3} & 93.7\\
			\textbf{Cross EMA} & \textbf{79.9} & 56.5 & 56.9 & \textbf{15.3} & 89.4\\
			\textbf{Cross CEA} & 79.6 & \textbf{56.7} & 56.9 & \textbf{15.3} & 91.7\\
			\bottomrule
		\end{tabular}
	\end{threeparttable}
	\vspace{0pt}
\end{table}
Exchanging attention maps improved each corresponding module: Cross ELA, Cross EMA, and Cross CEA increased mAP50/mAP50:95 by $0.5/0.2$, $0.5/0.7$, and $0.4/0.6$ points, respectively, over their non-cross counterparts. Cross EMA achieved the highest mAP50 ($79.9\%$), whereas Cross CEA achieved the highest mAP50:95 ($56.7\%$) with 56.9 GFLOPs, 15.3M parameters, and 91.7 FPS. Cross CEA was therefore retained for its stronger localization performance at unchanged model complexity.

\subsection{Visualization}
\label{sec_4_6}

\begin{figure}[!t]
	\centering
	\includegraphics[width=\linewidth, trim={5mm 5mm 5mm 5mm}, clip]{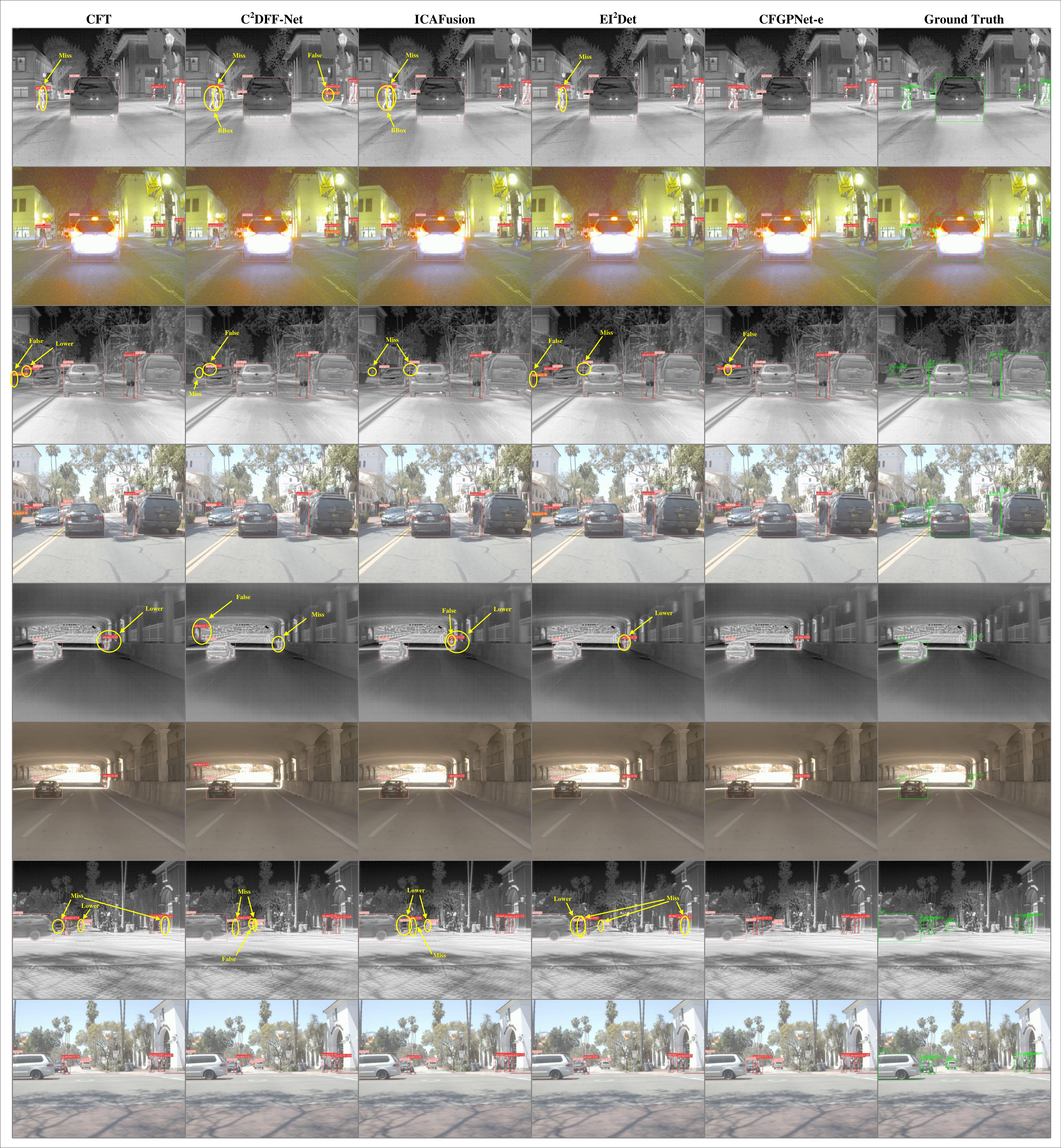}
	\caption{\centering Illustration of inference results for CFT, C$^2$DFF-Net, ICAFusion, EI$^2$Det, and CFGPNet-e on the FLIR dataset. The yellow arrow annotations show "Miss" (missed detection), "False" (false alarm), "Lower" (low-confidence detection), and "BBox" (inaccurately positioned bounding boxes).}\label{fig_13}
\end{figure}

\begin{figure}[!t]
	\centering
	\includegraphics[width=\linewidth, trim={5mm 5mm 5mm 5mm}, clip]{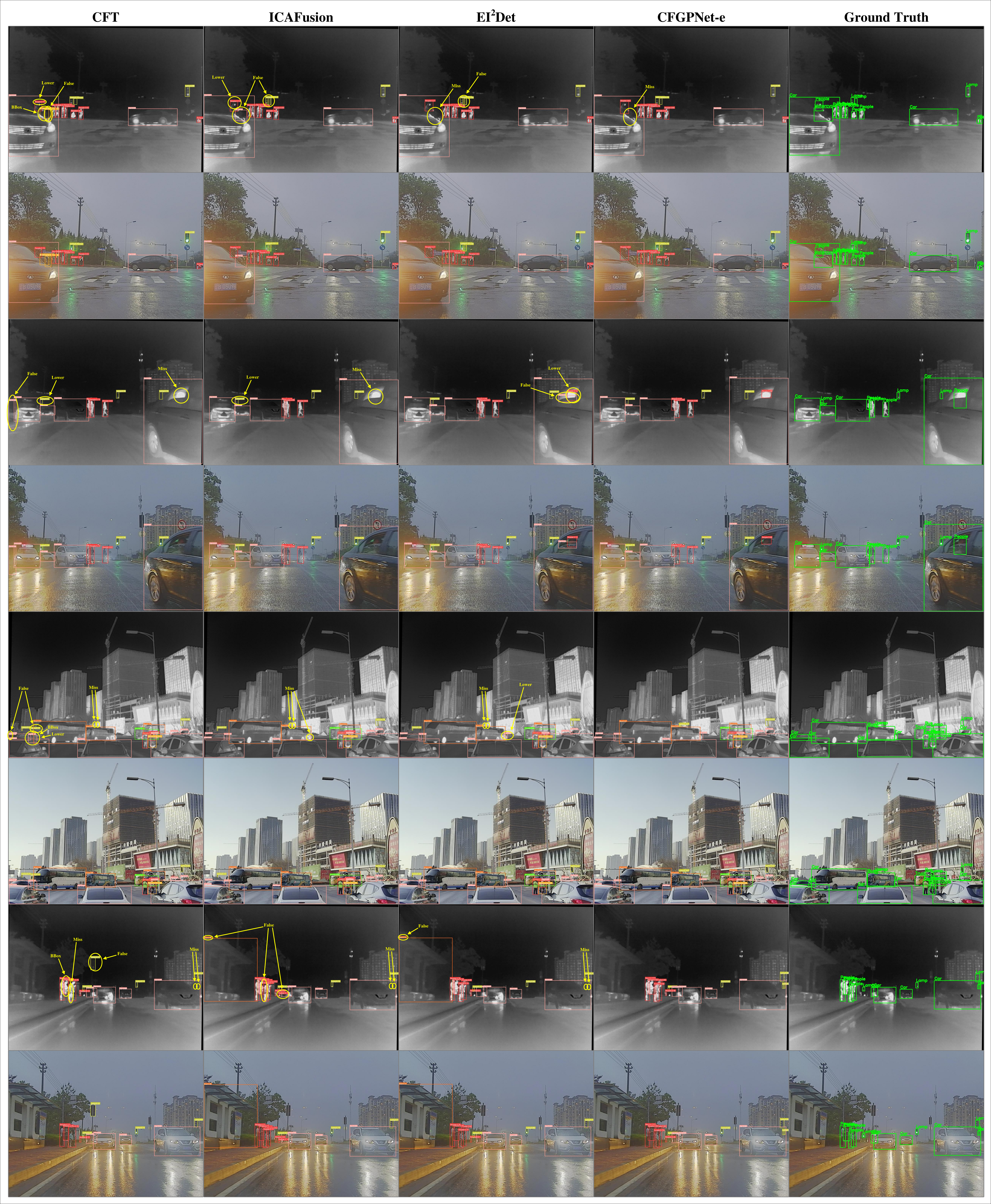}
	\caption{\centering Illustration of inference results for CFT, ICAFusion, EI$^2$Det, and CFGPNet-e on the M3FD dataset. The yellow arrow annotations show "Miss" (missed detection), "False" (false alarm), "Lower" (low-confidence detection), and "BBox" (inaccurately positioned bounding boxes).}\label{fig_14}
\end{figure}

\begin{figure}[!t]
	\centering
	\includegraphics[width=\linewidth, trim={5mm 5mm 5mm 5mm}, clip]{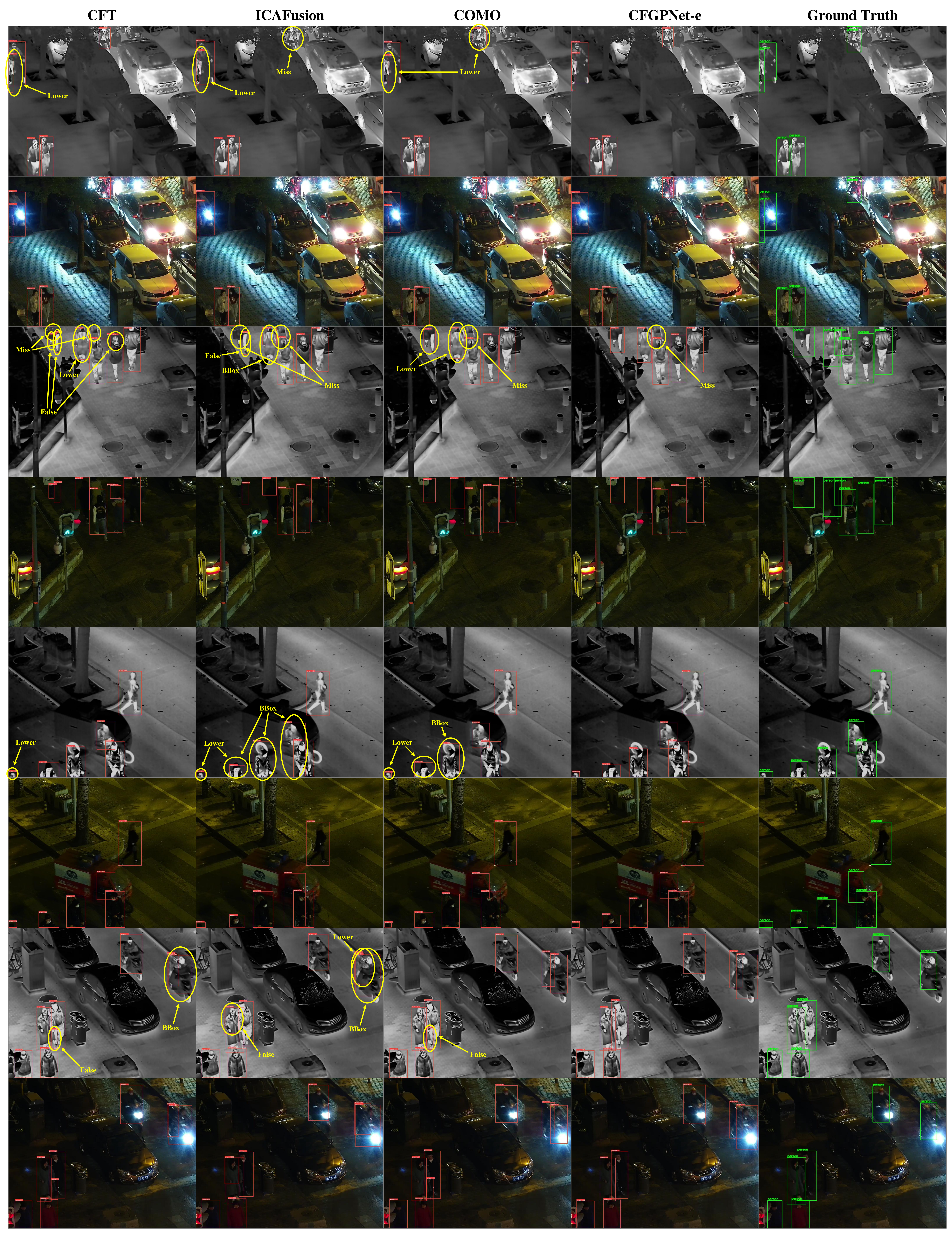}
	\caption{\centering Illustration of inference results for CFT, ICAFusion, COMO, and CFGPNet-e on the LLVIP dataset. The yellow arrow annotations show "Miss" (missed detection), "False" (false alarm), "Lower" (low-confidence detection), and "BBox" (inaccurately positioned bounding boxes).}\label{fig_15}
\end{figure}

\begin{figure}[!t]
	\centering
	\includegraphics[width=\linewidth, trim={5mm 5mm 5mm 5mm}, clip]{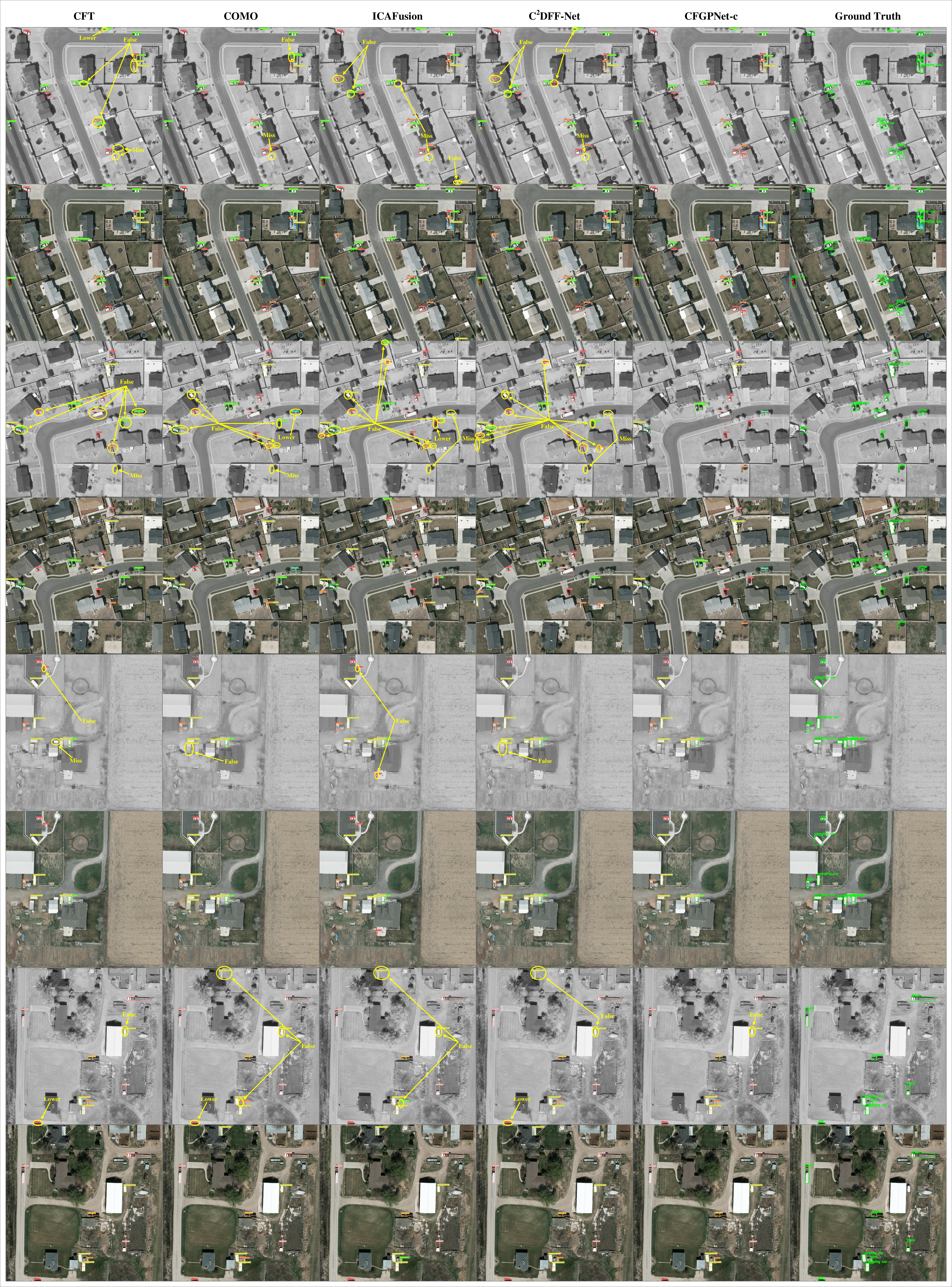}
	\caption{\centering Illustration of inference results for CFT, COMO, ICAFusion, C$^2$DFF-Net, and CFGPNet-c on VEDAI Split 1. The yellow arrow annotations show "Miss" (missed detection), "False" (false alarm), "Lower" (low-confidence detection), and "BBox" (inaccurately positioned bounding boxes).}\label{fig_16}
\end{figure}

\begin{figure}[!t]
	\centering
	\includegraphics[width=\linewidth, trim={5mm 5mm 5mm 5mm}, clip]{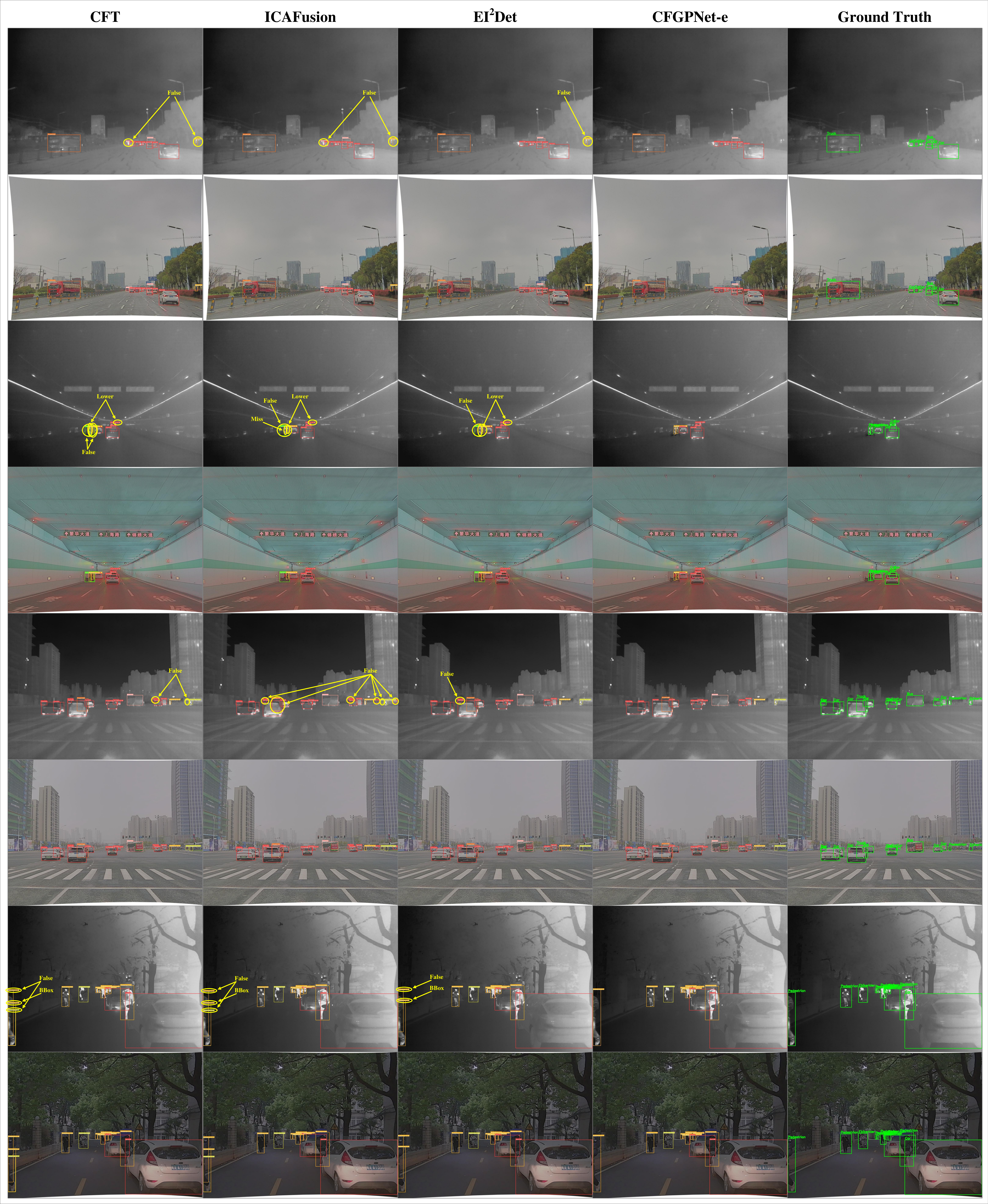}
	\caption{\centering Illustration of inference results for CFT, ICAFusion, EI$^2$Det, and CFGPNet-e on the MFAD dataset. The yellow arrow annotations show "Miss" (missed detection), "False" (false alarm), "Lower" (low-confidence detection), and "BBox" (inaccurately positioned bounding boxes).}\label{fig_17}
\end{figure}

Qualitative results on FLIR, M3FD, LLVIP, VEDAI, and MFAD are presented in Figs.~\ref{fig_13}--\ref{fig_17}. Representative cases are used to examine localization, confidence, missed detections, and false alarms under difficult illumination, occlusion, truncation, and scale conditions.

Across FLIR, M3FD, and LLVIP (Figs.~\ref{fig_13}--\ref{fig_15}), tighter localization and higher confidence in the illustrated cases were generally obtained with CFGPNet than with the compared detectors. 
The clearest differences were observed for crowded FLIR scenes, small or occluded objects in M3FD, and truncated LLVIP pedestrians. Isolated errors remained, including a false pedestrian on FLIR, a missed distant motorcycle on M3FD, and a missed heavily occluded pedestrian on LLVIP; these cases were visually ambiguous and were also difficult for the comparison methods.

On VEDAI and MFAD (Figs.~\ref{fig_16}--\ref{fig_17}), fewer false alarms and more precise boxes were generally observed for small, occluded, and truncated objects. 
The low-confidence camping-car prediction on VEDAI was visually plausible and was also produced by the comparison methods, while an additional border vehicle detected by CFGPNet-c may reflect incomplete annotation. The remaining difficult cases were concentrated around partial visibility and class ambiguity, consistent with the quantitative results.

\section{Conclusion}
\label{sec_5}

In this work, CFGPNet was developed for RGB--T object detection. An improved GELAN backbone with RepViT blocks was combined with CrossCEA for cross-modal attention exchange and ASAF for compact feature fusion. A PGI auxiliary branch was used only during training and removed at inference. Competitive performance was obtained on FLIR, M3FD, LLVIP, VEDAI, and MFAD across three model scales, providing distinct accuracy--efficiency operating points.

The final configuration was supported by the MFAD ablations: CE--KL was retained as the fusion loss, three MBatt branches were selected, learnable aggregation factors were omitted, and SE, MFE, MHA, and CrossCEA were jointly used. Since the PGI branch was removed after training, additional supervision was provided without enlarging the inference model. Further validation across multiple random seeds and additional VEDAI splits remains for future work.


\section*{Acknowledgements}
The authors acknowledge the Machine Vision Research Laboratory at Amirkabir University of Technology for providing the computational resources used in this study. 

\section*{Funding}
This research did not receive any grant from funding agencies in the public, commercial, or not-for-profit sectors.

\appendix 

\section{Implementation Details} 
\label{app_implementation_details} 

This appendix summarizes the implementation details required to reproduce the CFGPNet variants. The optimization schedule, loss gains, warm-up settings, and augmentation parameters are reported in Table~\ref{tab_15}, while the layer-wise network configurations are provided in Table~\ref{tab_16}.
\begin{table}[h]
	\centering
	\begin{threeparttable}[h]
		\caption{Training hyperparameter settings of CFGPNet, including the learning-rate schedule, warm-up configuration, loss gains, and data-augmentation parameters.}
		\label{tab_15}
		\begin{tabular}{lc}
			\toprule
			\textbf{hyperparameter} & \textbf{Value} \\	
			\midrule
			epochs & 600 \\
			optimizer & SGD \\
			initial learning rate & 0.01 \\
			final learning rate & 0.0001 \\
			learning rate decay & linear \\
			momentum & 0.937 \\
			weight decay & 0.0005 \\
			warm-up epochs & 3 \\
			warm-up momentum & 0.8 \\
			warm-up bias learning rate & 0.1 \\
			box loss gain & 7.5 \\
			class loss gain & 0.5 \\
			DFL loss gain & 1.5 \\
			HSV saturation augmentation & 0.7 \\
			HSV value augmentation & 0.4 \\
			translation augmentation & 0.1 \\
			scale augmentation & 0.9 \\
			mosaic augmentation & 1.0 \\
			MixUp augmentation & 0.15 \\
			copy \& paste augmentation & 0.3 \\
			close mosaic epochs & 15 \\
			\bottomrule
		\end{tabular}
	\end{threeparttable}
\end{table}
\begin{table}[!t]
	\centering
	
	\caption{Network configurations of the three CFGPNet variants. 
		The left subtable presents CFGPNet-m and CFGPNet-c, while the right
		subtable presents CFGPNet-e.}
	\label{tab_16}
	
	\captionsetup[subtable]{
		font=scriptsize,
		justification=centering,
		singlelinecheck=false,
		skip=2pt
	}
	
	\begin{subtable}[t]{0.49\columnwidth}
		\vspace{0pt}
		\centering
		
		\caption{Network configurations of CFGPNet-m and CFGPNet-c.
			The Filters column is reported as m/c when the two variants differ.}
		\label{tab_cfgpnet_mc_configuration}
		
		{%
			\fontsize{4}{4.4}\selectfont
			\setlength{\tabcolsep}{0.25pt}
			\renewcommand{\arraystretch}{0.68}
			
			\resizebox{\linewidth}{!}{%
				\begin{tabular}{lcccccc}
					\toprule
					\textbf{Index}
					& \textbf{Module}
					& \textbf{Route}
					& \textbf{Filters (m/c)}
					& \textbf{Depth/Select}
					& \textbf{Size}
					& \textbf{Stride} \\
					\midrule
					0 & Silence & -- & 3, 3 & -- & -- & 1 \\
					1 & Conv & 0 & 32 / 64 & -- & 3 & 2 \\
					2 & Conv & 1 & 64 / 128 & -- & 3 & 2 \\
					3 & RepViTCSPELAN4 & 2 & 128, 128, 64 / 256, 128, 64 & 2, 1 & -- & 1 \\
					4 & ADown & 3 & 128 / 256 & -- & 3 & 2 \\
					5 & RepViTCSPELAN4 & 4 & 128, 128, 64 / 512, 256, 128 & 2, 1 & -- & 1 \\
					6 & ADown & 5 & 256 / 512 & -- & 3 & 2 \\
					7 & RepViTCSPELAN4 & 6 & 256, 256, 128 / 512, 512, 256 & 2, 1 & -- & 1 \\
					8 & ADown & 7 & 384 / 512 & -- & 3 & 2 \\
					9 & RepViTCSPELAN4 & 8 & 384, 384, 192 / 512, 512, 256 & 2, 1 & -- & 1 \\
					10 & Conv & 0 & 32 / 64 & -- & 3 & 2 \\
					11 & Conv & 10 & 64 / 128 & -- & 3 & 2 \\
					12 & RepViTCSPELAN4 & 11 & 128, 128, 64 / 256, 128, 64 & 2, 1 & -- & 1 \\
					13 & ADown & 12 & 128 / 256 & -- & 3 & 2 \\
					14 & RepViTCSPELAN4 & 13 & 128, 128, 64 / 512, 256, 128 & 2, 1 & -- & 1 \\
					15 & ADown & 14 & 256 / 512 & -- & 3 & 2 \\
					16 & RepViTCSPELAN4 & 15 & 256, 256, 128 / 512, 512, 256 & 2, 1 & -- & 1 \\
					17 & ADown & 16 & 384 / 512 & -- & 3 & 2 \\
					18 & RepViTCSPELAN4 & 17 & 384, 384, 192 / 512, 512, 256 & 2, 1 & -- & 1 \\
					19 & CrossCEA & 5, 14 & 128 / 512 & -- & -- & 1 \\
					20 & CrossCEA & 7, 16 & 256 / 512 & -- & -- & 1 \\
					21 & CrossCEA & 9, 18 & 384 / 512 & -- & -- & 1 \\
					22 & ASAF & 19 & 128 / 512 & -- & -- & 1 \\
					23 & ASAF & 20 & 256 / 512 & -- & -- & 1 \\
					24 & ASAF & 21 & 384 / 512 & -- & -- & 1 \\
					25 & SPPELAN & 24 & 384, 192, 192 / 512, 256, 256 & 3, 1 & -- & 1 \\
					26 & Upsample & 25 & 384 / 512 & -- & -- & 2 \\
					27 & Concat & 26, 23 & 640 / 1024 & -- & -- & 1 \\
					28 & RepViTCSPELAN4 & 27 & 256, 256, 128 / 512, 512, 256 & 2, 1 & -- & 1 \\
					29 & Upsample & 28 & 256 / 512 & -- & -- & 2 \\
					30 & Concat & 29, 22 & 384 / 1024 & -- & -- & 1 \\
					31 & RepViTCSPELAN4 & 30 & 128, 128, 64 / 256, 256, 128 & 2, 1 & -- & 1 \\
					32 & ADown & 31 & 128 / 256 & -- & 3 & 2 \\
					33 & Concat & 32, 28 & 384 / 768 & -- & -- & 1 \\
					34 & RepViTCSPELAN4 & 33 & 256, 256, 128 / 512, 512, 256 & 2, 1 & -- & 1 \\
					35 & ADown & 34 & 384 / 512 & -- & 3 & 2 \\
					36 & Concat & 35, 25 & 768 / 1024 & -- & -- & 1 \\
					37 & RepViTCSPELAN4 & 36 & 384, 384, 192 / 512, 512, 256 & 2, 1 & -- & 1 \\
					38 & CBLinear & 22 & 128 / 256 & -- & -- & 1 \\
					39 & CBLinear & 23 & 128, 256 / 256, 512 & -- & -- & 1 \\
					40 & CBLinear & 24 & 128, 256, 384 / 256, 512, 512 & -- & -- & 1 \\
					41 & Conv & 0 & 32 / 64 & -- & 3 & 2 \\
					42 & Conv & 41 & 64 / 128 & -- & 3 & 2 \\
					43 & RepViTCSPELAN4 & 42 & 128, 128, 64 / 256, 128, 64 & 2, 1 & -- & 1 \\
					44 & ADown & 43 & 128 / 256 & -- & 3 & 2 \\
					45 & Conv & 0 & 32 / 64 & -- & 3 & 2 \\
					46 & Conv & 45 & 64 / 128 & -- & 3 & 2 \\
					47 & RepViTCSPELAN4 & 46 & 128, 128, 64 / 256, 128, 64 & 2, 1 & -- & 1 \\
					48 & ADown & 47 & 128 / 256 & -- & 3 & 2 \\
					49 & CrossCEA & 44, 48 & 128 / 256 & -- & -- & 1 \\
					50 & ASAF & 49 & 128 / 256 & -- & -- & 1 \\
					51 & CBFuse & 38, 39, 40, 50 & 128 / 256 & 0, 0, 0 & -- & 1 \\
					52 & RepViTCSPELAN4 & 51 & 128, 128, 64 / 512, 256, 128 & 2, 1 & -- & 1 \\
					53 & ADown & 52 & 256 / 512 & -- & 3 & 2 \\
					54 & CBFuse & 39, 40, 53 & 256 / 512 & 1, 1 & -- & 1 \\
					55 & RepViTCSPELAN4 & 54 & 256, 256, 128 / 512, 512, 256 & 2, 1 & -- & 1 \\
					56 & ADown & 55 & 384 / 512 & -- & 3 & 2 \\
					57 & CBFuse & 40, 56 & 384 / 512 & 2 & -- & 1 \\
					58 & RepViTCSPELAN4 & 57 & 384, 384, 192 / 512, 512, 256 & 2, 1 & -- & 1 \\
					59 & DualDDetect & 52, 55, 58, 31, 34, 37 & -- & -- & -- & -- \\
					\bottomrule
				\end{tabular}%
			}
		}
	\end{subtable}
	\hfill
	\begin{subtable}[t]{0.49\columnwidth}
		\vspace{0pt}
		\centering
		
		\caption{Network configuration of CFGPNet-e.}
		\label{tab_cfgpnet_e_configuration}
		
		{%
			\fontsize{4}{4.4}\selectfont
			\setlength{\tabcolsep}{0.25pt}
			\renewcommand{\arraystretch}{0.68}
			
			\resizebox{\linewidth}{!}{%
				\begin{tabular}{lcccccc}
					\toprule
					\textbf{Index}
					& \textbf{Module}
					& \textbf{Route}
					& \textbf{Filters}
					& \textbf{Depth/Select}
					& \textbf{Size}
					& \textbf{Stride} \\
					\midrule
					0 & Silence & -- & 3, 3 & -- & -- & 1 \\
					1 & Conv & 0 & 64 & 0 & 3 & 2 \\
					2 & Conv & 1 & 128 & -- & 3 & 2 \\
					3 & RepViTCSPELAN4 & 2 & 256, 128, 64 & 2, 2 & -- & 1 \\
					4 & ADown & 3 & 256 & -- & 3 & 2 \\
					5 & RepViTCSPELAN4 & 4 & 512, 256, 128 & 2, 2 & -- & 1 \\
					6 & ADown & 5 & 512 & -- & 3 & 2 \\
					7 & RepViTCSPELAN4 & 6 & 1024, 512, 256 & 2, 2 & -- & 1 \\
					8 & ADown & 7 & 1024 & -- & 3 & 2 \\
					9 & RepViTCSPELAN4 & 8 & 1024, 512, 256 & 2, 2 & -- & 1 \\
					10 & CBLinear & 1 & 64 & -- & 1 & 1 \\
					11 & CBLinear & 3 & 64, 128 & -- & 1 & 1 \\
					12 & CBLinear & 5 & 64, 128, 256 & -- & 1 & 1 \\
					13 & CBLinear & 7 & 64, 128, 256, 512 & -- & 1 & 1 \\
					14 & CBLinear & 9 & 64, 128, 256, 512, 1024 & -- & 1 & 1 \\
					15 & Conv & 0 & 64 & 2 & 3 & 2 \\
					16 & CBFuse & 10, 11, 12, 13, 14, 15 & 64 & 0, 0, 0, 0, 0 & -- & 1 \\
					17 & Conv & 16 & 128 & -- & 3 & 2 \\
					18 & CBFuse & 11, 12, 13, 14, 17 & 128 & 1, 1, 1, 1 & -- & 1 \\
					19 & RepViTCSPELAN4 & 18 & 256, 128, 64 & 2, 2 & -- & 1 \\
					20 & ADown & 19 & 256 & -- & 3 & 2 \\
					21 & CBFuse & 12, 13, 14, 20 & 256 & 2, 2, 2 & -- & 1 \\
					22 & RepViTCSPELAN4 & 21 & 512, 256, 128 & 2, 2 & -- & 1 \\
					23 & ADown & 22 & 512 & -- & 3 & 2 \\
					24 & CBFuse & 13, 14, 23 & 512 & 3, 3 & -- & 1 \\
					25 & RepViTCSPELAN4 & 24 & 1024, 512, 256 & 2, 2 & -- & 1 \\
					26 & ADown & 25 & 1024 & -- & 3 & 2 \\
					27 & CBFuse & 14, 26 & 1024 & 4 & -- & 1 \\
					28 & RepViTCSPELAN4 & 27 & 1024, 512, 256 & 2, 2 & -- & 1 \\
					29 & Conv & 0 & 64 & 1 & 3 & 2 \\
					30 & Conv & 29 & 128 & -- & 3 & 2 \\
					31 & RepViTCSPELAN4 & 30 & 256, 128, 64 & 2, 2 & -- & 1 \\
					32 & ADown & 31 & 256 & -- & 3 & 2 \\
					33 & RepViTCSPELAN4 & 32 & 512, 256, 128 & 2, 2 & -- & 1 \\
					34 & ADown & 33 & 512 & -- & 3 & 2 \\
					35 & RepViTCSPELAN4 & 34 & 1024, 512, 256 & 2, 2 & -- & 1 \\
					36 & ADown & 35 & 1024 & -- & 3 & 2 \\
					37 & RepViTCSPELAN4 & 36 & 1024, 512, 256 & 2, 2 & -- & 1 \\
					38 & CBLinear & 1 & 64 & -- & 1 & 1 \\
					39 & CBLinear & 3 & 64, 128 & -- & 1 & 1 \\
					40 & CBLinear & 5 & 64, 128, 256 & -- & 1 & 1 \\
					41 & CBLinear & 7 & 64, 128, 256, 512 & -- & 1 & 1 \\
					42 & CBLinear & 9 & 64, 128, 256, 512, 1024 & -- & 1 & 1 \\
					43 & Conv & 0 & 64 & 2 & 3 & 2 \\
					44 & CBFuse & 10, 11, 12, 13, 14, 43 & 64 & 0, 0, 0, 0, 0 & -- & 1 \\
					45 & Conv & 44 & 128 & -- & 3 & 2 \\
					46 & CBFuse & 11, 12, 13, 14, 45 & 128 & 1, 1, 1, 1 & -- & 1 \\
					47 & RepViTCSPELAN4 & 46 & 256, 128, 64 & 2, 2 & -- & 1 \\
					48 & ADown & 47 & 256 & -- & 3 & 2 \\
					49 & CBFuse & 12, 13, 14, 48 & 256 & 2, 2, 2 & -- & 1 \\
					50 & RepViTCSPELAN4 & 49 & 512, 256, 128 & 2, 2 & -- & 1 \\
					51 & ADown & 50 & 512 & -- & 3 & 2 \\
					52 & CBFuse & 13, 14, 51 & 512 & 3, 3 & -- & 1 \\
					53 & RepViTCSPELAN4 & 52 & 1024, 512, 256 & 2, 2 & -- & 1 \\
					54 & ADown & 53 & 1024 & -- & 3 & 2 \\
					55 & CBFuse & 14, 54 & 1024 & 4 & -- & 1 \\
					56 & RepViTCSPELAN4 & 55 & 1024, 512, 256 & 2, 2 & -- & 1 \\
					57 & CrossCEA & 5, 33 & 512 & -- & -- & 1 \\
					58 & CrossCEA & 7, 35 & 1024 & -- & -- & 1 \\
					59 & CrossCEA & 9, 37 & 1024 & -- & -- & 1 \\
					60 & ASAF & 57 & 512 & -- & -- & 1 \\
					61 & ASAF & 58 & 1024 & -- & -- & 1 \\
					62 & ASAF & 59 & 1024 & -- & -- & 1 \\
					63 & SPPCSPC & 62 & 512 & 1 & 5, 9, 13 & 1 \\
					64 & Upsample & 63 & 512 & -- & -- & $\times 2$ \\
					65 & Concat & 64, 61 & 1536 & -- & -- & 1 \\
					66 & RepViTCSPELAN4 & 65 & 512, 512, 256 & 2, 2 & -- & 1 \\
					67 & Upsample & 66 & 512 & -- & -- & $\times 2$ \\
					68 & Concat & 67, 60 & 1024 & -- & -- & 1 \\
					69 & RepViTCSPELAN4 & 68 & 256, 256, 128 & 2, 2 & -- & 1 \\
					70 & CrossCEA & 22, 50 & 512 & -- & -- & 1 \\
					71 & CrossCEA & 25, 53 & 1024 & -- & -- & 1 \\
					72 & CrossCEA & 28, 56 & 1024 & -- & -- & 1 \\
					73 & ASAF & 70 & 512 & -- & -- & 1 \\
					74 & ASAF & 71 & 1024 & -- & -- & 1 \\
					75 & ASAF & 72 & 1024 & -- & -- & 1 \\
					76 & SPPCSPC & 75 & 512 & 1 & 5, 9, 13 & 1 \\
					77 & Upsample & 76 & 512 & -- & -- & $\times 2$ \\
					78 & Concat & 77, 74 & 1536 & -- & -- & 1 \\
					79 & RepViTCSPELAN4 & 78 & 512, 512, 256 & 2, 2 & -- & 1 \\
					80 & Upsample & 79 & 512 & -- & -- & $\times 2$ \\
					81 & Concat & 80, 73 & 1024 & -- & -- & 1 \\
					82 & RepViTCSPELAN4 & 81 & 256, 256, 128 & 2, 2 & -- & 1 \\
					83 & ADown & 82 & 256 & -- & 3 & 2 \\
					84 & Concat & 83, 79 & 768 & -- & -- & 1 \\
					85 & RepViTCSPELAN4 & 84 & 512, 512, 256 & 2, 2 & -- & 1 \\
					86 & ADown & 85 & 512 & -- & 3 & 2 \\
					87 & Concat & 86, 76 & 1024 & -- & -- & 1 \\
					88 & RepViTCSPELAN4 & 87 & 512, 1024, 512 & 2, 2 & -- & 1 \\
					89 & DualDDetect & 69, 66, 63, 82, 85, 88 & -- & -- & -- & -- \\
					\bottomrule
				\end{tabular}%
			}
		}
	\end{subtable}
	
	\vspace{-3pt}
\end{table} 

The same overall computational architecture is used for CFGPNet-m and CFGPNet-c, with different channel widths. CFGPNet-e is implemented with wider channels, deeper blocks, and an expanded multilevel auxiliary pathway. In Table~\ref{tab_16}, the route, filter, depth/select, kernel-size, and stride settings specify the input connections, channel dimensions, repetitions, and spatial operations of each layer. These specifications make the differences among the three variants explicit and support their reproduction.

%
%




%
%
%

\bibliographystyle{elsarticle-num}
\bibliography{refs}

@ARTICLE{11180153,
	author={Zhang, Yue and Chen, Jinbao and Wang, Jianyuan and Shi, Donghao and Han, Shu and Deng, Lixiao},
	journal={IEEE Transactions on Geoscience and Remote Sensing}, 
	title={C2DFF-Net for Object Detection in Multimodal Remote Sensing Images}, 
	year={2025},
	volume={63},
	number={},
	pages={1-16},
	doi={10.1109/TGRS.2025.3614295}}

@inproceedings{cao2023multimodal,
	title={Multimodal object detection by channel switching and spatial attention},
	author={Cao, Yue and Bin, Junchi and Hamari, Jozsef and Blasch, Erik and Liu, Zheng},
	booktitle={Proceedings of the IEEE/CVF Conference on Computer Vision and Pattern Recognition},
	pages={403--411},
	year={2023}
}

@inproceedings{chen2022mobile,
	title={Mobile-former: Bridging mobilenet and transformer},
	author={Chen, Yinpeng and Dai, Xiyang and Chen, Dongdong and Liu, Mengchen and Dong, Xiaoyi and Yuan, Lu and Liu, Zicheng},
	booktitle={Proceedings of the IEEE/CVF conference on computer vision and pattern recognition},
	pages={5270--5279},
	year={2022}
}

@article{chen2025amfd,
  title={AMFD: Distillation via adaptive multimodal fusion for multispectral pedestrian detection},
  author={Chen, Zizhao and Qian, Yeqiang and Yang, Xiaoxiao and Wang, Chunxiang and Yang, Ming},
  journal={IEEE Transactions on Multimedia},
  year={2025},
  publisher={IEEE}
}

@inproceedings{chen2025vif,
	title={VIF-YOLO: Visible-Infrared Fusion YOLO for Remote Sensing Small Object Detection},
	author={Chen, Renjie and Sun, Hua and Fan, Haiyang and Wu, Pingxiang},
	booktitle={2025 IEEE 31th International Conference on Parallel and Distributed Systems (ICPADS)},
	pages={1--8},
	year={2025},
	organization={IEEE}
}

@inproceedings{ding2021repvgg,
  title={Repvgg: Making vgg-style convnets great again},
  author={Ding, Xiaohan and Zhang, Xiangyu and Ma, Ningning and Han, Jungong and Ding, Guiguang and Sun, Jian},
  booktitle={Proceedings of the IEEE/CVF conference on computer vision and pattern recognition},
  pages={13733--13742},
  year={2021}
}

@inproceedings{girshick2015fast,
  title={Fast r-cnn},
  author={Girshick, Ross},
  booktitle={Proceedings of the IEEE international conference on computer vision},
  pages={1440--1448},
  year={2015}
}

@inproceedings{hou2021coordinate,
  title={Coordinate attention for efficient mobile network design},
  author={Hou, Qibin and Zhou, Daquan and Feng, Jiashi},
  booktitle={Proceedings of the IEEE/CVF conference on computer vision and pattern recognition},
  pages={13713--13722},
  year={2021}
}

@inproceedings{hu2018squeeze,
  title={Squeeze-and-excitation networks},
  author={Hu, Jie and Shen, Li and Sun, Gang},
  booktitle={Proceedings of the IEEE conference on computer vision and pattern recognition},
  pages={7132--7141},
  year={2018}
}

@article{hu2025ei,
  title={Ei 2 det: Edge-guided illumination-aware interactive learning for visible-infrared object detection},
  author={Hu, Ke and He, Yudong and Li, Yuan and Zhao, Jiayu and Chen, Song and Kang, Yi},
  journal={IEEE Transactions on Circuits and Systems for Video Technology},
  year={2025},
  publisher={IEEE}
}

@inproceedings{huang2017densely,
  title={Densely connected convolutional networks},
  author={Huang, Gao and Liu, Zhuang and Van Der Maaten, Laurens and Weinberger, Kilian Q},
  booktitle={Proceedings of the IEEE conference on computer vision and pattern recognition},
  pages={4700--4708},
  year={2017}
}

@article{huang2026cdfit,
	title={CDFIT: A Transformer Using Cross-Modal Dual-Stream Feature Interaction for Multispectral Pedestrian Detection},
	author={Huang, Zihao and Li, Wenshi and Zhang, Yuzhen and Guo, Jiaren and Zheng, Jianyin and Ji, Guang and Tao, Yanyun},
	journal={IEEE Transactions on Intelligent Transportation Systems},
	year={2026},
	publisher={IEEE}
}

@inproceedings{jang2025camdet,
	title={CAMDet: Condition-adaptive multispectral object detection using a visible-thermal translation model},
	author={Jang, Junbo and Lee, Jiyoon and Paik, Joonki},
	booktitle={ICASSP 2025-2025 IEEE International Conference on Acoustics, Speech and Signal Processing (ICASSP)},
	pages={1--5},
	year={2025},
	organization={IEEE}
}

@inproceedings{jang2025multispectral,
	title={Multispectral Object Detection Enhanced by Cross-Modal Information Complementary and Cosine Similarity Channel Resampling Modules},
	author={Jang, Junbo and Park, Chanyeong and Kim, Heegwang and Lee, Jiyoon and Paik, Joonki},
	booktitle={2025 IEEE/CVF Winter Conference on Applications of Computer Vision (WACV)},
	pages={9437--9446},
	year={2025},
	organization={IEEE}
}

@inproceedings{jia2021llvip,
  title={LLVIP: A visible-infrared paired dataset for low-light vision},
  author={Jia, Xinyu and Zhu, Chuang and Li, Minzhen and Tang, Wenqi and Zhou, Wenli},
  booktitle={Proceedings of the IEEE/CVF international conference on computer vision},
  pages={3496--3504},
  year={2021}
}

@article{jocher2022ultralyticsv7,
	title={ultralytics/yolov5: v7. 0-yolov5 sota realtime instance segmentation},
	author={Jocher, Glenn and Chaurasia, Ayush and Stoken, Alex and Borovec, Jirka and Kwon, Yonghye and Michael, Kalen and Fang, Jiacong and Yifu, Zeng and Wong, Colin and Montes, Diego and others},
	journal={Zenodo},
	year={2022}
}

@article{khanam2024yolov11,
	title={Yolov11: An overview of the key architectural enhancements},
	author={Khanam, Rahima and Hussain, Muhammad},
	journal={arXiv preprint arXiv:2410.17725},
	year={2024}
}

@book{kullback1997information,
  title={Information theory and statistics},
  author={Kullback, Solomon},
  year={1997},
  publisher={Courier Corporation}
}

@article{lan2025kdet,
	title={KDET-HPFL: A Personalized Federated Learning Framework for Multimodal Pedestrian Detection With Adaptive Feature Selection},
	author={Lan, Rukai and Zhang, Yong and Wang, Zidong and Liu, Weibo and Yang, Rui},
	journal={IEEE Internet of Things Journal},
	year={2025},
	publisher={IEEE}
}

@article{lee2024crossformer,
	title={CrossFormer: Cross-guided attention for multi-modal object detection},
	author={Lee, Seungik and Park, Jaehyeong and Park, Jinsun},
	journal={Pattern Recognition Letters},
	volume={179},
	pages={144--150},
	year={2024},
	publisher={Elsevier}
}

@article{li2025cross,
	title={Cross-modal edge-enhanced detector for UAV-based multispectral object detection},
	author={Li, Gong and Ren, Guoyin and Wang, Jingyu and Zhi, Mobing and Yu, Zhijie and Jiang, Bo and Guan, Haoliang and Guo, Qidan},
	journal={Scientific Reports},
	year={2025},
	publisher={Nature Publishing Group UK London}
}

@article{li2025crossmodalnet,
	title={CrossModalNet: A dual-modal object detection network based on cross-modal fusion and channel interaction},
	author={Li, Hanyun and Xiao, Linsong and Cao, Lihua and Wu, Di and Liu, Yangfan and Li, Yi and Zhang, Yunfeng and Bao, Haiyang},
	journal={Expert Systems with Applications},
	pages={129677},
	year={2025},
	publisher={Elsevier}
}

@inproceedings{liu2022target,
  title={Target-aware dual adversarial learning and a multi-scenario multi-modality benchmark to fuse infrared and visible for object detection},
  author={Liu, Jinyuan and Fan, Xin and Huang, Zhanbo and Wu, Guanyao and Liu, Risheng and Zhong, Wei and Luo, Zhongxuan},
  booktitle={Proceedings of the IEEE/CVF conference on computer vision and pattern recognition},
  pages={5802--5811},
  year={2022}
}

@inproceedings{liu2023efficientvit,
	title={Efficientvit: Memory efficient vision transformer with cascaded group attention},
	author={Liu, Xinyu and Peng, Houwen and Zheng, Ningxin and Yang, Yuqing and Hu, Han and Yuan, Yixuan},
	booktitle={Proceedings of the IEEE/CVF conference on computer vision and pattern recognition},
	pages={14420--14430},
	year={2023}
}

@article{liu2025fusion,
	title={A Fusion-Enhanced Network for Infrared and Visible High-Level Vision Tasks},
	author={Liu, Fangcen and Gao, Chenqiang and Chen, Fang and Li, Pengcheng and Guo, Junjie and Meng, Deyu},
	journal={IEEE Transactions on Multimedia},
	year={2025},
	publisher={IEEE}
}

@article{liu2026cross,
	title={COMO: Cross-mamba interaction and offset-guided fusion for multimodal object detection},
	author={Liu, Chang and Ma, Xin and Yang, Xiaochen and Zhang, Yuxiang and Dong, Yanni},
	journal={Information Fusion},
	volume={125},
	pages={103414},
	year={2026},
	publisher={Elsevier}
}

@inproceedings{lu2026alignfree,
	title={AlignFree-Net: A Registration-Free Hybrid Framework for Infrared-Visible Object Detection via Semantic Feature Matching},
	author={Lu, Peng and Yu, Shuntong and Li, Zeyang},
	booktitle={2026 11th International Conference on Intelligent Computing and Signal Processing (ICSP)},
	pages={2107--2111},
	year={2026},
	organization={IEEE}
}

@article{luo2016understanding,
  title={Understanding the effective receptive field in deep convolutional neural networks},
  author={Luo, Wenjie and Li, Yujia and Urtasun, Raquel and Zemel, Richard},
  journal={Advances in neural information processing systems},
  volume={29},
  year={2016}
}

@article{luo2026scfdet,
	title={SCFDet: Selective Cross-modal Fusion Network for RGB-Infrared Object Detection in Remote Sensing},
	author={Luo, Fulin and Guan, Yan and Fu, Chuan and Guo, Tan and Nan, Zhixiong and Xiang, Tao and Du, Bo},
	journal={IEEE Transactions on Geoscience and Remote Sensing},
	year={2026},
	publisher={IEEE}
}

@inproceedings{ouyang2023efficient,
  title={Efficient multi-scale attention module with cross-spatial learning},
  author={Ouyang, Daliang and He, Su and Zhang, Guozhong and Luo, Mingzhu and Guo, Huaiyong and Zhan, Jian and Huang, Zhijie},
  booktitle={ICASSP 2023-2023 IEEE international conference on acoustics, speech and signal processing (ICASSP)},
  pages={1--5},
  year={2023},
  organization={IEEE}
}

@article{qingyun2021cross,
	title={Cross-modality fusion transformer for multispectral object detection},
	author={Qingyun, Fang and Dapeng, Han and Zhaokui, Wang},
	journal={arXiv preprint arXiv:2111.00273},
	year={2021}
}

@article{razakarivony2016vehicle,
  title={Vehicle detection in aerial imagery: A small target detection benchmark},
  author={Razakarivony, Sebastien and Jurie, Frederic},
  journal={Journal of Visual Communication and Image Representation},
  volume={34},
  pages={187--203},
  year={2016},
  publisher={Elsevier}
}

@article{shannon1948mathematical,
  title={A mathematical theory of communication},
  author={Shannon, Claude E},
  journal={The Bell system technical journal},
  volume={27},
  number={3},
  pages={379--423},
  year={1948},
  publisher={Nokia Bell Labs}
}

@article{shen2024icafusion,
	title={ICAFusion: Iterative cross-attention guided feature fusion for multispectral object detection},
	author={Shen, Jifeng and Chen, Yifei and Liu, Yue and Zuo, Xin and Fan, Heng and Yang, Wankou},
	journal={Pattern Recognition},
	volume={145},
	pages={109913},
	year={2024},
	publisher={Elsevier}
}

@article{tang2023camf,
	title={CAMF: An interpretable infrared and visible image fusion network based on class activation mapping},
	author={Tang, Linfeng and Chen, Ziang and Huang, Jun and Ma, Jiayi},
	journal={IEEE Transactions on Multimedia},
	volume={26},
	pages={4776--4791},
	year={2023},
	publisher={IEEE}
}

@article{tang2026adaptive,
	title={Adaptive fine-grained fusion network for multimodal UAV object detection},
	author={Tang, Zhanyan and Wu, Zhihao and Li, Mu and Wen, Jie and Zhang, Bob and Xu, Yong and Li, Jianqiang},
	journal={IEEE Transactions on Image Processing},
	year={2026},
	publisher={IEEE}
}

@misc{teledyneflir_adas_dataset,
  author       = {{Teledyne FLIR}},
  title        = {{FREE Teledyne FLIR Thermal Dataset for Algorithm Training}},
  howpublished = {Available online: \url{https://oem.flir.com/solutions/automotive/adas-dataset-form/}},
  year         = {2018},
  note         = {(Accessed on 25 December 2025)}
}

@article{vaswani2017attention,
  title={Attention is all you need},
  author={Vaswani, Ashish and Shazeer, Noam and Parmar, Niki and Uszkoreit, Jakob and Jones, Llion and Gomez, Aidan N and Kaiser, {\L}ukasz and Polosukhin, Illia},
  journal={Advances in neural information processing systems},
  volume={30},
  year={2017}
}

@article{wang2004image,
  title={Image quality assessment: from error visibility to structural similarity},
  author={Wang, Zhou and Bovik, Alan C and Sheikh, Hamid R and Simoncelli, Eero P},
  journal={IEEE transactions on image processing},
  volume={13},
  number={4},
  pages={600--612},
  year={2004},
  publisher={IEEE}
}

@inproceedings{wang2023yolov7,
  title={YOLOv7: Trainable bag-of-freebies sets new state-of-the-art for real-time object detectors},
  author={Wang, Chien-Yao and Bochkovskiy, Alexey and Liao, Hong-Yuan Mark},
  booktitle={Proceedings of the IEEE/CVF conference on computer vision and pattern recognition},
  pages={7464--7475},
  year={2023}
}

@inproceedings{wang2024repvit,
  title={Repvit: Revisiting mobile cnn from vit perspective},
  author={Wang, Ao and Chen, Hui and Lin, Zijia and Han, Jungong and Ding, Guiguang},
  booktitle={Proceedings of the IEEE/CVF conference on computer vision and pattern recognition},
  pages={15909--15920},
  year={2024}
}

@article{wang2024yolov10,
	title={Yolov10: Real-time end-to-end object detection},
	author={Wang, Ao and Chen, Hui and Liu, Lihao and Chen, Kai and Lin, Zijia and Han, Jungong and Ding, Guiguang},
	journal={Advances in neural information processing systems},
	volume={37},
	pages={107984--108011},
	year={2024}
}

@inproceedings{wang2024yolov9,
  title={Yolov9: Learning what you want to learn using programmable gradient information},
  author={Wang, Chien-Yao and Yeh, I-Hau and Mark Liao, Hong-Yuan},
  booktitle={European conference on computer vision},
  pages={1--21},
  year={2024},
  organization={Springer}
}

@article{wang2025carnet,
	title={CARNet: Cross-Attention guided feature reconstruction for RGB-T object detection},
	author={Wang, Sirui and Sun, Guiling and Dong, Liang and Zheng, Bowen},
	journal={Expert Systems with Applications},
	pages={129865},
	year={2025},
	publisher={Elsevier}
}

@article{wang2025pfi,
	title={PFI-Net: A parallel feature interaction network for infrared and visible target detection},
	author={Wang, Xiaoxia and Xi, Jiangtao and Yang, Fengbao and Yang, Yunjia and Li, Minglu},
	journal={Pattern Recognition},
	pages={113003},
	year={2025},
	publisher={Elsevier}
}

@article{wang2026cmms,
	title={CMMS-DETR: Cross-Modal Multi-Scale Feature Fusion Network for UAV Remote Sensing Detection},
	author={Wang, Yin and Mu, Jiliang and Yang, QiHong and Gao, Guanglei and He, Jian and Yu, Junbin and Chou, Xiujian},
	journal={IEEE Journal of Selected Topics in Applied Earth Observations and Remote Sensing},
	year={2026},
	publisher={IEEE}
}

@inproceedings{woo2018cbam,
  title={Cbam: Convolutional block attention module},
  author={Woo, Sanghyun and Park, Jongchan and Lee, Joon-Young and Kweon, In So},
  booktitle={Proceedings of the European conference on computer vision (ECCV)},
  pages={3--19},
  year={2018}
}

@article{wu2025dhanet,
	title={DHANet: Dual-stream Hierarchical Interaction Networks for Multimodal Drone Object Detection},
	author={Wu, Xin and Wang, Li and Guan, Jian and Ji, Haoyang and Xu, Lianming and Hou, Yingyan and Fei, Aiguo},
	journal={IEEE Transactions on Geoscience and Remote Sensing},
	year={2025},
	publisher={IEEE}
}

@article{wu2026cdfnet,
	title={CDFNet: Cross-Dimension Fusion Network with Dual Feature Enhancement for Multimodal Object Detection},
	author={Wu, Wencong and Zhang, Xiuwei and Yin, Hanlin and Zeng, Haorui and Wei, Chenxu and Yu, Lei and Zhang, Yanning},
	journal={Expert Systems with Applications},
	pages={132380},
	year={2026},
	publisher={Elsevier}
}

@article{wu2026ivd,
	title={IVD-NET: An Adaptive Dual-Branch Network for Small Object Detection in Multi-Modal Remote Sensing Images},
	author={Wu, Haibin and Yuan, Pengfei and Liu, Wenbai and Wang, Aili and Yu, Liang and Molnar, Gabor},
	journal={IEEE Journal of Selected Topics in Applied Earth Observations and Remote Sensing},
	year={2026},
	publisher={IEEE}
}

@article{xiao2026saff,
	title={SAFF: A Spatially-Aware Fusion Framework for Effective and Efficient Aerial Object Detection},
	author={Xiao, Hongru and Zhuang, Jiankun and Yang, Bin and Hu, Jinming and Zhu, Junze and Lian, Zhen and Zhao, Sijie and Zhou, Yanmin},
	journal={IEEE Transactions on Geoscience and Remote Sensing},
	year={2026},
	publisher={IEEE}
}

@article{xu2025ela,
  title={ELA: efficient location attention for deep convolution neural networks},
  author={Xu, Wei and Wan, Yi and Zhao, Weina},
  journal={Journal of Real-Time Image Processing},
  volume={22},
  number={4},
  pages={1--14},
  year={2025},
  publisher={Springer}
}

@article{xue2026transfer,
	title={Transfer graph reasoning network for misaligned visible-thermal object detection},
	author={Xue, Xiaotong and Chen, Hongshu and Song, Kechen and Yan, Yunhui and Li, Baihua and Meng, Qinggang},
	journal={Knowledge-Based Systems},
	pages={116271},
	year={2026},
	publisher={Elsevier}
}

@article{yu2015multi,
  title={Multi-scale context aggregation by dilated convolutions},
  author={Yu, Fisher and Koltun, Vladlen},
  journal={arXiv preprint arXiv:1511.07122},
  year={2015}
}

@article{yu2026m2fre,
	title={M2FRe-LDF: Multiscale intra-Modality Feature Refinement and Local Deformable convolution Fusion For multispectral object detection},
	author={Yu, Yaonan and Zhang, Zheng and Song, Zhiming and Ou, Xinyu},
	journal={Journal of Visual Communication and Image Representation},
	pages={104936},
	year={2026},
	publisher={Elsevier}
}

@article{zeng2024mmi,
	title={MMI-Det: Exploring multi-modal integration for visible and infrared object detection},
	author={Zeng, Yuqiao and Liang, Tengfei and Jin, Yi and Li, Yidong},
	journal={IEEE Transactions on Circuits and Systems for Video Technology},
	volume={34},
	number={11},
	pages={11198--11213},
	year={2024},
	publisher={IEEE}
}

@inproceedings{zhai2025mcff,
	title={MCFF-Det: Multispectral Coarse-to-Fine Fusion for Object Detection},
	author={Zhai, Xiaoguang and Xu, Fan and Pan, Yaowen},
	booktitle={2025 IEEE 35th International Workshop on Machine Learning for Signal Processing (MLSP)},
	pages={1--6},
	year={2025},
	organization={IEEE}
}

@inproceedings{zhang2018shufflenet,
  title={Shufflenet: An extremely efficient convolutional neural network for mobile devices},
  author={Zhang, Xiangyu and Zhou, Xinyu and Lin, Mengxiao and Sun, Jian},
  booktitle={Proceedings of the IEEE conference on computer vision and pattern recognition},
  pages={6848--6856},
  year={2018}
}

@inproceedings{zhang2020multispectral,
  title={Multispectral fusion for object detection with cyclic fuse-and-refine blocks},
  author={Zhang, Heng and Fromont, Elisa and Lefevre, S{\'e}bastien and Avignon, Bruno},
  booktitle={2020 IEEE International conference on image processing (ICIP)},
  pages={276--280},
  year={2020},
  organization={IEEE}
}

@article{zhang2026adaptive,
	title={Adaptive Dual Cross-Attention Network for Multispectral Object Detection in Autonomous Driving},
	author={Zhang, Jinlai and Song, Xiaolong and Li, Yucheng and Liang, Diqing and Zhang, Zhiyong and Cai, Jinhu},
	journal={Expert Systems with Applications},
	pages={132012},
	year={2026},
	publisher={Elsevier}
}

@inproceedings{zhang2026mdsf,
	title={MDSF-Det: Modality Decoupling and Synergistic fusion detector},
	author={Zhang, Zhengyang and Lv, Guohua and Gao, Yongbiao and Ma, Guangxiao},
	booktitle={ICASSP 2026-2026 IEEE International Conference on Acoustics, Speech and Signal Processing (ICASSP)},
	pages={10377--10381},
	year={2026},
	organization={IEEE}
}

\end{document}